\documentclass[sigconf]{acmart}

\usepackage{subcaption}
\usepackage{multirow}
\usepackage[table,xcdraw]{xcolor}
\usepackage{balance}
\usepackage{amsmath}
\usepackage{mathtools}
\usepackage{amsthm}
\usepackage{hyperref}
\usepackage{graphicx}
\usepackage{booktabs}
\usepackage{wrapfig}
\usepackage{float}
\usepackage{algorithm}
\usepackage{algorithmic}
\usepackage{etoc}

\AtBeginDocument{%
  }

\copyrightyear{2026}
\acmYear{2026}
\setcopyright{cc}
\setcctype{by}
\acmConference[MM '26] {Proceedings of the 34th ACM International Conference on Multimedia}{November 10--14, 2026}{Rio de Janeiro, Brazil.}
\acmBooktitle{Proceedings of the 34th ACM International Conference on Multimedia (MM '26), November 10--14, 2026, Rio de Janeiro, Brazil}
\acmISBN{979-8-4007-2213-4/2026/11}
\acmDOI{10.1145/3767308.3835408}

\begin{document}

\title{VisionSelector: End-to-End Learnable Visual Token Compression \\ for Efficient Multimodal LLMs}

\author{Jiaying Zhu}
\authornote{Work done during internship at ZTE Corporation.}
\email{zhujy53@mail.ustc.edu.cn}
\affiliation{%
  \institution{University of Science and Technology of China}
  \city{Hefei}
  \country{China}}

\author{Yurui Zhu}
\correspondingauthor
\email{zhu.yurui@zte.com.cn}
\affiliation{%
  \institution{ZTE Corporation}
  \city{Changsha}
  \country{China}}

\author{Xin Lu}
\email{luxion@mail.ustc.edu.cn}
\affiliation{%
  \institution{University of Science and Technology of China}
  \city{Hefei}
  \country{China}
}

\author{Wenrui Yan}
\email{yan.wenrui@zte.com.cn}
\affiliation{%
  \institution{ZTE Corporation}
  \city{Changsha}
  \country{China}}

\author{Dong Li}
\email{dongli6@mail.ustc.edu.cn}
\affiliation{%
  \institution{University of Science and Technology of China}
  \city{Hefei}
  \country{China}
}

\author{Kunlin Liu}
\email{liu.kunlin@zte.com.cn}
\affiliation{%
  \institution{ZTE Corporation}
  \city{Changsha}
  \country{China}}

\author{Xueyang Fu}
\correspondingauthor
\email{xyfu@ustc.edu.cn}
\affiliation{%
  \institution{University of Science and Technology of China}
  \city{Hefei}
  \country{China}
}

\author{Zheng-Jun Zha}
\email{zhazj@ustc.edu.cn}
\affiliation{%
  \institution{University of Science and Technology of China}
  \city{Hefei}
  \country{China}
}

\renewcommand{\shortauthors}{Zhu et al.}

\begin{abstract}
Multimodal Large Language Models (MLLMs) encounter significant computational and memory bottlenecks from the massive number of visual tokens generated by high-resolution images or multi-image inputs. 
Previous token compression techniques are often constrained by heuristic rules that risk discarding critical information. They may suffer from biases, such as attention sinks, that lead to sharp performance drops under aggressive compression ratios. To address these limitations, we reformulate token compression as a lightweight plug-and-play framework and turn it into an end-to-end learnable decision process. Specifically, we propose VisionSelector, a scorer module decoupled from the MLLM backbone that incorporates a differentiable Top-K mechanism and a curriculum annealing strategy to bridge the training–inference gap, enabling efficient and adaptive token selection across various compression rates. Remarkably lightweight with only 12.85M trainable parameters, VisionSelector demonstrates generalization across various compression rates and adaptively identifies critical tokens. This leads to superior performance across the evaluated compression budgets, evidenced by preserving 100$\%$ performance on MME with a 30$\%$ retention budget, outperforming representative heuristic baselines by 12.14 percentage points at a 10$\%$ retention budget, and doubling prefill speed. Our code is available at \url{https://github.com/JulietChoo/VisionSelector}.
\end{abstract}



\maketitle

\section{Introduction}
Multimodal Large Language Models (MLLMs) \cite{liu2023visualllava,bai2025qwen25vl,wang2025internvl35advancingopensourcemultimodal} exhibit remarkable capabilities in complex vision-language tasks \cite{wang2026skyfind,feng2026training,min2026bootstrapping,wang2023large}. However, their superior performance hinges on effectively handling high-information-density visual inputs, such as high-resolution images, multi-image sequences, and videos. This inevitably results in an explosion of visual tokens, exacerbating computational and memory bottlenecks during training and inference. Recent studies reveal significant redundancy in visual information, suggesting that effective token compression \cite{wang2025effivlm,shao2025tokensurvey,yao2025timechat} could alleviate these bottlenecks. Consequently, visual token compression has emerged as a critical research frontier in MLLMs, aiming to enhance efficiency while preserving critical information.

\begin{figure*}[!htbp]
    \centering
    \begin{subfigure}{0.31\textwidth}
        \centering
        \includegraphics[width=\linewidth]{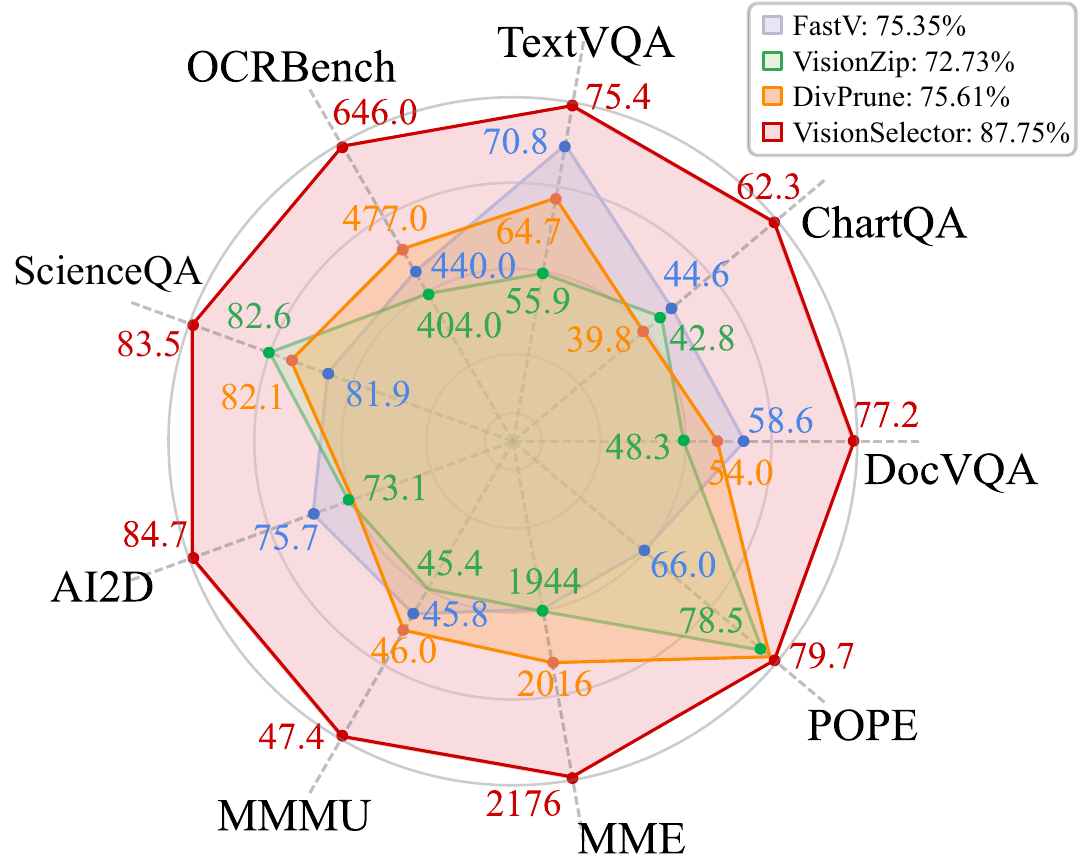}
        \caption{}
        \label{fig:radar}
    \end{subfigure}
    \hfill
    \begin{subfigure}{0.32\textwidth}
        \centering
        \includegraphics[width=\linewidth]{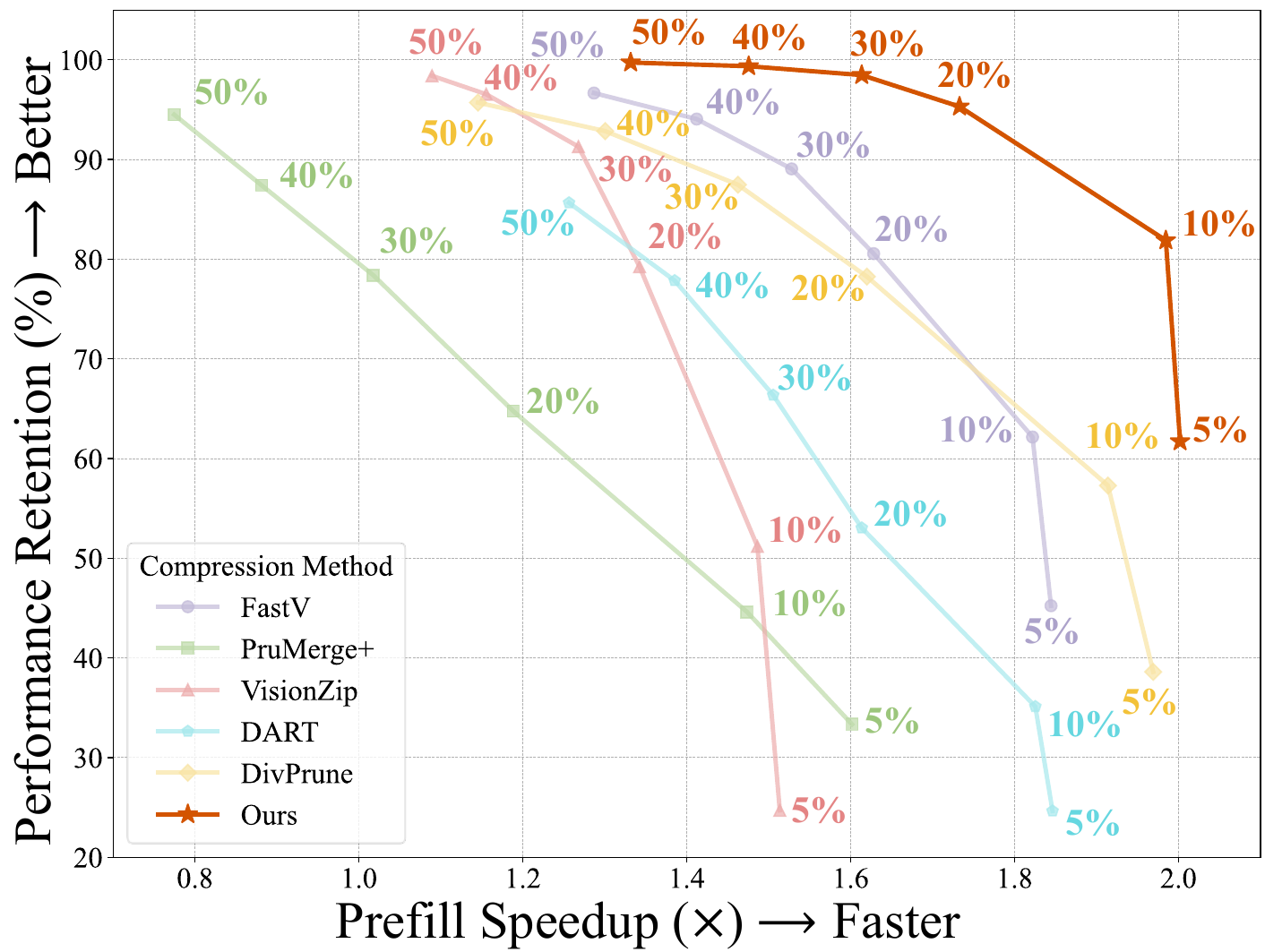}
        \caption{}
        \label{fig:speedup}
    \end{subfigure}
    \hfill
    \begin{subfigure}{0.3\textwidth} 
        \centering
        \includegraphics[width=\linewidth]{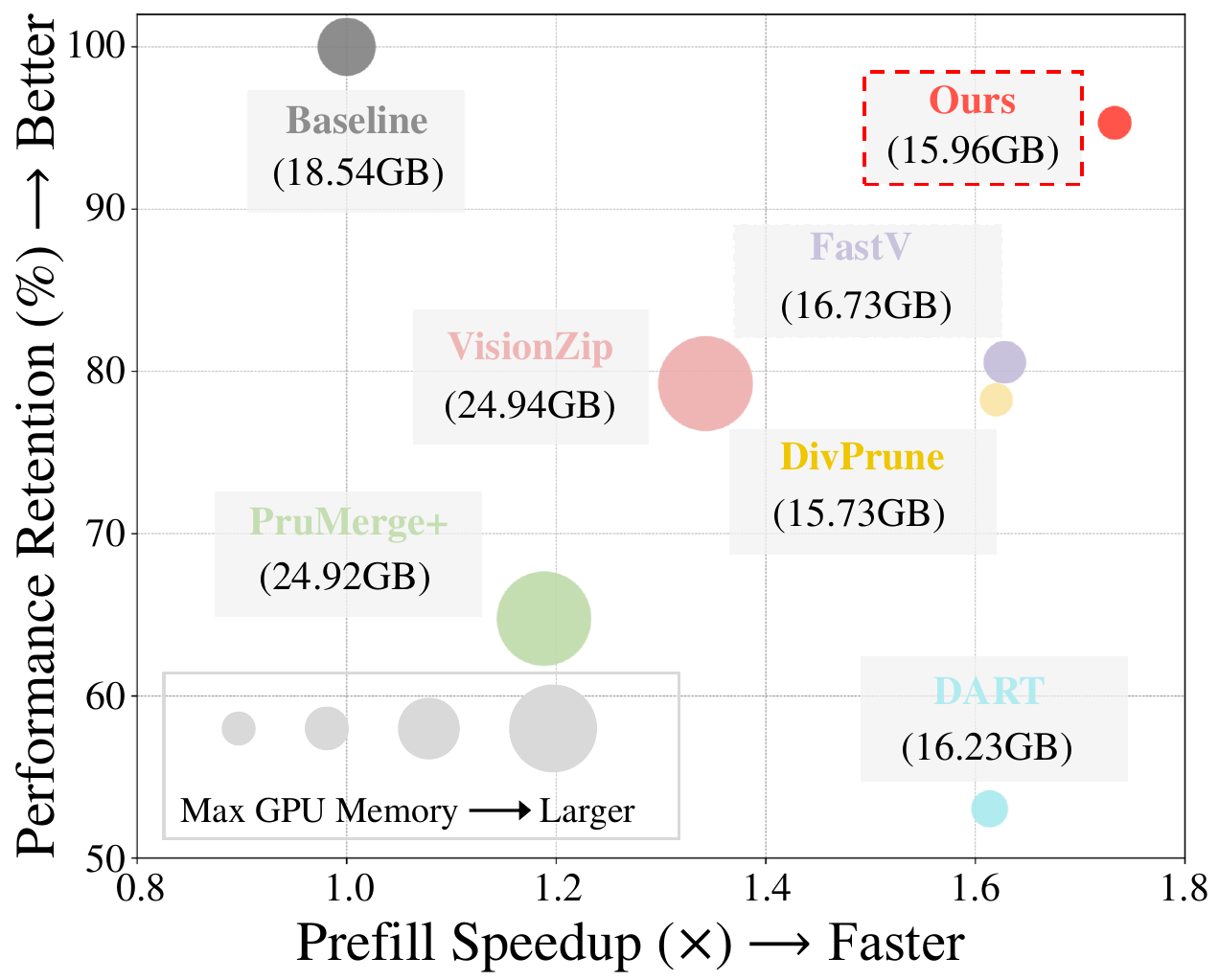}
        \caption{}
        \label{fig:memory}
    \end{subfigure}

    \caption{
    \textbf{Performance and Efficiency of VisionSelector.}
    \textbf{(a)} VisionSelector outperforms representative heuristic baselines under a $10\%$ token budget on Qwen2.5-VL-7B, retaining $87.75\%$ performance on average. 
    \textbf{(b)} Performance–speedup trade-off on DocVQA~\cite{mathew2021docvqa} across 6 retention budgets. VisionSelector is optimal across all ratios, achieving a $2\times$ prefill speedup with only $5\%$ tokens. 
    \textbf{(c)} DocVQA with a $20\%$ token budget: comparison of performance, GPU memory, and speedup. VisionSelector delivers a leading three‑way trade‑off.
    }
    \label{fig:comparison}
    \vspace{-3.5mm}
\end{figure*}

Current token compression methods for vision models exhibit inherent structural trade-offs that constrain their effectiveness. Similarity-based methods \cite{ToMe,wen2025dart,alvar2025divprune} prioritize token diversity but often discard fine-grained, task-critical signals, compromising performance. Attention-based techniques \cite{chen2024imagefastv,yang2025visionzip,shang2024llavaprumerge}, though intuitive, are vulnerable to biases such as attention sink~\cite{zhang2024fastervlm, xiao2023efficient, yang2025visionzip, weng2024longvlm}, which can significantly degrade performance under aggressive compression. Compounding these issues, the reliance on model-specific feature statistics limits generalization \cite{shao2025tokensurvey}, underscoring the need for a more robust, adaptable compression framework.

To overcome these limitations, we reformulate visual token compression as an end-to-end learnable decision process, replacing heuristic rules with optimization-driven token selection. Our approach, VisionSelector, addresses the limitations of prior methods by introducing a lightweight, plug-and-play framework that seamlessly integrates with existing vision-language models and supports adaptive compression. VisionSelector comprises three key components: (1) a Differentiable Top-K Selection Mechanism that employs continuous relaxation to preserve end-to-end gradients while maintaining compatibility with high-performance kernels such as FlashAttention \cite{dao2022flashattention,dao2023flashattention2}; (2) a Curriculum Annealing Strategy paired with the composite loss, which bridges the train–inference gap by progressively transitioning from soft to hard selection for robust importance learning; and (3) a backbone-decoupled Learnable Importance Scorer (LIS) that computes global token importance within a single forward pass, enabling a model trained at a fixed compression rate to generalize to various compression budgets at inference time.
As shown in Figure \ref{fig:comparison}, VisionSelector delivers substantial advancements. Specifically, at a 10\% token retention, it improves over heuristic baselines by 12.14 percentage points. At $20\%$ retention, it accelerates the prefill phase by a factor of 1.73× while simultaneously reducing memory consumption to $86.08\%$.  The contributions of this paper could be summarized as:

\begin{itemize}

\item We propose VisionSelector, a novel framework that reformulates visual token compression from heuristic-based post-processing to an end-to-end learnable decision process, optimized directly by downstream losses.
    
\item We design a lightweight, plug-and-play, and training-efficient module (12.85M parameters) that seamlessly integrates with existing MLLMs. It requires no backbone modification, maintains compatibility with various acceleration techniques (e.g., FlashAttention), and its learned mechanism effectively alleviates heuristic-driven biases such as attention sink.

\item Trained at a single fixed compression rate, our approach demonstrates strong adaptability by generalizing across a range of evaluated compression budgets during inference. Our method significantly reduces memory usage and prefill time while maintaining superior performance, outperforming baselines across various compression budgets on 13 image and video understanding benchmarks.  
\end{itemize}

\vspace{-1.5mm}

\section{Related Work}

\begin{figure*}[!ht]
    \centering
    \includegraphics[width=0.92\textwidth]{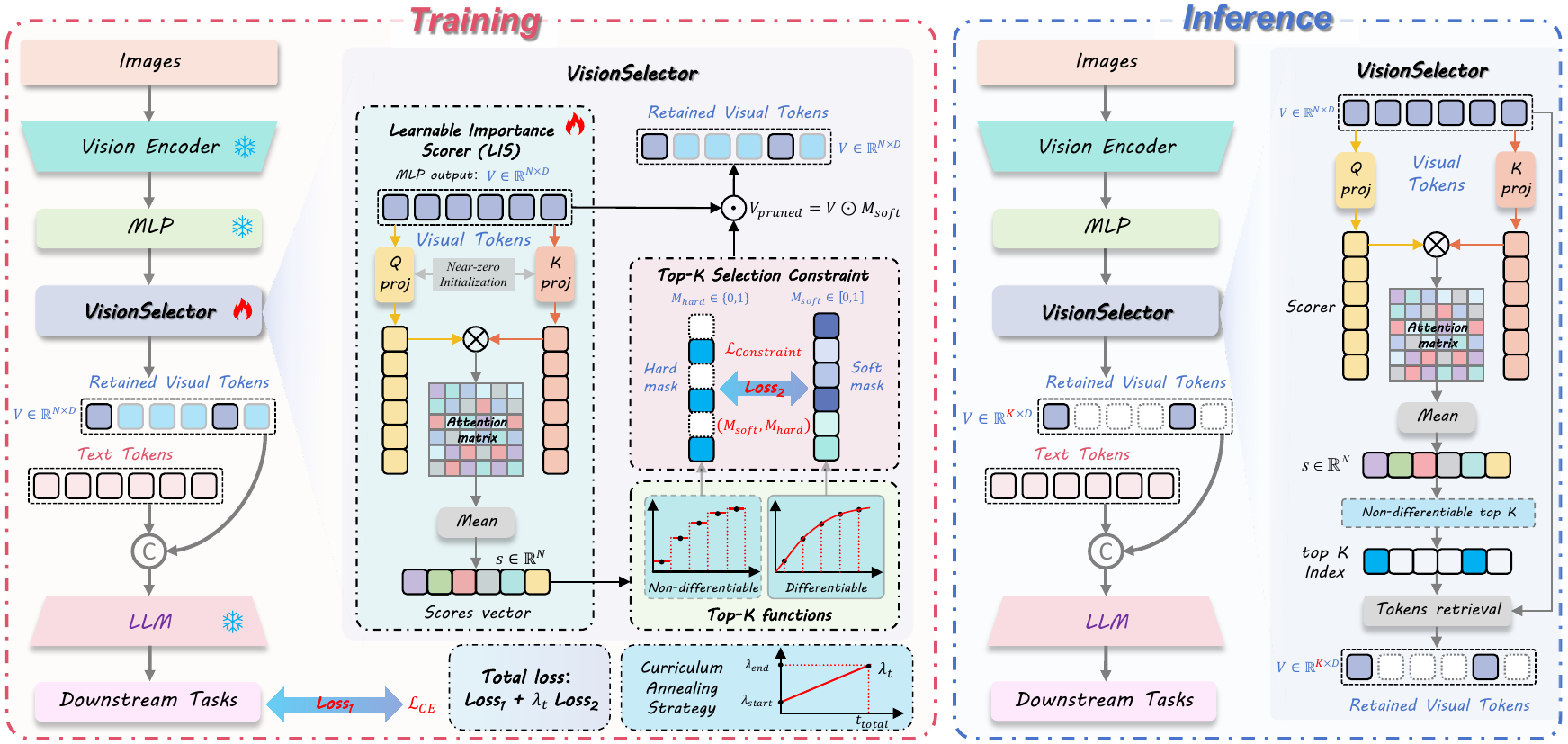}
    \caption{Overview of our VisionSelector. The framework introduces a lightweight, learnable importance scorer to evaluate token importance. During training, the Differentiable Top-K Selection produces a soft mask for gradient propagation, while a constraint loss integrated with a Curriculum Annealing Strategy progressively transforms it into a hard mask. At inference, the standard Top-K operation is applied for efficient token selection.}
    \label{fig:framwork}
    \vspace{-2mm}
\end{figure*}

\textbf{Training-free Token Compression Methods.} Existing training-free token compression methods rely primarily on heuristic rules rather than end-to-end optimization \cite{ma2025short,chen2025lvlm_csp,jiang2025visa,bao2025bolt,dong2025mmtok,wang2025accelerating,zhang2026vscan}. Representative approaches mainly include attention-based pruning and similarity-based merging, along with several other heuristic variants \cite{chen2024imagefastv,shang2024llavaprumerge,arif2025hired,yang2025visionzip,zhangsparsevlm,shao2025holitom,zou2025holov,ToMe,alvar2025divprune,wen2025dart,wang2025folderacceleratingmultimodallarge,yang2025topv,tan2025tokencarve,sun2025llavascissor}. Attention-based strategies compress sequences by discarding tokens with low attention scores. For example, FastV \cite{chen2024imagefastv} selects visual tokens according to the attention they receive from text tokens, while PruMerge+ \cite{shang2024llavaprumerge} identifies important tokens via attention sparsity and merges the remaining ones with K-Nearest Neighbors (KNN) clustering. VisionZip \cite{yang2025visionzip} further combines attention-based dominant token selection with similarity-based contextual token merging, and HoloV \cite{zou2025holov} improves robustness under high pruning ratios by preserving holistic visual context across spatial crops. By contrast, similarity-based approaches compress sequences by grouping redundant tokens. DART \cite{wen2025dart} retains one representative token from near-duplicate groups based on cosine similarity, whereas DivPrune \cite{alvar2025divprune} preserves tokens by maximizing the informational diversity of the retained subset. VISA \cite{jiang2025visa} further introduces a graph-based aggregation strategy, where removed tokens are summarized into retained ones according to semantic similarity, improving the compactness of the compressed representation. Despite their efficiency, these heuristic methods may suffer from performance degradation caused by attention sink \cite{wang2025effivlm} and attention dispersion \cite{zhang2024fastervlm}, risk discarding critical fine-grained information \cite{shao2025tokensurvey}, and often generalize poorly across different architectures and scales. Moreover, because their selection rules are fixed and hand-crafted, they usually cannot adapt their token importance criteria to downstream objectives. This limitation becomes particularly evident under aggressive compression ratios, where small design biases may lead to severe performance collapse.

\textbf{Learning-based Token Selection Methods.}
\label{sec:moredis_learnable}
In contrast to heuristic pruning, learning-based token selection methods train lightweight scoring modules to identify informative tokens while preserving the original token features for dynamic inference. Dynamic-LLaVA \cite{huangdynamicllva} learns an image predictor for binary keep-or-drop token decisions, while LightVLA \cite{jiang2025lightvla} follows a Straight-Through Estimation (STE) \cite{bengio2013ste} style pruning paradigm to optimize discrete token selection. Despite their different implementations, both methods make token decisions locally rather than through a global ranking under an explicit budget constraint. As a result, they remain sensitive to compression-rate variations and may suffer from the mismatch between training-time relaxation and inference-time hard selection. To address this limitation, we propose VisionSelector, a general end-to-end framework that treats MLLM visual token selection as a discrete optimization problem, shifting the paradigm from local independent gating to global differentiable ranking. 

\vspace{-3mm}
\section{Method}
To overcome the limitations of current visual token compression techniques, we propose VisionSelector, a framework that recasts token compression as an end-to-end, task-oriented, and learnable decision process. The framework is built upon three key components: a lightweight Learnable Importance Scorer (LIS) for evaluating global token significance, a Differentiable Top-K Selection (DTS) mechanism that enables gradient backpropagation, and a Curriculum Annealing Strategy (CAS) to align training and inference. This integrated design achieves efficient, flexible token compression while maintaining full compatibility with modern acceleration libraries like FlashAttention.

\vspace{-1.5mm}
\subsection{Overall Framework}
Our method, VisionSelector, is designed as a plug-and-play solution for seamless integration into advanced MLLMs. We deploy it between the modality interface and the large language model. The overall process, as illustrated in Figure \ref{fig:framwork}, can be divided into the following steps: First, the modality interface projects the features from the visual encoder into visual tokens, $\mathbf{V}\in \mathbb{R}^{N \times D}$, required by the LLM, where $N$ is the number of tokens and $D$ is the feature dimension. This set of features $\mathbf{V}$ is then fed into our proposed LIS, which generates an importance score for each token, yielding a score vector $\mathbf{s}\in \mathbb{R}^{N}$. Based on a predefined retention budget $b\in(0,1)$, the DTS utilizes the score vector $\mathbf{s}$ to generate a soft mask, $\mathbf{M}_{\text{soft}}\in [0,1]^{N}$. This soft mask is applied to the original visual token features via element-wise multiplication to suppress the expression of non-critical tokens: $\mathbf{V}_{\text{pruned}}=\mathbf{M}_{\text{soft}}\odot \mathbf{V}$. Finally, the pruned visual features $\mathbf{V}_{\text{pruned}}$ are concatenated with the text token embeddings and fed into the LLM for subsequent processing. The entire model is trained end-to-end using the loss from the downstream task and a constraint loss designed to optimize the selection process.

\vspace{-1.5mm}
\subsection{Learnable Importance Scorer}
As illustrated in Figure~\ref{fig:framwork}, our lightweight Learnable Importance Scorer (LIS) captures each token's relative importance in the global visual context. 
Given visual token embeddings $\mathbf{V}\in\mathbb{R}^{N\times D}$, LIS applies two linear projections to obtain Query and Key:
\begin{equation}
\mathbf{Q}=\mathbf{V}\mathbf{W}_{q},\qquad
\mathbf{K}=\mathbf{V}\mathbf{W}_{k},
\end{equation}
where $\mathbf{W}_{q},\mathbf{W}_{k}\in\mathbb{R}^{D\times d}$ are learnable projection matrices and $d$ is the hidden dimension (thus $\mathbf{Q},\mathbf{K}\in\mathbb{R}^{N\times d}$).
We then compute a simplified self-attention score matrix $\mathbf{A}\in\mathbb{R}^{N\times N}$ and derive per-token importance scores by row-wise averaging:
\begin{equation}
\mathbf{A}=\frac{\mathbf{Q}\mathbf{K}^{\top}}{\sqrt{d}},\qquad
s_i=\frac{1}{N}\sum_{j=1}^{N}A_{ij}.
\end{equation}
Stacking $\{s_i\}_{i=1}^{N}$ yields the score vector $\mathbf{s}\in\mathbb{R}^{N}$.
This design enables global context perception while remaining computationally efficient. To smoothly integrate LIS with the frozen pre-trained model and stabilize early training, we use near-zero weight initialization.

\subsection{Differentiable Top-K Selection Mechanism}
\label{sec:dts}

Selecting the Top-K visual tokens is inherently discrete, making the standard Top-K operator non-differentiable and blocking gradient flow to the importance scorer.
Existing learning-based token compression methods typically bypass this issue by introducing \textbf{independent token-wise gating}, where each token is retained or discarded based on a local threshold.
However, such local decisions fail to explicitly model the relative importance among tokens at the sequence level, and are highly sensitive to compression rate variations at inference time.

Inspired by previous work \cite{xie2020DifferentiableOptimal,petersen2022difftopk,sander2023fast,struski2025lapsum,Differentiable,zhu2025dftopk}, we instead learn token selection via a \textbf{differentiable ranking} with an implicit cardinality constraint.
Given token importance scores $\mathbf{s}\in\mathbb{R}^{B\times N}$ from the LIS and a target budget $k=N \times b$, we define a continuous relaxation that outputs a soft mask
\begin{equation}
\mathbf{M}_{\text{soft}}=\mathrm{DiffTopK}(\mathbf{s})\in(0,1)^{B\times N},
\end{equation}
This soft mask is used only for training-time gradient propagation and at inference we revert to the standard hard Top-K for selection.


\subsubsection{Forward Pass}
\label{sec:dts_forward}

For each sample, we compute $\mathbf{M}_{\text{soft}}$ by applying a shared score shift $t$ to all tokens:
\begin{equation}
\mathbf{M}_{\text{soft}}=\sigma(\mathbf{s}+t),
\quad \text{s.t.} \quad
\sum_{i=1}^{N} \sigma(s_i+t)=k .
\label{eq:dts_forward}
\end{equation}
Here $\sigma(\cdot)$ is the sigmoid function and $t$ is the unique root of
$F(t)=\sum_{i=1}^{N}\sigma(s_i+t)-k$.
Since $F(t)$ is continuous and strictly increasing in $t$, the root can be found efficiently by bisection.

Equation~\eqref{eq:dts_forward} defines a globally coupled relaxation: $t=t(\mathbf{s})$ depends on the entire score vector, hence each mask entry $M_i$ is a function of all scores.
Intuitively, under a fixed budget, increasing one score tends to increase its selection probability while decreasing others through the compensating shift $t$.

The forward operator is order-preserving:
for any $i,j$, $\; s_i>s_j \iff [\mathrm{DiffTopK}(\mathbf{s})]_i>[\mathrm{DiffTopK}(\mathbf{s})]_j$,
because $\sigma(\cdot)$ is strictly increasing and $t$ is shared.
It is also shift-invariant: adding a constant $c$ to all scores leaves $\mathbf{M}_{\text{soft}}$ unchanged, since the solution shifts as $t(\mathbf{s}+c)=t(\mathbf{s})-c$.
These properties ensure a consistent, scale-robust approximation to discrete Top-K during training. 


\subsubsection{Backward Pass via Implicit Constraint}
To enable end-to-end training, we must propagate gradients through the LIS module. Since the threshold $t$ is implicitly determined by the cardinality constraint $\sum_{i=1}^{N} \sigma(s_i + t) = k$, we treat $t$ as a function of the score vector $\mathbf{s}$. By differentiating the constraint equation with respect to an input score $s_j$, we obtain:
\begin{equation}
\begin{split}
\frac{\partial}{\partial s_j}\sum_{i=1}^N \sigma(s_i+t)
&= \sigma'(s_j+t)\Big(1+\frac{\partial t}{\partial s_j}\Big) \\
&\quad + \sum_{i\neq j}\sigma'(s_i+t)\frac{\partial t}{\partial s_j}
= 0.
\end{split}
\end{equation}
Let $v_i = \sigma'(s_i + t)$. Solving for $\frac{\partial t}{\partial s_j}$ yields:
\begin{equation}
\frac{\partial t}{\partial s_j} = -\frac{v_j}{\sum_{i=1}^N v_i}.
\end{equation}
Applying the chain rule to the soft mask $M_i = \sigma(s_i + t)$, the entries of the Jacobian matrix $\mathbf{J} = \frac{\partial \mathbf{M}}{\partial \mathbf{s}}$ are given by:

\begin{equation}
\begin{aligned}
J_{ij} = \frac{\partial M_i}{\partial s_j}
&= \sigma'(s_i + t)\left( \frac{\partial s_i}{\partial s_j}
    + \frac{\partial t}{\partial s_j} \right) \\
&= v_i \left( \delta_{ij} - \frac{v_j}{\sum_{l=1}^N v_l} \right),
\end{aligned}
\end{equation}

where $\delta_{ij}$ is the Kronecker delta. In matrix form, this is expressed as $\mathbf{J} = \text{diag}(\mathbf{v}) - \frac{\mathbf{v}\mathbf{v}^T}{\sum v_i}$. Given the upstream gradient $\mathbf{g} = \frac{\partial L}{\partial \mathbf{M}}$, the gradient with respect to the input scores $\mathbf{s}$ is computed via the vector-Jacobian product:
\begin{equation}
\frac{\partial L}{\partial \mathbf{s}} = \mathbf{g}^T \mathbf{J} = \mathbf{v} \odot \mathbf{g} - \frac{\mathbf{v}^T \mathbf{g}}{\sum_{i=1}^N v_i} \mathbf{v},
\end{equation}
where $\odot$ denotes element-wise multiplication. This formulation ensures stable gradient flow and encourages the scorer to focus on tokens near the decision boundary. 


\subsubsection{Inference}
This continuous relaxation is used only during training for gradient propagation. At inference time, we apply the standard Top‑K operator directly to the score vector to obtain a hard binary mask that selects the top $k$ tokens.

\subsection{Training Objective with Curriculum Annealing Strategy}

To effectively train the LIS and bridge the discrepancy between soft selection during training and hard selection during inference, we design a composite loss function and introduce a curriculum learning-based weight scheduling strategy. The total loss function $L_{\text{total}}$ is defined as:
\begin{equation}
L_{\text{total}}=L_{\text{CE}}+\lambda_{t} L_{\text{constraint}},
\end{equation}
where $L_{\text{CE}}$ is the cross-entropy loss for the downstream task, $L_{\text{constraint}}$ regularizes the selection behavior, and $\lambda_t$ is a time-varying weight.

\begin{table*}[!ht]
\centering
\caption{Comparison results with different methods on Image-Language benchmarks. Note that our method, trained with a fixed 20$\%$ retention budget, exhibits adaptability to varying compression budgets during inference. Dynamic* denotes a learnable baseline. Evaluation is conducted with LMMs-Eval~\cite{zhang2024lmms}.}

\setlength{\tabcolsep}{2pt}
\resizebox{0.87\textwidth}{!}{
\begin{tabular}{ccccccccccc}
\toprule
\multicolumn{1}{c|}{}                                                & \textbf{DocVQA} & \textbf{ChartQA} & \textbf{TextVQA} & \textbf{OCRBench} & \textbf{ScienceQA} & \textbf{AI2D}  & \textbf{MMMU}  & \textbf{MME}     & \multicolumn{1}{c|}{\textbf{POPE}}                          &                                \\
\multicolumn{1}{c|}{\multirow{-2}{*}{\textbf{Method}}}               & ANLS            & Relaxed          & EM               & Score               & EM                 & EM             & Acc            & Score            & \multicolumn{1}{c|}{F1}                                     & \multirow{-2}{*}{\textbf{Avg}} \\ \midrule
\rowcolor[HTML]{EFEFEF} 
\multicolumn{11}{c}{\cellcolor[HTML]{EFEFEF}\textit{Dynamic Resolution(MinPix=256×28×28,MaxPix=2048×28×28),Upper Bound (100\%)}}                                                                                                                                                                                          \\
\rowcolor[HTML]{EFEFEF} 
\multicolumn{1}{c|}{\cellcolor[HTML]{EFEFEF}Avg. Visual Tokens}      & 1951.61         & 596.06           & 976.58           & 652.82            & 323.05             & 510.19         & 601.15         & 867.67           & \multicolumn{1}{c|}{\cellcolor[HTML]{EFEFEF}359.55}         &                                \\
\multicolumn{1}{c|}{Qwen2.5-VL-7B}                                  & 94.33           & 83.40            & 82.84            & 838               & 87.26              & 93.59          & 50.78          & 2342.15          & \multicolumn{1}{c|}{86.19}                                  & 100\%                          \\ \midrule
\rowcolor[HTML]{EFEFEF} 
\multicolumn{11}{c}{\cellcolor[HTML]{EFEFEF}\textit{Retain 30\% Tokens (70\% Compression Ratio)}}                                                                                                                                                                                                                         \\
\multicolumn{1}{c|}{FastV (ECCV2024)}                                & 84.01           & 67.64            & 80.22            & 687               & 83.06              & 86.92          & 49.44          & 2263.58          & \multicolumn{1}{c|}{80.47}                                  & 91.61\%                        \\
\multicolumn{1}{c|}{PruMerge+ (ICCV2025)}                            & 73.95           & 62.24            & 73.71            & 648               & 85.22              & 82.77          & 47.67          & 2239.64          & \multicolumn{1}{c|}{83.69}                                  & 88.00\%                        \\
\multicolumn{1}{c|}{VisionZip (CVPR2025)}                            & 86.11           & 72.28            & 77.30            & 711               & \textbf{86.61}     & 87.86          & 49.44          & 2276.04          & \multicolumn{1}{c|}{84.73}                                  & 93.57\%                        \\
\multicolumn{1}{c|}{HoloV(NeurIPS2025)}                               & 59.24           & 49.28            & 68.79            & 589               & 85.13              & 80.89          & 48.33          & 2196.12          & \multicolumn{1}{c|}{83.27}                                  & 82.75\%                        \\
\multicolumn{1}{c|}{DART (EMNLP2025)}                                & 62.60           & 56.88            & 74.45            & 629               & 84.33              & 75.94          & 47.89          & 2218.83          & \multicolumn{1}{c|}{83.43}                                  & 84.79\%                        \\
\multicolumn{1}{c|}{DivPrune (CVPR2025)}                             & 82.51           & 67.52            & 78.52            & 720               & 86.01              & 88.28          & 48.33          & 2224.06          & \multicolumn{1}{c|}{84.68}                                  & 92.27\%                        \\
\multicolumn{1}{c|}{Dynamic* (ICLR2025)}                             & 86.32           & 68.88            & 73.56            & 750               & 78.78              & 83.29          & 41.78          & 2180.42          & \multicolumn{1}{c|}{80.87}                                  & 88.99\%                        \\
\rowcolor[HTML]{ECF4FF} 
\multicolumn{1}{c|}{\cellcolor[HTML]{ECF4FF}\textbf{VisionSelector}} & \textbf{92.89}  & \textbf{72.96}   & \textbf{81.45}   & \textbf{809}      & 85.77              & \textbf{92.00} & \textbf{50.11} & \textbf{2343.77} & \multicolumn{1}{c|}{\cellcolor[HTML]{ECF4FF}\textbf{85.05}} & \textbf{97.20\%}               \\ \midrule
\rowcolor[HTML]{EFEFEF} 
\multicolumn{11}{c}{\cellcolor[HTML]{EFEFEF}\textit{Retain 20\% Tokens (80\% Compression Ratio)}}                                                                                                                                                                                                                         \\
\multicolumn{1}{c|}{FastV (ECCV2024)}                                & 75.99           & 60.48            & 78.01            & 597               & 82.75              & 82.35          & 49.00          & 2152.74          & \multicolumn{1}{c|}{76.12}                                  & 86.45\%                        \\
\multicolumn{1}{c|}{PruMerge+ (ICCV2025)}                            & 61.09           & 52.56            & 68.72            & 562               & 83.79              & 77.62          & 46.11          & 2219.30          & \multicolumn{1}{c|}{81.74}                                  & 81.91\%                        \\
\multicolumn{1}{c|}{VisionZip (CVPR2025)}                            & 74.75           & 62.04            & 72.03            & 591               & 84.68              & 82.32          & 47.00          & 2168.86          & \multicolumn{1}{c|}{83.23}                                  & 86.43\%                        \\
\multicolumn{1}{c|}{HoloV(NeurIPS2025)}                               & 50.53           & 40.24            & 62.10            & 497               & 82.45              & 75.42          & 47.44          & 2134.89          & \multicolumn{1}{c|}{81.63}                                  & 76.72\%                        \\
\multicolumn{1}{c|}{DART (EMNLP2025)}                                & 50.03           & 47.16            & 67.53            & 537               & 83.34              & 71.04          & 47.00          & 2138.61          & \multicolumn{1}{c|}{80.16}                                  & 78.16\%                        \\
\multicolumn{1}{c|}{DivPrune (CVPR2025)}                             & 73.81           & 57.88            & 73.86            & 648               & 84.33              & 82.29          & 46.33          & 2198.82          & \multicolumn{1}{c|}{83.55}                                  & 86.75\%                        \\
\multicolumn{1}{c|}{Dynamic* (ICLR2025)}                             & 79.21           & 65.92            & 71.73            & 674               & 77.00              & 81.87          & 42.56          & 2117.37          & \multicolumn{1}{c|}{77.22}                                  & 85.51\%                        \\
\rowcolor[HTML]{ECF4FF} 
\multicolumn{1}{c|}{\cellcolor[HTML]{ECF4FF}\textbf{VisionSelector}} & \textbf{89.91}  & \textbf{68.84}   & \textbf{80.05}   & \textbf{770}      & \textbf{85.67}     & \textbf{90.15} & \textbf{49.22} & \textbf{2293.54} & \multicolumn{1}{c|}{\cellcolor[HTML]{ECF4FF}\textbf{84.27}} & \textbf{94.83\%}               \\ \midrule
\rowcolor[HTML]{EFEFEF} 
\multicolumn{11}{c}{\cellcolor[HTML]{EFEFEF}\textit{Retain 10\% Tokens (90\% Compression Ratio)}}                                                                                                                                                                                                                         \\
\multicolumn{1}{c|}{FastV (ECCV2024)}                                & 58.64           & 44.64            & 70.83            & 440               & 81.95              & 75.74          & 45.78          & 1940.91          & \multicolumn{1}{c|}{65.99}                                  & 75.35\%                        \\
\multicolumn{1}{c|}{PruMerge+ (ICCV2025)}                            & 42.08           & 41.56            & 56.87            & 417               & 81.56              & 71.08          & 45.22          & 1948.58          & \multicolumn{1}{c|}{76.52}                                  & 71.48\%                        \\
\multicolumn{1}{c|}{VisionZip (CVPR2025)}                            & 48.29           & 42.84            & 55.94            & 404               & 82.60              & 73.09          & 45.44          & 1944.04          & \multicolumn{1}{c|}{78.46}                                  & 72.73\%                        \\
\multicolumn{1}{c|}{HoloV(NeurIPS2025)}                               & 34.40           & 29.16            & 47.69            & 371               & 78.98              & 69.59          & 45.44          & 1941.13          & \multicolumn{1}{c|}{75.84}                                  & 66.50\%                        \\
\multicolumn{1}{c|}{DART (EMNLP2025)}                                & 33.13           & 34.00            & 53.97            & 415               & 81.85              & 67.10          & 46.44          & 1980.70          & \multicolumn{1}{c|}{71.91}                                  & 68.39\%                        \\
\multicolumn{1}{c|}{DivPrune (CVPR2025)}                             & 54.04           & 39.80            & 64.65            & 477               & 82.15              & 72.80          & 46.00          & 2015.66          & \multicolumn{1}{c|}{79.27}                                  & 75.61\%                        \\
\multicolumn{1}{c|}{Dynamic* (ICLR2025)}                             & 61.17           & 59.96            & 68.25            & 557               & 75.71              & 76.10          & 43.11          & 1977.58          & \multicolumn{1}{c|}{70.80}                                  & 78.35\%                        \\
\rowcolor[HTML]{ECF4FF} 
\multicolumn{1}{c|}{\cellcolor[HTML]{ECF4FF}\textbf{VisionSelector}} & \textbf{77.25}  & \textbf{62.28}   & \textbf{75.37}   & \textbf{646}      & \textbf{83.54}     & \textbf{84.72} & \textbf{47.44} & \textbf{2175.75} & \multicolumn{1}{c|}{\cellcolor[HTML]{ECF4FF}\textbf{79.73}} & \textbf{87.75\%}               \\ \bottomrule
\end{tabular}
}
\label{tab:image}
\end{table*}

\subsubsection{Constraint loss.}
Let $\mathbf{s}\in\mathbb{R}^{N}$ denote the score vector predicted by LIS. During training, the selector outputs a soft mask $\mathbf{M}_{\text{soft}}\in[0,1]^N$, while inference uses a hard Top-K mask $\mathbf{M}_{\text{hard}}\in\{0,1\}^N$. We define the hard mask by
\begin{equation}
\mathbf{M}_{\text{hard}} = \mathrm{TopK}(\mathbf{s};k),
\end{equation}
where $k=\lfloor bN\rfloor$ is the target number of retained tokens under retention ratio $b\in(0,1)$.
We then penalize the discrepancy between $\mathbf{M}_{\text{soft}}$ and $\mathbf{M}_{\text{hard}}$ using a binary cross-entropy (BCE) loss:
\begin{equation}
L_{\text{constraint}}
= \mathrm{BCE}\!\left(\mathbf{M}_{\text{soft}}, \mathbf{M}_{\text{hard}}\right).
\end{equation}
This constraint encourages $\mathbf{M}_{\text{soft}}$ to become polarized towards $\{0,1\}$, making the learned score distribution directly transferable to hard selection during inference.

\subsubsection{Curriculum annealing schedule.}
We progressively increase $\lambda_t$ from a small initial weight $\lambda_{\text{start}}$ to a maximum weight $\lambda_{\text{end}}$:
\begin{equation}
\lambda_t
= \lambda_{\text{start}}
+ \left(\lambda_{\text{end}}-\lambda_{\text{start}}\right)\,
\min\!\left(\frac{t}{T},\,1\right),
\end{equation}
where $t$ is the current training step and $T$ is the total number of training steps. This curriculum annealing strategy lets the model first learn to solve the downstream task and then gradually strengthens the regularization on token selection, leading to more stable joint optimization.

\section{Experiments}

\subsection{Experimental Setting}
\label{sec:experimental_setting}
\textbf{Models, Training Data and Hardware.} All experiments are based on the Qwen2.5-VL-7B \cite{bai2025qwen25vl}, a powerful open-source MLLM that serves as a high-performance baseline for our study. We conduct supervised training by integrating our proposed LIS module into this model. For training, we employ a mixed-dataset strategy. Our training data is a composite of several datasets from Cambrian-737K \cite{tong2024cambrian}, totaling approximately 144K samples. It primarily includes ChartQA \cite{masry2022chartqa}, OCRVQA \cite{mishra2019ocrvqa}, and a $10\%$ random sample of the COCO \cite{lin2014microsoftcoco}. For datasets included in both the training mixture and the evaluation suite (e.g., ChartQA), we strictly use their official training splits for optimization and the held-out validation or test splits for evaluation, ensuring no sample-level overlap between training and evaluation. This composition exposes the model to diverse visual scenarios, including chart understanding, document OCR, and natural images, which enhances its generalization capability. All experiments are conducted on 8 NVIDIA A800 GPUs (80GB). We utilize a distributed data-parallel strategy and leverage DeepSpeed ZeRO Stage 3 \cite{rasley2020deepspeed} to optimize the training process for efficient management of memory and computational resources.

\textbf{Comparison Methods, Evaluation Framework, Datasets, and Tasks.} For comparison, we select the attention-based methods FastV~\cite{chen2024imagefastv}, PruMerge+~\cite{shang2024llavaprumerge}, VisionZip~\cite{yang2025visionzip} and HoloV~\cite{zou2025holov}, as well as the similarity-based methods DART~\cite{wen2025dart} and
DivPrune~\cite{alvar2025divprune} as our baselines. 
We additionally include a re-implementation of the image predictor from Dynamic-LLaVA~\cite{huangdynamicllva} (denoted as Dynamic* in tables) as a learnable baseline. Since the original Dynamic-LLaVA is developed for LLaVA-1.5~\cite{liu2023visualllava} and jointly sparsifies visual and language contexts, we adapt only its image predictor to Qwen2.5-VL-7B to match our visual token compression setting. We follow the original hyperparameter settings where applicable and train it on the same data as VisionSelector. 
To ensure fairness and reproducibility, we conduct a comprehensive evaluation under the LMMs-Eval \cite{zhang2024lmms} framework. 
Our evaluation covers both image and video modalities, utilizing 9 image-language datasets and 4 video-language datasets. \textbf{For image-language understanding}, we focus on assessing performance in information-dense visual tasks. We select a suite of OCR and text-centric VQA datasets, including TextVQA \cite{singh2019textvqa}, DocVQA \cite{mathew2021docvqa}, OCRBench \cite{Liu_2024ocrbench}, and ChartQA \cite{masry2022chartqa}, to test the model's ability to recognize and comprehend embedded text. To measure complex reasoning about object relationships, attributes, and spatial layouts, we employ AI2D \cite{kembhavi2016ai2d} and ScienceQA \cite{lu2022scienceqa}. Furthermore, we test cross-domain knowledge and expert-level reasoning using two challenging comprehensive benchmarks: MME \cite{fu2024mme} and MMMU \cite{yue2024mmmu}. Finally, we use the POPE \cite{li2023pope} benchmark to quantify the model's hallucination levels and ensure content faithfulness. \textbf{For video-language understanding}, we utilize MVBench \cite{li2024mvbench}, SEEDBench \cite{li2023seed}, VideoMME \cite{fu2025video}, and NExT-QA \cite{xiao2021next}. These benchmarks collectively form a comprehensive evaluation suite to measure performance on advanced tasks such as temporal event understanding, dynamic relationship reasoning, and causal question-answering over video content.

\subsection{Implementation Details}
\label{sec:implementation}
\textbf{Model Configuration and Training Strategy.} We seamlessly integrate the LIS module following the modality interface of the baseline model. During training, we adopt a parameter-efficient approach, exclusively updating the parameters of the LIS module while keeping all pre-trained weights frozen. This strategy not only preserves the powerful pre-trained knowledge of the baseline model and is suitable for scenarios with proprietary training sets, but also significantly reduces the computational overhead of training.

\textbf{Hyperparameter Settings.} We train the model for 1 epoch on the mixed dataset. The projection dimension $d$ for $\mathbf{W}_{q}$ and $\mathbf{W}_{k}$ is set to 1792. We use the AdamW optimizer with a cosine annealing learning rate scheduler, setting the initial learning rate to $5e-5$ with a linear warm-up for the first 0.03 epochs. The per-device batch size is 16, and we use 4 steps of gradient accumulation, resulting in an effective global batch size of 256. 
The retention budget for visual tokens is set to $20\%$. For the constraint loss weight, $\lambda_{t}$, we adopt a Curriculum Annealing Strategy, linearly increasing it from an initial value of 0.1 to a final value of 2.0 over the course of training.

\begin{table}[!ht]
\centering
\caption{Comparison results on LLaVA-OneVision-1.5-8B.}
\vspace{-2.0mm}
\setlength{\tabcolsep}{2pt}
\resizebox{0.95\columnwidth}{!}{
\begin{tabular}{cccccc}
\toprule
\multicolumn{1}{c|}{}                                                & \textbf{ChartQA} & \textbf{TextVQA} & \textbf{ScienceQA} & \multicolumn{1}{c|}{\textbf{AI2D}}                          &                                \\
\multicolumn{1}{c|}{\multirow{-2}{*}{\textbf{Method}}}               & Relaxed          & EM               & EM                 & \multicolumn{1}{c|}{EM}                                     & \multirow{-2}{*}{\textbf{Avg}} \\ \midrule
\rowcolor[HTML]{EFEFEF} 
\multicolumn{6}{c}{\cellcolor[HTML]{EFEFEF}\textit{Dynamic Resolution(MinPix=256×28×28,MaxPix=2048×28×28),Upper Bound (100\%)}}                                                                                                \\
\rowcolor[HTML]{EFEFEF} 
\multicolumn{1}{c|}{\cellcolor[HTML]{EFEFEF}Avg. Visual Tokens}      & 595.92           & 978.56           & 320.4              & \multicolumn{1}{c|}{\cellcolor[HTML]{EFEFEF}523.27}         &                                \\
\multicolumn{1}{c|}{LLaVA-OneVision-1.5-8B}                          & 86.72            & 79.51            & 98.56              & \multicolumn{1}{c|}{94.01}                                  & 100\%                          \\ \midrule
\rowcolor[HTML]{EFEFEF} 
\multicolumn{6}{c}{\cellcolor[HTML]{EFEFEF}\textit{Retain 30\% Tokens (70\% Compression Ratio)}}                                                                                                                               \\
\multicolumn{1}{c|}{FastV (ECCV2024)}                                & 68.64            & 72.27            & 91.22              & \multicolumn{1}{c|}{83.52}                                  & 87.86\%                        \\
\multicolumn{1}{c|}{VisionZip (CVPR2025)}                            & 78.04            & 73.25            & 96.23              & \multicolumn{1}{c|}{85.33}                                  & 92.63\%                        \\
\multicolumn{1}{c|}{DART (EMNLP2025)}                                & 75.76            & 76.63            & 96.78              & \multicolumn{1}{c|}{85.01}                                  & 93.09\%                        \\
\multicolumn{1}{c|}{DivPrune (CVPR2025)}                             & 70.64            & 76.03            & 94.65              & \multicolumn{1}{c|}{89.35}                                  & 92.04\%                        \\
\rowcolor[HTML]{ECF4FF} 
\multicolumn{1}{c|}{\cellcolor[HTML]{ECF4FF}\textbf{VisionSelector}} & \textbf{79.68}   & \textbf{78.12}   & \textbf{97.27}     & \multicolumn{1}{c|}{\cellcolor[HTML]{ECF4FF}\textbf{91.77}} & \textbf{96.61\%}               \\ \midrule
\rowcolor[HTML]{EFEFEF} 
\multicolumn{6}{c}{\cellcolor[HTML]{EFEFEF}\textit{Retain 20\% Tokens (80\% Compression Ratio)}}                                                                                                                               \\
\multicolumn{1}{c|}{FastV (ECCV2024)}                                & 59.08            & 67.33            & 87.95              & \multicolumn{1}{c|}{78.89}                                  & 81.49\%                        \\
\multicolumn{1}{c|}{VisionZip (CVPR2025)}                            & 67.56            & 66.28            & 91.87              & \multicolumn{1}{c|}{79.18}                                  & 84.68\%                        \\
\multicolumn{1}{c|}{DART (EMNLP2025)}                                & 67.28            & 72.77            & 95.44              & \multicolumn{1}{c|}{79.57}                                  & 87.65\%                        \\
\multicolumn{1}{c|}{DivPrune (CVPR2025)}                             & 58.64            & 72.73            & 92.46              & \multicolumn{1}{c|}{84.26}                                  & 85.63\%                        \\
\rowcolor[HTML]{ECF4FF} 
\multicolumn{1}{c|}{\cellcolor[HTML]{ECF4FF}\textbf{VisionSelector}} & \textbf{74.72}   & \textbf{76.02}   & \textbf{95.79}     & \multicolumn{1}{c|}{\cellcolor[HTML]{ECF4FF}\textbf{90.45}} & \textbf{93.79\%}               \\ \midrule
\rowcolor[HTML]{EFEFEF} 
\multicolumn{6}{c}{\cellcolor[HTML]{EFEFEF}\textit{Retain 10\% Tokens (90\% Compression Ratio)}}                                                                                                                               \\
\multicolumn{1}{c|}{FastV (ECCV2024)}                                & 38.08            & 56.47            & 81.51              & \multicolumn{1}{c|}{73.15}                                  & 68.86\%                        \\
\multicolumn{1}{c|}{VisionZip (CVPR2025)}                            & 43.12            & 47.17            & 88.70              & \multicolumn{1}{c|}{71.83}                                  & 68.87\%                        \\
\multicolumn{1}{c|}{DART (EMNLP2025)}                                & 47.32            & 63.56            & 89.34              & \multicolumn{1}{c|}{73.51}                                  & 75.84\%                        \\
\multicolumn{1}{c|}{DivPrune (CVPR2025)}                             & 38.24            & 65.48            & 88.20              & \multicolumn{1}{c|}{76.85}                                  & 74.42\%                        \\
\rowcolor[HTML]{ECF4FF} 
\multicolumn{1}{c|}{\cellcolor[HTML]{ECF4FF}\textbf{VisionSelector}} & \textbf{62.12}   & \textbf{67.40}   & \textbf{89.89}     & \multicolumn{1}{c|}{\cellcolor[HTML]{ECF4FF}\textbf{84.29}} & \textbf{84.32\%}               \\ \bottomrule
\end{tabular}
}
\vspace{-6.00mm}
\label{tab:llava-ov-mini}
\end{table}

\begin{table*}[!ht]
\centering
\caption{Performance and efficiency comparisons. Task performance is evaluated across various video-language benchmarks, while efficiency metrics are benchmarked on MVBench.}
\setlength{\tabcolsep}{2pt}
\vspace{-1.5mm}
\resizebox{0.9\textwidth}{!}{
\begin{tabular}{c|ccccc||ccc}
\toprule
\multirow{2}{*}{\textbf{Method}} & \textbf{MVBench} & \textbf{SEEDBench} & \textbf{VideoMME} & \textbf{NExT-QA} & \textbf{Performance} & \textbf{Max GPU} & \textbf{Prefill Time (ms)} & \textbf{E2E Latency} \\
                                 & Acc              & Acc                & Score             & WUPS            & (\%)                 & mem (GB)         & (ms)                       & (ms)                 \\ \midrule
\rowcolor[HTML]{EFEFEF}
Qwen2.5-VL-7B                    & 68.10             & 60.48              & 60.67             & 27.58           & 100\%                & 25.97            & 1413.34                    & 1605.31              \\
FastV (ECCV2024)                            & 65.75            & \textit{OOM}                & 59.15             & 27.00              & 97.31\%              & 27.59            & 851.76                     & 1021.83              \\
DART (EMNLP2025)                             & 65.80             & \textbf{61.00}        & 57.74             & 26.84           & 97.49\%              & 18.93            & 832.58                     & 996.55               \\
DivPrune (CVPR2025)                        & 65.85            & 59.79              & 57.78             & 27.00              & 97.17\%              & \textbf{17.55}   & 1184.00                    & 1345.24              \\
VisionSelector                             & \textbf{66.55}   & 59.82              & \textbf{59.22}    & \textbf{27.10}   & \textbf{98.13\%}     & 17.57            & \textbf{760.82}            & \textbf{924.57}      \\ \bottomrule
\end{tabular}
}
\vspace{-2.5mm}
\label{tab:video}
\end{table*}

\subsection{Image-Language Understanding Evaluation}
To evaluate VisionSelector, we compare a model trained with a $20\%$ budget against multiple baselines across varying budgets, and the results in Table \ref{tab:image} validate the effectiveness of our method while highlighting the advantage of an end-to-end learning paradigm over fixed, heuristic rules.

The experimental results show that VisionSelector consistently outperforms all baselines across compression ratios. At a 20\% budget, VisionSelector retains 94.83\% of the full model’s average performance, exceeding the next-best method by over 8 percentage points; at 10\%, the margin remains 9.40 percentage points even when the learnable baseline Dynamic* is included. When the budget tightens from 20\% to 10\%, attention-based baselines drop sharply (e.g., VisionZip by $\sim$14 points), whereas VisionSelector degrades more gracefully (by $\sim$6 points). We attribute the collapse of baselines to their reliance on pre-trained attention maps: under extreme budgets, biases such as attention sink may preserve early but semantically irrelevant tokens, leading to a performance collapse. In contrast, VisionSelector’s learnable selection is more robust.

On MME, VisionSelector at a $30\%$ retention budget reaches 2343.77, compared with 2342.15 for the full-token baseline, corresponding to $100.07\%$ relative performance in this run. This result indicates that substantial visual token compression can preserve MME performance under this setting. Given the very small margin over the full-token baseline, we do not over-interpret it as conclusive evidence of consistent performance gains from compression.

\textbf{Comparison with Learnable Alternatives.}
As shown in Table~\ref{tab:image}, Dynamic* demonstrates competitive performance on text-oriented benchmarks (e.g., DocVQA, ChartQA, and OCRBench) but exhibits reduced generalization on broader reasoning tasks such as MMMU and POPE.
VisionSelector consistently surpasses Dynamic* across all benchmarks and compression ratios.
The advantage is particularly pronounced under aggressive compression: at 10\% retention, VisionSelector outperforms Dynamic* by 9.40 percentage points in average performance
retention, highlighting the benefit of global differentiable ranking over local independent gating. 

\textbf{Experiments on LLaVA-OneVision-1.5-8B.} To further examine the plug-and-play generality of VisionSelector across different MLLM architectures, we integrate it into LLaVA-OneVision-1.5-8B~\cite{an2025llavaov15}. We freeze the pre-trained backbone and train only the selector for one epoch on approximately 101K samples from the OCRVQA and TextVQA subsets of Cambrian-737K. The selector is trained with a fixed 20\% retention budget and contains 16.87M trainable parameters. As shown in Table~\ref{tab:llava-ov-mini}, VisionSelector achieves average performance retention rates of 96.61\%, 93.79\%, and 84.32\% at the 30\%, 20\%, and 10\% token budgets, consistently outperforming the compared methods. These results demonstrate that VisionSelector generalizes effectively to another MLLM backbone and adapts to unseen compression budgets without retraining.

\subsection{Video-Language Understanding Evaluation}
To further validate the generalization capability of VisionSelector, we evaluate it on video–language understanding tasks. Specifically, we take the model trained exclusively on image data with a $20\%$ budget and directly apply it to four representative video benchmarks: MVBench, SEED-Bench, VideoMME, and NExT-QA.

In selecting baselines, we note that VisionZip and PruMerge+ cause out-of-memory (OOM) errors when integrated with the Qwen2.5-VL architecture, owing to their operational placement. Qwen2.5-VL utilizes a PatchMerger module for initial token compression after the visual encoder. Since both VisionZip and PruMerge+ compute attention at the direct output of the visual encoder, they incur prohibitive computational and memory costs. Consequently, we select FastV, DART, and DivPrune as baselines for this evaluation.

The results in Table \ref{tab:video} show that VisionSelector outperforms FastV, DART and DivPrune across most video datasets. On MVBench, VisionSelector achieves $66.55\%$ accuracy, significantly higher than the baselines. 
Overall, VisionSelector retains $98.13\%$ of the baseline performance on these video tasks.
This demonstrates that importance criteria learned through an end-to-end process generalize more effectively to temporal data and enable the precise identification of key video tokens compared to fixed heuristic rules.

\subsection{Efficiency Analysis}
This section quantifies the computational efficiency of our method. We analyze efficiency on video tasks across three key metrics: max memory usage, prefill time, and end-to-end (E2E) latency.

As detailed in Table \ref{tab:video}, we analyze this using the MVBench dataset, which has a high average token count of 6,828. VisionSelector’s prefill time is only 760.82 ms and its end-to-end latency is 924.57 ms, achieving $1.86\times$ and $1.74\times$ speedups over the baseline model, respectively. It is also significantly faster than all comparison methods. Our method matches DivPrune in memory efficiency by lowering peak memory usage to $17.57$ GB, a $32.3\%$ reduction from the baseline model. This implies that under the same hardware constraints, our method can process longer visual contexts than competing methods, which is crucial for advancing the application of MLLMs in real-world scenarios like long-video understanding.

\subsection{Latency Breakdown Analysis}
\label{sec:latency_breakdown}

To examine whether the LIS overhead offsets the efficiency gains from token compression, we profile the prefill latency of Qwen2.5-VL-7B with and without VisionSelector ($b\!=\!20\%$) on a single A800 GPU using a single-image input. The input yields $N\!=\!682$ visual tokens (after PatchMerger) and $T\!=\!27$ text tokens, forming a full prefix of $S\!=\!N\!+\!T\!=\!709$ tokens. With $20\%$ retention, the compressed prefix is $S'\!=\!\lfloor bN \rfloor\!+\!T\!=\!163$ tokens. All timings are recorded with CUDA events and averaged over 100 runs.

\begin{table}[h]
\centering
\caption{Prefill latency breakdown on Qwen2.5-VL-7B
(single image, A800 GPU, $b\!=\!20\%$).}
\label{tab:latency_breakdown}
\small
\vspace{-1.5mm}
\begin{tabular}{lcccc}
\toprule
\multirow{2}{*}{\textbf{Component}}
  & \multicolumn{2}{c}{\textbf{Baseline} ($S\!=\!709$)}
  & \multicolumn{2}{c}{\textbf{Ours} ($S'\!=\!163$)} \\
\cmidrule(lr){2-3}\cmidrule(lr){4-5}
 & ms & Frac. & ms & Frac. \\
\midrule
ViT Backbone + Merger       & 26.55 & 38.6\% & 26.42 & 55.4\% \\
LIS (Importance Scorer)     & ---   & ---    & 0.14  & 0.3\%  \\
Token Selection (Top-$K$)   & ---   & ---    & 0.09  & 0.2\%  \\
\midrule
LLM Prefill                 & 42.18 & 61.4\% & 21.01 & 44.1\% \\
\midrule
\textbf{End-to-End Prefill} & \textbf{68.73} & 100\%  & \textbf{47.65} & 100\% \\
\midrule
\textbf{Prefill Speedup}    & \multicolumn{4}{c}{$\mathbf{1.44\times}$} \\
\bottomrule
\end{tabular}
\end{table}

As shown in Table~\ref{tab:latency_breakdown}, the ViT backbone latency is nearly identical in both configurations (26.55\,ms vs.\ 26.42\,ms), as VisionSelector does not modify the vision encoder. The LIS scorer and token selection together add only 0.23\,ms, accounting for less than $0.5\%$ of the end-to-end prefill time. The primary speedup comes from the LLM decoder, whose prefill time drops from 42.18\,ms to 21.01\,ms ($2.01\times$) due to the shortened prefix. This yields an overall $1.44\times$ end-to-end prefill speedup. In this case, the additional latency introduced by LIS and token selection is small relative to the decoder-side reduction.

\begin{table*}[!htbp]
\centering
\caption{Ablation study on the impact of training data composition and curriculum annealing strategy ($\lambda_{t}$) on model performance. }
\vspace{-1.5mm}
\setlength{\tabcolsep}{2pt}
\resizebox{0.92\textwidth}{!}{
\begin{tabular}{c|ccc|cccc|ccccc|c}
\toprule
                                  & \multicolumn{3}{c|}{\textbf{Dataset}} &                                          &                                      &                                         &                                & \textbf{DocVQA} & \textbf{OCRBench} & \textbf{MMMU}  & \textbf{MME}     & \textbf{POPE}  &                                \\ \cmidrule{2-4}
\multirow{-2}{*}{\textbf{Config}} & OCRVQA     & ChartQA    & 10\%COCO    & \multirow{-2}{*}{\textbf{$\lambda_{t}$}} & \multirow{-2}{*}{\textbf{BatchSize}} & \multirow{-2}{*}{\textbf{LearningRate}} & \multirow{-2}{*}{\textbf{Dim}} & ANLS            & Score               & Acc            & Score            & F1             & \multirow{-2}{*}{\textbf{Avg}} \\ \midrule
1                                 & \checkmark          &            &             & 0.1$\sim$3                               & 8                                    & 1e-4                                    & 1024                           & 86.85           & 700               & 48.78          & 2238.43          & 81.58          & 92.38\%                        \\
2                                 & \checkmark          & \checkmark          &             & 0.1$\sim$3                               & 8                                    & 1e-4                                    & 1024                           & 87.59           & 701               & 48.78          & 2270.09          & 81.84          & 92.89\%                        \\
3                                 & \checkmark          & \checkmark          & \checkmark           & 0.1$\sim$3                               & 8                                    & 1e-4                                    & 1024                           & 89.34           & 763               & 48.89          & 2282.37          & 83.93          & 95.37\%                        \\
4                                 & \checkmark          & \checkmark          & \checkmark           & 0                                        & 8                                    & 1e-4                                    & 1024                           & 88.88           & 760               & 47.67          & 2284.06          & 83.43          & 94.62\%                        \\
5                                 & \checkmark          & \checkmark          & \checkmark           & 3                                        & 8                                    & 1e-4                                    & 1024                           & 83.69           & 613               & 47.22          & 2172.38          & 83.70          & 88.94\%                        \\
6                                 & \checkmark          & \checkmark          & \checkmark           & 0.1$\sim$2                               & 8                                    & 1e-4                                    & 1024                           & 89.54           & \textbf{779}      & 49.00          & 2272.71          & 83.90          & 95.75\%                        \\
\rowcolor[HTML]{ECF4FF} 
VisionSelector                    & \checkmark          & \checkmark          & \checkmark           & 0.1$\sim$2                               & 16                                   & 5e-5                                    & 1792                           & \textbf{89.91}  & 770               & \textbf{49.22} & \textbf{2293.54} & \textbf{84.27} & \textbf{95.96\%}               \\ \bottomrule
\end{tabular}
}
\label{tab:ablation}
\end{table*}

\begin{figure*}[!htbp]
    \centering
    \includegraphics[width=0.92\textwidth]{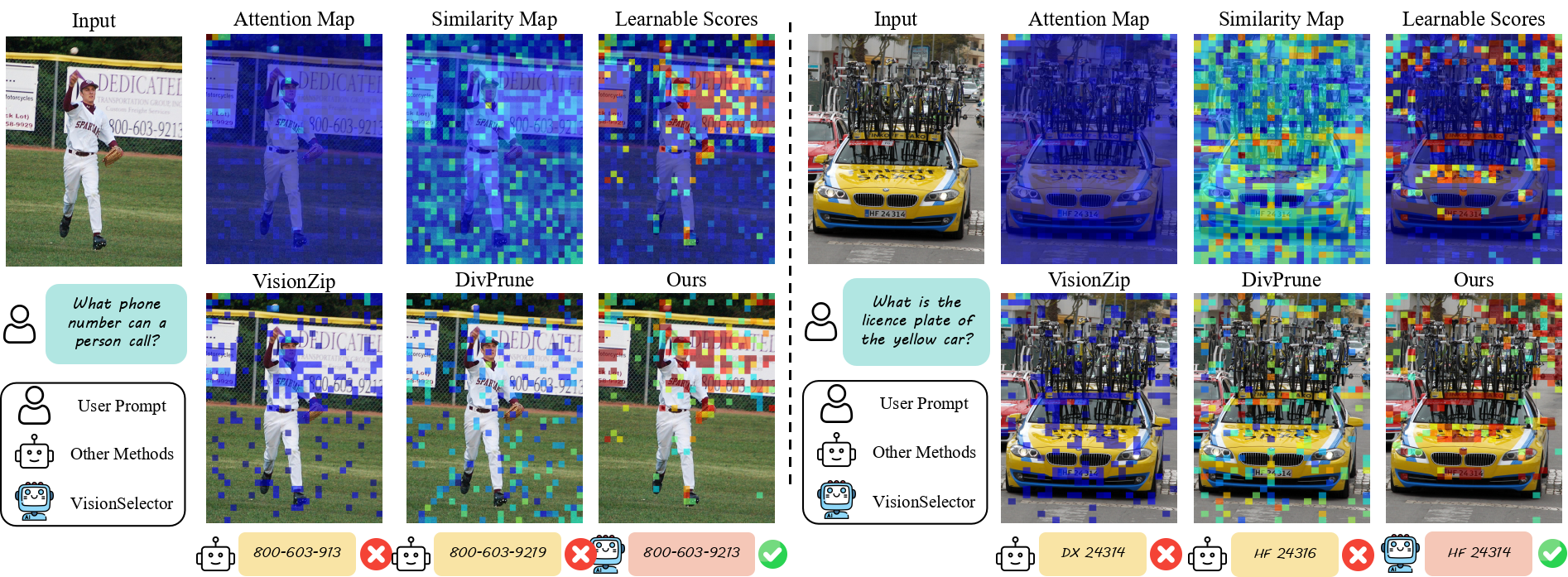}
    \caption{Qualitative results of VisionSelector on TextVQA. The top row compares three token-importance scoring strategies, where warmer colors (red) indicate higher importance scores. In the bottom row, colored overlays denote retained visual tokens, whereas non-overlaid regions correspond to pruned tokens. The corresponding model predictions are obtained under a 20\% retention budget.}
    \vspace{-2.5mm}
\label{fig:vis}
\end{figure*}

\subsection{Ablation Study}
We conduct a series of comprehensive ablation studies to validate the effectiveness of VisionSelector's key components and to determine the optimal hyperparameter settings. All experiments are performed on the Qwen2.5-VL-7B model, and we evaluate the results across several benchmarks spanning multiple dimensions: document understanding (DocVQA, OCRBench), scientific reasoning (ScienceQA, MMMU), and hallucination evaluation (MME, POPE). The detailed configurations and results are presented in Table \ref{tab:ablation}.

\textbf{Impact on training data composition.} As detailed in configurations 1, 2, and 3, we progressively enrich the training data. The results reveal a clear trend: a more diverse training set leads to significant performance gains. Notably, with the inclusion of the general-purpose COCO dataset (Config 3 vs. Config 2), the model's performance on OCR-related tasks improves substantially: the ANLS score on DocVQA increases from $87.59$ to $89.34$, and the OCRBench score jumps from $701$ to $763$. This indicates that exposure to diverse natural images enhances the model's general visual representation capabilities, which in turn benefits its performance on domain-specific tasks. Furthermore, the increase in the F1-score on the POPE dataset suggests that diverse training data also helps to suppress model hallucinations.


\textbf{Impact of the Curriculum Annealing Strategy for $\lambda_{t}$.}
To assess the curriculum annealing of $\lambda_t$, we compare Config~6 with alternative weighting schemes.
Removing the constraint loss (Config~4) yields a lower average score than Config~6, indicating that aligning the soft mask with the hard mask can further improve overall stability and performance, especially during the early stages of joint optimization.
In contrast, using a constant high weight (Config~5) drops the score to $88.94\%$, suggesting that an overly strong early constraint prematurely polarizes token scores and hinders downstream learning.

\subsection{Visualization of Token Selection Results}
As illustrated in Figure~\ref{fig:vis}, heuristic-based pruning methods exhibit clear limitations. VisionZip is prone to issues like attention sink and dispersion, which obscure salient tokens. DivPrune tends to drop semantically rich tokens, such as the logo. Both methods incorrectly preserve low-information background tokens, highlighting their dependence on model-specific feature distributions. In contrast, VisionSelector pinpoints and retains the critical tokens containing the phone number to answer correctly, demonstrating the superiority of its learned selection mechanism.


\vspace{-5mm}
\section{Conclusion}
In this work, we propose VisionSelector, a learnable visual token pruning framework based on differentiable Top-K selection. A lightweight scorer decoupled from the MLLM backbone predicts token importance. During training, soft masks enable stable gradient flow, while a hard-mask constraint with curriculum annealing reduces the training–inference gap. During inference, standard Top-K enables efficient token selection. Extensive experiments on image and video benchmarks show consistent gains over prior token compression baselines, achieving a better balance among performance, latency, and GPU memory across retention budgets. Results on Qwen2.5-VL and LLaVA-OneVision-1.5 further show that VisionSelector is plug-and-play and supports end-to-end training in MLLMs.

\begin{acks}
This work was supported in part by the National Natural Science Foundation of China (NSFC) under Grant 62422609, in part by the CAS Project for Young Scientists in Basic Research under Grant YSBR-122, and in part by the Fundamental Research Funds for the Central Universities under Grant WK2102026001.
\end{acks}

\bibliographystyle{ACM-Reference-Format}
\balance
\bibliography{main}

\clearpage
\appendix

\section{Details about Comparison Method}
\label{sec:app_method}
We select five representative visual token compression methods for comparison, covering several mainstream technical approaches. FastV \cite{chen2024imagefastv} is an attention-based pruning method that operates after the second layer of the MLLM. It selects visual tokens based on the attention scores they receive from text tokens. PruMerge+ \cite{shang2024llavaprumerge} also leverages attention mechanisms but at the visual encoder stage. It identifies key tokens via attention sparsity and then merges the remaining tokens using K-Nearest Neighbors (KNN) clustering. VisionZip \cite{yang2025visionzip} is a text-agnostic compression method. It selects "dominant tokens" based on the attention map from the final visual encoder layer and merges the remaining "contextual tokens" according to semantic similarity. HoloV \cite{zou2025holov} is a plug-and-play visual token pruning framework that preserves holistic visual context under high pruning ratios by combining variance-modulated diversity scoring with attention saliency and dynamically allocating the pruning budget across spatial crops. DART \cite{wen2025dart} identifies groups of near-duplicate tokens by computing the cosine similarity between all visual tokens. It then retains only one representative token from each group, achieving efficient compression while preserving key information. DivPrune \cite{alvar2025divprune} models the token pruning task as a Max-Min Diversity Problem, aiming to maximize the informational diversity of the preserved subset of tokens.

\section{More Detailed Discussions}
\subsection{Discussion on Learning-Based Token Compression Methods and VisionSelector.} 
\label{sec:moredis_learnable}
These approaches explicitly train auxiliary modules to enhance compression and generally fall into three categories. \textbf{Transformation-based} methods reduce token counts or dimensionality via structured operations like pixel-shuffle \cite{wang2025internvl35advancingopensourcemultimodal}, convolution \cite{wei2025deepseekocr}, resampling \cite{li2025tokenpacker,liu2024textmonkey} or MLP projection \cite{bai2025qwen25vl}, while \textbf{query-based} methods distill tokens into fixed representations using learnable queries \cite{li2023blip2,ye2024mplugowl3,alayrac2022flamingo,ye2025voco,zhang2025llavamini}.  Both strategies suffer from fixed compression ratios and altered feature distributions, often necessitating expensive retraining to maintain performance. In contrast, \textbf{token selection-based} methods identify informative tokens through lightweight scoring, preserving original features for dynamic inference. Dynamic-LLaVA \cite{huangdynamicllva} learns an image predictor that assigns each visual token a binary keep-or-drop decision, which is a local gating mechanism because each token is classified independently rather than selected by a globally coordinated ranking under a fixed budget. LightVLA \cite{jiang2025lightvla} introduces a learnable pruning module that selects visual tokens through query-conditioned hard one-hot decisions from token scores, which is local because token selection is decomposed into independent discrete choices instead of a single global ranking over all tokens. These approaches lack explicit modeling of relative token importance and fail to adapt to varying compression rates during inference. To address these bottlenecks, we design a general end-to-end framework. The LIS serves as a lightweight scoring head that provides context-aware input for global ranking. The differentiable Top-K operator acts as a bridge between discrete selection and continuous optimization. It performs a soft relaxation on the ranking to enable gradient propagation to the LIS and allows end-to-end learning of relative token order. The Curriculum Annealing Strategy minimizes the discrepancy between soft and hard masks to ensure inference consistency. VisionSelector is a general end-to-end framework for the discrete optimization problem of MLLM visual token selection. This marks a paradigm shift from local independent gating to global differentiable ranking.

\begin{table*}[!t]
\centering
\caption{Comparison results of our method and learnable alternatives on image-language understanding datasets under Qwen2.5-VL-7B.}
\setlength{\tabcolsep}{2pt}
\resizebox{0.9\linewidth}{!}{
\begin{tabular}{ccccccccccc}
\toprule
\multicolumn{1}{c|}{}                                                                  & \textbf{DocVQA} & \textbf{ChartQA} & \textbf{TextVQA} & \textbf{OCRBench} & \textbf{ScienceQA} & \textbf{AI2D}  & \textbf{MMMU}  & \textbf{MME}     & \multicolumn{1}{c|}{\textbf{POPE}}                          &                                                            \\
\multicolumn{1}{c|}{\multirow{-2}{*}{\textbf{Method}}}                                 & ANLS            & Relaxed          & EM               & Acc               & EM                 & EM             & Acc            & Score            & \multicolumn{1}{c|}{F1}                                     & \multirow{-2}{*}{\textbf{Avg}}                             \\ \midrule
\rowcolor[HTML]{EFEFEF} 
\multicolumn{11}{c}{\cellcolor[HTML]{EFEFEF}\textit{Dynamic Resolution(MinPix=256×28×28,MaxPix=2048×28×28),Upper Bound (100\%)}}                                                                                                                                                                                                                                        \\
\rowcolor[HTML]{EFEFEF} 
\multicolumn{1}{c|}{\cellcolor[HTML]{EFEFEF}Avg. Visual Tokens}                        & 1951.61         & 596.06           & 976.58           & 652.82            & 323.05             & 510.19         & 601.15         & 867.67           & \multicolumn{1}{c|}{\cellcolor[HTML]{EFEFEF}359.55}         &                                                            \\
\multicolumn{1}{c|}{Qwen-2.5-VL-7B}                                                    & 94.33           & 83.40            & 82.84            & 838               & 87.26              & 93.56          & 50.78          & 2342.15          & \multicolumn{1}{c|}{86.19}                                  & 100\%                                                      \\ \midrule
\rowcolor[HTML]{EFEFEF} 
\multicolumn{11}{c}{\cellcolor[HTML]{EFEFEF}\textit{Retain 30\% Tokens (70\% Compression Ratio)}}                                                                                                                                                                                                                                                                       \\
\multicolumn{1}{c|}{}                                                                  & 86.32           & 68.88            & 73.56            & 750               & 78.78              & 83.29          & 41.78          & 2180.42          & \multicolumn{1}{c|}{80.87}                                  &                                                            \\
\multicolumn{1}{c|}{\multirow{-2}{*}{Dynamic* (ICLR2025)}}                              & 91.51\%         & 82.59\%          & 88.80\%          & 89.50\%           & 90.28\%            & 88.99\%        & 82.28\%        & 93.09\%          & \multicolumn{1}{c|}{93.83\%}                                & \multirow{-2}{*}{88.99\%}                                  \\
\multicolumn{1}{c|}{}                                                                  & 71.09           & 49.72            & 71.74            & 559               & 83.79              & 80.54          & 46.89          & 2204.46          & \multicolumn{1}{c|}{84.28}                                  &                                                            \\
\multicolumn{1}{c|}{\multirow{-2}{*}{STE-style Pruning (Arxiv2025)}}                   & 75.36\%         & 59.62\%          & 86.60\%          & 66.71\%           & 96.02\%            & 86.06\%        & 92.34\%        & 94.12\%          & \multicolumn{1}{c|}{97.78\%}                                & \multirow{-2}{*}{83.85\%}                                  \\
\multicolumn{1}{c|}{}                                                                  & 92.00           & 72.84            & 80.67            & 783               & 86.07              & 91.45          & 48.78          & 2311.28          & \multicolumn{1}{c|}{84.05}                                  &                                                            \\
\multicolumn{1}{c|}{\multirow{-2}{*}{SoftSort (ICML2020)}}                             & 97.53\%         & 87.34\%          & 97.38\%          & 93.44\%           & 98.64\%            & 97.71\%        & 96.06\%        & 98.68\%          & \multicolumn{1}{c|}{97.52\%}                                & \multirow{-2}{*}{96.03\%}                                  \\
\rowcolor[HTML]{ECF4FF} 
\multicolumn{1}{c|}{\cellcolor[HTML]{ECF4FF}}                                          & \textbf{92.89}  & \textbf{72.96}   & \textbf{81.45}   & \textbf{809}      & \textbf{85.77}     & \textbf{92.00} & \textbf{50.11} & \textbf{2343.77} & \multicolumn{1}{c|}{\cellcolor[HTML]{ECF4FF}\textbf{85.05}} & \cellcolor[HTML]{ECF4FF}                                   \\
\rowcolor[HTML]{ECF4FF} 
\multicolumn{1}{c|}{\multirow{-2}{*}{\cellcolor[HTML]{ECF4FF}\textbf{VisionSelector}}} & 98.47\%         & 87.48\%          & 98.32\%          & 96.54\%           & 98.29\%            & 98.30\%        & 98.68\%        & 100.07\%         & \multicolumn{1}{c|}{\cellcolor[HTML]{ECF4FF}98.68\%}        & \multirow{-2}{*}{\cellcolor[HTML]{ECF4FF}\textbf{97.20\%}} \\ \midrule
\rowcolor[HTML]{EFEFEF} 
\multicolumn{11}{c}{\cellcolor[HTML]{EFEFEF}\textit{Retain 20\% Tokens (80\% Compression Ratio)}}                                                                                                                                                                                                                                                                       \\
\multicolumn{1}{c|}{}                                                                  & 79.21           & 65.92            & 71.73            & 674               & 77.00              & 81.87          & 42.56          & 2117.37          & \multicolumn{1}{c|}{77.22}                                  &                                                            \\
\multicolumn{1}{c|}{\multirow{-2}{*}{Dynamic* (ICLR2025)}}                              & 83.97\%         & 79.04\%          & 86.59\%          & 80.43\%           & 88.24\%            & 87.48\%        & 83.81\%        & 90.40\%          & \multicolumn{1}{c|}{89.59\%}                                & \multirow{-2}{*}{85.51\%}                                  \\
\multicolumn{1}{c|}{}                                                                  & 57.48           & 39.48            & 59.93            & 428               & 82.99              & 75.16          & 45.56          & 1985.39          & \multicolumn{1}{c|}{81.71}                                  &                                                            \\
\multicolumn{1}{c|}{\multirow{-2}{*}{STE-style Pruning (Arxiv2025)}}                   & 60.94\%         & 47.34\%          & 72.34\%          & 51.07\%           & 95.11\%            & 80.31\%        & 89.72\%        & 84.77\%          & \multicolumn{1}{c|}{94.80\%}                                & \multirow{-2}{*}{75.16\%}                                  \\
\multicolumn{1}{c|}{}                                                                  & 86.90           & 68.20            & 77.76            & 731               & 86.22              & 89.15          & 48.11          & 2259.01          & \multicolumn{1}{c|}{82.77}                                  &                                                            \\
\multicolumn{1}{c|}{\multirow{-2}{*}{SoftSort (ICML2020)}}                             & 92.12\%         & 81.77\%          & 93.87\%          & 87.23\%           & 98.81\%            & 95.26\%        & 94.74\%        & 96.45\%          & \multicolumn{1}{c|}{96.03\%}                                & \multirow{-2}{*}{92.92\%}                                  \\
\rowcolor[HTML]{ECF4FF} 
\multicolumn{1}{c|}{\cellcolor[HTML]{ECF4FF}}                                          & \textbf{89.91}  & \textbf{68.84}   & \textbf{80.05}   & \textbf{770}      & \textbf{85.67}     & \textbf{90.15} & \textbf{49.22} & \textbf{2293.54} & \multicolumn{1}{c|}{\cellcolor[HTML]{ECF4FF}\textbf{84.27}} & \cellcolor[HTML]{ECF4FF}                                   \\
\rowcolor[HTML]{ECF4FF} 
\multicolumn{1}{c|}{\multirow{-2}{*}{\cellcolor[HTML]{ECF4FF}\textbf{VisionSelector}}} & 95.31\%         & 82.54\%          & 96.63\%          & 91.89\%           & 98.18\%            & 96.32\%        & 96.93\%        & 97.92\%          & \multicolumn{1}{c|}{\cellcolor[HTML]{ECF4FF}97.77\%}        & \multirow{-2}{*}{\cellcolor[HTML]{ECF4FF}\textbf{94.83\%}} \\ \midrule
\rowcolor[HTML]{EFEFEF} 
\multicolumn{11}{c}{\cellcolor[HTML]{EFEFEF}\textit{Retain 10\% Tokens (90\% Compression Ratio)}}                                                                                                                                                                                                                                                                       \\
\multicolumn{1}{c|}{}                                                                  & 61.17           & 59.96            & 68.25            & 557               & 75.71              & 76.10          & 43.11          & 1977.58          & \multicolumn{1}{c|}{70.80}                                  &                                                            \\
\multicolumn{1}{c|}{\multirow{-2}{*}{Dynamic* (ICLR2025)}}                              & 64.85\%         & 71.89\%          & 82.39\%          & 66.47\%           & 86.76\%            & 81.31\%        & 84.90\%        & 84.43\%          & \multicolumn{1}{c|}{82.14\%}                                & \multirow{-2}{*}{78.35\%}                                  \\
\multicolumn{1}{c|}{}                                                                  & 36.32           & 22.52            & 37.22            & 244               & 78.38              & 68.91          & 44.56          & 1744.56          & \multicolumn{1}{c|}{75.55}                                  &                                                            \\
\multicolumn{1}{c|}{\multirow{-2}{*}{STE-style Pruning (Arxiv2025)}}                   & 38.50\%         & 27.00\%          & 44.93\%          & 29.12\%           & 89.82\%            & 73.63\%        & 87.75\%        & 74.49\%          & \multicolumn{1}{c|}{87.66\%}                                & \multirow{-2}{*}{61.43\%}                                  \\
\multicolumn{1}{c|}{}                                                                  & 69.61           & 58.24            & 65.62            & 546               & 82.40              & 81.09          & 46.56          & 2014.89          & \multicolumn{1}{c|}{78.09}                                  &                                                            \\
\multicolumn{1}{c|}{\multirow{-2}{*}{SoftSort (ICML2020)}}                             & 73.79\%         & 69.83\%          & 79.21\%          & 65.16\%           & 94.43\%            & 86.64\%        & 91.69\%        & 86.03\%          & \multicolumn{1}{c|}{90.60\%}                                & \multirow{-2}{*}{81.93\%}                                  \\
\rowcolor[HTML]{ECF4FF} 
\multicolumn{1}{c|}{\cellcolor[HTML]{ECF4FF}}                                          & \textbf{77.25}  & \textbf{62.28}   & \textbf{75.37}   & \textbf{646}      & \textbf{83.54}     & \textbf{84.72} & \textbf{47.44} & \textbf{2175.75} & \multicolumn{1}{c|}{\cellcolor[HTML]{ECF4FF}\textbf{79.73}} & \cellcolor[HTML]{ECF4FF}                                   \\
\rowcolor[HTML]{ECF4FF} 
\multicolumn{1}{c|}{\multirow{-2}{*}{\cellcolor[HTML]{ECF4FF}\textbf{VisionSelector}}} & 81.89\%         & 74.68\%          & 90.98\%          & 77.09\%           & 95.74\%            & 90.52\%        & 93.42\%        & 92.90\%          & \multicolumn{1}{c|}{\cellcolor[HTML]{ECF4FF}92.50\%}        & \multirow{-2}{*}{\cellcolor[HTML]{ECF4FF}\textbf{87.75\%}} \\ \bottomrule
\end{tabular}
}
\label{tab:compare_dynamic}
\end{table*}

\subsection{Discussion on Differentiable Sorting Operators and Differentiable Top-K operator}
\label{sec:moredis_learnable}
Differentiable sorting operator \cite{grover2019neuralsort,prillo2020softsort,petersen2022diffsort} relaxes discrete permutations to enable end-to-end optimization. Early methods like NeuralSort \cite{grover2019neuralsort} and SoftSort \cite{prillo2020softsort} construct soft permutation matrices based on pairwise comparisons or distances. However, these approaches involve $O(N^{2})$ complexity and suffer from optimization instability due to strict normalization constraints. Furthermore, they lack precise control over the active token count which causes a sparsity gap between training and inference. In contrast, the Differentiable Top-K operator \cite{xie2020DifferentiableOptimal,petersen2022difftopk,sander2023fast,struski2025lapsum,Differentiable,zhu2025dftopk} directly approximates the selection process with linear time complexity $O(N)$. By avoiding full permutation matrix construction, it mitigates gradient conflicts and balances smoothness with precision via temperature adjustment. We therefore adopt this operator in VisionSelector to ensure stable gradient propagation during training and accurate hard selection during inference.

\section{Comparison with Learnable Alternatives on Qwen2.5-VL-7B}
\label{sec:compare_learnable}

\subsection{Comparison with Dynamic-LLaVA}
\label{sec:dynamic_analy}
To ensure a fair comparison and validate the effectiveness of our approach, we re-implemented the trainable image predictor module from Dynamic-LLaVA \cite{huangdynamicllva} (referred to as Dynamic*) on Qwen2.5-VL-7B. During training, only the parameters of the image predictor are updated, while all other components remain frozen. The training configuration strictly follows the settings described in the original paper: the temperature of the Gumbel Softmax is annealed from $\tau_{\text{start}} =1.0$ to $\tau_{\text{end}} =0.1$, the learning rate of the image predictor is set to $2e-4$, and the batch size is 8. The model is trained for 1 epoch on the same datasets used by our method, including OCRVQA, TextVQA, and $10\%$ randomly sampled COCO images.

The results are summarized in Table~\ref{tab:image} and Table~\ref{tab:compare_dynamic}. Compared with training-free methods, Dynamic* demonstrates strong capability on text-oriented benchmarks (e.g., DocVQA, ChartQA, TextVQA, and OCRBench), though its generalization to broader visual reasoning datasets such as MMMU and POPE is somewhat reduced. In contrast, VisionSelector consistently surpasses Dynamic* across all benchmarks and also outperforms training-free methods in most cases. Remarkably, it maintains strong robustness and generalization even under severe compression (retaining only $20\%$ or $10\%$ of tokens), demonstrating its superior capability in adaptive visual token selection.

\subsection{Comparison with STE-style Pruning}

The Straight-Through Estimator (STE) \cite{bengio2013ste} is a widely adopted technique for estimating gradients through non-differentiable discrete operations. To validate the effectiveness of Differentiable Top-K Operator, we implement an STE-style pruning baseline inspired by LightVLA \cite{jiang2025lightvla}. Specifically, we replace our differentiable Top-K mechanism with their STE-based selection strategy while keeping all other experimental settings identical.

Specifically, after obtaining the importance score map $\mathbf{A} \in \mathbb{R}^{N \times N}$ from the LIS, we convert it into an indicator matrix $\mathbf{I} \in \mathbb{R}^{N \times N}$. The process involves generating both a soft differentiable score and a hard discrete mask:

\begin{align}
\mathbf{A}_{\text{soft}}=\mathrm{softmax}_{j}(\mathbf{A}),
\end{align}
\begin{align}
\mathbf{A}_{\text{hard}}=\mathrm{one-hot}(\mathrm{argmax}_{j}(\mathbf{A})),
\end{align}
\begin{align}
\mathbf{I}=\mathbf{A}_{\text{hard}}+\mathbf{A}_{\text{soft}}-\mathbf{A}_{\text{soft}}^{SG},
\end{align}

where $SG$ denotes the stop-gradient operation. In this formulation, $\mathbf{A}_{\text{hard}}$ represents the binary selection mask used during the forward pass, while gradients are propagated through $\mathbf{A}_{\text{soft}}$ during backpropagation. The pruned visual tokens are then computed as $\mathbf{V}_{\text{pruned}} = \mathbf{I}\mathbf{V}$, retaining their original positional embeddings. Throughout this process, only the LIS module is trainable, with all other components frozen.

The suboptimal results of STE-style pruning in Table \ref{tab:compare_dynamic} likely stem from the absence of global sorting and the inherent gradient mismatch issue \cite{bengio2013ste,jang2017gumbelsoftmax}. Conversely, our DiffTopK approach ensures theoretically guaranteed gradient stability and monotonicity, as demonstrated in DFTopK \cite{zhu2025dftopk} and \ref{sec:gradient_sta}, enabling more effective end-to-end optimization.

\subsection{Comparison with SoftSort}
To investigate the versatility of the VisionSelector framework and the specific contribution of the Differentiable Top-K operator, we conduct an ablation study by replacing the Differentiable Top-K module with the established SoftSort operator \cite{prillo2020softsort}. We maintain identical network architectures and hyperparameters to ensure that any performance variations result solely from the ranking approximation mechanism. Table \ref{tab:compare_dynamic} presents the comparative results on the Qwen2.5-VL-7B backbone. The SoftSort-based variant consistently outperforms heuristic-based methods and the learnable Dynamic* approach across all tested token retention rates. Specifically, it achieves substantial average performance gains ranging from 3.58\% to 7.41\% over Dynamic*. Furthermore, the SoftSort variant yields results comparable to our proposed DiffTop-K implementation and even exhibits marginal superiority on ScienceQA. However, the standard VisionSelector with DiffTop-K generally offers better stability and generalization. This advantage arises because the DiffTop-K operator effectively mitigates gradient conflicts and resolves the sparsity gap between training and inference phases. These findings demonstrate that the VisionSelector framework is compatible with various differentiable operators and does not rely exclusively on a single sorting mechanism. Even with SoftSort, our framework provides stable performance improvements and surpasses baselines such as sparse selection or Gumbel-Softmax gating.

\section{Complexity Analysis of VisionSelector}
\label{app:visionselector_complexity}

We analyze the computational complexity of VisionSelector during training and inference. Let $\mathbf{V}\in\mathbb{R}^{N\times D}$ denote the visual token embeddings output by the modality interface, where $N$ is the number of visual tokens and $D$ is the feature dimension. Given a retention ratio $b\in(0,1)$, the target budget is
\begin{equation}
    k=\max(1,\lfloor bN\rfloor).
\end{equation}
Let $T$ denote the number of non-visual tokens. The multimodal prefix length before and after compression is
\begin{equation}
    S=T+N,\qquad S'=T+k,
\end{equation}
respectively.

\paragraph{LIS complexity.}
The Learnable Importance Scorer computes $\mathbf{Q}=\mathbf{V}\mathbf{W}_{q}$ and $\mathbf{K}=\mathbf{V}\mathbf{W}_{k}$ with $\mathbf{W}_{q},\mathbf{W}_{k}\in\mathbb{R}^{D\times d}$, followed by the score matrix $\mathbf{A}=\mathbf{Q}\mathbf{K}^{\!\top}\!/\sqrt{d}\in\mathbb{R}^{N\times N}$ and row-wise averaging,
\begin{equation}
    s_i=\frac{1}{N}\sum_{j=1}^{N}A_{ij}.
\end{equation}
Under the current implementation, the arithmetic complexity and temporary activation memory of LIS are
\begin{equation}
    \mathcal{C}_{\mathrm{LIS}}=\mathcal{O}(NDd+N^{2}d),\qquad
    \mathcal{M}_{\mathrm{LIS}}=\mathcal{O}(N^{2}+Nd).
\end{equation}

\paragraph{Training-time complexity.}
During training, the differentiable Top-$K$ relaxation solves for a shared threshold via $T_b$ bisection steps, costing $\mathcal{O}(T_bN)$. The soft mask application costs $\mathcal{O}(ND)$, and the hard mask used in the constraint loss costs $\mathcal{O}(N\log N)$. Thus, the selector-side training overhead is
\begin{equation}
    \mathcal{C}_{\mathrm{VS}}^{\mathrm{train}}
    =
    \mathcal{O}(NDd+N^{2}d)
    +
    \mathcal{O}(T_bN+N\log N+ND).
\end{equation}
Since training uses soft reweighting without physically removing tokens, the decoder still processes the full prefix of length~$S$:
\begin{equation}
    \mathcal{C}_{\mathrm{total}}^{\mathrm{train}}
    =
    \mathcal{C}_{\mathrm{VS}}^{\mathrm{train}}
    +
    \mathcal{C}_{\mathrm{prefill}}(S).
\end{equation}

\paragraph{Inference-time complexity.}
At inference, the standard hard Top-$K$ is applied to the score vector, and the retained tokens are gathered to form $\mathbf{V}_{\text{pruned}}\in\mathbb{R}^{k\times D}$. The multimodal prefix is then reconstructed with the shortened visual tokens. The selector overhead is
\begin{equation}
    \mathcal{C}_{\mathrm{VS}}^{\mathrm{infer}}
    =
    \underbrace{\mathcal{O}(NDd+N^{2}d)}_{\text{LIS}}
    +
    \underbrace{\mathcal{O}(N\log N)}_{\text{hard Top-}K}
    +
    \underbrace{\mathcal{O}(S\log S+S'D)}_{\text{index sort \& gather}}.
\end{equation}

\paragraph{Decoder prefill complexity.}
For an $L$-layer decoder with GQA ($h$ query heads, $h_{\mathrm{kv}}$ key-value heads, head dimension $d_h=D/h$) and SwiGLU FFN with intermediate dimension $d_{\mathrm{ff}}$, the prefill complexity at sequence length $S$ is
\begin{equation}\label{eq:prefill}
    \mathcal{C}_{\mathrm{prefill}}(S)
    =
    L\!\left(
    \underbrace{2SD(2h+2h_{\mathrm{kv}})d_h}_{\text{QKV+O proj.}}
    +
    \underbrace{4S^{2}D}_{\text{attention}}
    +
    \underbrace{6SDd_{\mathrm{ff}}}_{\text{SwiGLU FFN}}
    \right).
\end{equation}
FlashAttention~\cite{dao2023flashattention2} substantially reduces the IO cost and memory of the attention term but does not change its arithmetic complexity. The total inference complexity is
\begin{equation}
    \mathcal{C}_{\mathrm{total}}^{\mathrm{infer}}
    =
    \mathcal{C}_{\mathrm{VS}}^{\mathrm{infer}}
    +
    \mathcal{C}_{\mathrm{prefill}}(S'),
\end{equation}
and the net prefill FLOPs reduction is
\begin{equation}
    \Delta\mathcal{C}
    =
    \mathcal{C}_{\mathrm{prefill}}(S)
    -
    \mathcal{C}_{\mathrm{total}}^{\mathrm{infer}}.
\end{equation}

\paragraph{Decoding-stage savings.}
During autoregressive decoding with KV cache, each step computes attention between the new token's query and all cached keys. The context-dependent attention cost per decoding step decreases from $\mathcal{O}(LSD)$ to $\mathcal{O}(LS'D)$, and the KV cache memory decreases from $\mathcal{O}(LSh_{\mathrm{kv}}d_h)$ to $\mathcal{O}(LS'h_{\mathrm{kv}}d_h)$. Over $G$ generated tokens, the cumulative decoding saving is $\mathcal{O}(GL(N-k)D)$.

\paragraph{Numerical example on Qwen2.5-VL-7B.}
Table~\ref{tab:flops_breakdown} instantiates the analysis on Qwen2.5-VL-7B ($L=28$, $D=3584$, $d_{\mathrm{ff}}=18944$, $h=28$, $h_{\mathrm{kv}}=4$, $d_h=128$, $d=1792$) with $N=500$ (a typical single-image input after PatchMerger), $T=50$, and $b=20\%$ (thus $k=100$).

\begin{table}[!t]
\centering
\caption{One-time selector overhead and decoder-side prefill reduction on Qwen2.5-VL-7B with $N=500$, $T=50$, and $b=20\%$.}
\label{tab:flops_breakdown}
\small
\setlength{\tabcolsep}{6pt}
\begin{tabular}{lc}
\toprule
\textbf{Item} & \textbf{GFLOPs} \\
\midrule
\multicolumn{2}{l}{\textit{One-time selector overhead}} \\
\quad LIS: Q/K projections & 12.85 \\
\quad LIS: score matrix & 0.90 \\
\quad Hard Top-$K$ + gather & $<0.01$ \\
\quad Total selector overhead & 13.74 \\
\addlinespace[2pt]
\multicolumn{2}{l}{\textit{Decoder-side prefill computation}} \\
\quad Baseline ($S=550$) & 7{,}299 \\
\quad Compressed prefix ($S'=150$) & 1{,}967 \\
\quad Prefill reduction & 5{,}333 \\
\midrule
\textbf{Net prefill FLOPs reduction} & \textbf{5{,}319} \\
\textbf{Overhead / prefill reduction} & \textbf{0.26\%} \\
\bottomrule
\end{tabular}
\end{table}

\noindent
Here, the selector overhead is
$\mathcal{C}_{\mathrm{VS}}^{\mathrm{infer}}$,
the prefill reduction is
$\mathcal{C}_{\mathrm{prefill}}(S)-\mathcal{C}_{\mathrm{prefill}}(S')$,
and the net reduction is
$\Delta\mathcal{C}
=
\mathcal{C}_{\mathrm{prefill}}(S)-\mathcal{C}_{\mathrm{prefill}}(S')
-\mathcal{C}_{\mathrm{VS}}^{\mathrm{infer}}$.

The one-time selector overhead is only $0.26\%$ of the decoder-side prefill reduction. Under a more demanding scenario with $N=5{,}000$ (e.g., long video), the LIS cost grows to ${\sim}218$\,GFLOPs while the decoder-side reduction exceeds $62$\,TFLOPs, keeping the overhead-to-reduction ratio below $0.35\%$. These results indicate that the $\mathcal{O}(N^{2}d)$ term in LIS remains negligible relative to the decoder-side savings under practical operating regimes.

\section{Theoretical Analysis of the Differentiable Top‑K Operator}
\label{sec:theoretical_analysis}

We provide theoretical insights into the differentiable Top‑K operator in VisionSelector, highlighting its key properties: (i) monotonicity, (ii) shift invariance, (iii) gradient stability, (iv) proof of gradient maximization at the decision boundary, (v) backpropagation and constraint satisfaction, and (vi) optimization dynamics for token retention.

Specifically, the differentiable Top-K operator is defined as:

\begin{equation}
\mathbf{M} = \mathrm{DiffTopK}(\mathbf{s}) = \sigma(\mathbf{s} + t),
\end{equation}
where $\sigma(z) = (1 + e^{-z})^{-1}$ denotes the standard Sigmoid function. The threshold parameter $t$ is dynamically adjusted to satisfy the \textbf{cardinality constraint}:
\begin{equation}
\sum_{i=1}^N \sigma(s_i + t) = k,
\end{equation}
ensuring that the expected number of selected tokens equals the target number $k$.

\subsection{Monotonicity}
\label{sec:mono}
The differentiable Top‑K operator preserves the order of token importance scores, i.e., it is \emph{monotonic} with respect to the input $\mathbf{s}$.
For any two tokens $i$ and $j$, if $s_i > s_j$, then their corresponding selection probabilities satisfy
\begin{equation}
M_i > M_j.
\end{equation}
To see this, note that $M_i = \sigma(s_i + t)$, where $\sigma(s) = (1 + e^{-s})^{-1}$.  
Hence, we analyze its derivative:
\begin{equation}
\frac{d}{ds}\sigma(s) = \frac{d}{ds} (1+e^{-s})^{-1} = \frac{e^{-s}}{(1+e^{-s})^2}.
\end{equation}
This can be rewritten as $\sigma'(s) = \sigma(s)(1-\sigma(s))$. Since the range of the Sigmoid function is $(0, 1)$, both $\sigma(s)$ and $(1-\sigma(s))$ are strictly positive. Therefore, $\sigma'(s) > 0$ for all $s \in \mathbb{R}$, proving that the Sigmoid function is strictly monotonically increasing.

Given $s_i > s_j$, it follows that $s_i + t > s_j + t$. Applying the strictly increasing Sigmoid function, we get:
\begin{equation}
\sigma(s_i + t) > \sigma(s_j + t),
\end{equation}
which implies $M_i > M_j$. Therefore, the ranking structure of $\mathbf{s}$ is preserved under the continuous relaxation $M_i$, and the soft operator provides an order‑consistent approximation of discrete Top‑K, an essential property for differentiable ranking consistency.

\subsection{Shift Invariance}
\label{sec:shift_invar}
The differentiable Top‑K operator is invariant under additive shifts to the input scores, provided the threshold $t$ adapts to satisfy the cardinality constraint.
Formally, for any constant shift $c \in \mathbb{R}$, let the shifted input be $\tilde{\mathbf{s}} = \mathbf{s} + c$. We show that the resulting selection probabilities remain unchanged:
\begin{equation}
M_i(\mathbf{s} + c) = M_i(\mathbf{s}).
\end{equation}
Recall that the selection probability is defined as $M_i = \sigma(s_i + t)$, subject to the constraint $\sum_{i=1}^N \sigma(s_i + t) \approx k$. This implies that $t$ is implicitly determined by the input $\mathbf{s}$ to satisfy this sum.
Let $t^*$ be the threshold that satisfies the constraint for the original input $\mathbf{s}$:
\begin{equation}
\sum_{i=1}^N \sigma(s_i + t^*) = k.
\end{equation}
Now, consider the shifted input $\tilde{s}_i = s_i + c$. We seek a new threshold $\tilde{t}$ such that the constraint holds:
\begin{equation}
\sum_{i=1}^N \sigma(\tilde{s}_i + \tilde{t}) = \sum_{i=1}^N \sigma(s_i + c + \tilde{t}) = k.
\end{equation}
Comparing the two equations, since the function $f(t) = \sum \sigma(s_i + t)$ is strictly monotonic, the argument inside the Sigmoid must be effectively identical to satisfy the same sum $k$. Thus, we must have:
\begin{equation}
c + \tilde{t} = t^* \implies \tilde{t} = t^* - c.
\end{equation}
Substituting $\tilde{t}$ back into the probability formula for the shifted input:
\begin{equation}
\tilde{M}_i = \sigma(\tilde{s}_i + \tilde{t}) = \sigma((s_i + c) + (t^* - c)) = \sigma(s_i + t^*) = M_i.
\end{equation}
This demonstrates that the adaptive threshold $\tilde{t}$ automatically compensates for the input shift $c$, rendering the final selection probabilities $M_i$ invariant to translation.

\subsection{Gradient Stability}
\label{sec:gradient_sta}
The differentiable operator provides stable gradients for end-to-end training. To analyze this, we examine the Jacobian of the selection vector $\mathbf{M}$ with respect to the input scores $\mathbf{s}$. Since the threshold $t$ is determined by the cardinality constraint $\sum_{i=1}^N \sigma(s_i + t) = k$, $t$ is an implicit function of $\mathbf{s}$, which we denote as $t(\mathbf{s})$.

The derivative of the selection probability $M_i = \sigma(s_i + t(\mathbf{s}))$ with respect to an input score $s_j$ is given by the chain rule:
\begin{equation}
\frac{\partial M_i}{\partial s_j} = \frac{\partial \sigma(s_i + t)}{\partial s_j} = \sigma'(s_i + t) \cdot \left( \frac{\partial s_i}{\partial s_j} + \frac{\partial t}{\partial s_j} \right).
\end{equation}
Here, $\frac{\partial s_i}{\partial s_j}$ is the Kronecker delta, $\delta_{ij}$ (which is 1 if $i=j$ and 0 otherwise). Let's denote $\sigma'_i = \sigma'(s_i + t)$. The equation becomes:
\begin{equation}
\frac{\partial M_i}{\partial s_j} = \sigma'_i \left( \delta_{ij} + \frac{\partial t}{\partial s_j} \right).
\label{eq:grad_mij}
\end{equation}
To find the term $\frac{\partial t}{\partial s_j}$, we differentiate the constraint equation $\sum_{l=1}^N \sigma(s_l + t) = k$ with respect to $s_j$:
\begin{equation}
\frac{\partial}{\partial s_j} \sum_{l=1}^N \sigma(s_l + t) = \sum_{l=1}^N \left[ \sigma'_l \cdot \left( \delta_{lj} + \frac{\partial t}{\partial s_j} \right) \right] = 0.
\end{equation}
Solving for $\frac{\partial t}{\partial s_j}$:
\begin{equation}
\sigma'_j + \frac{\partial t}{\partial s_j} \sum_{l=1}^N \sigma'_l = 0 \implies \frac{\partial t}{\partial s_j} = - \frac{\sigma'_j}{\sum_{l=1}^N \sigma'_l}.
\end{equation}
Substituting this back into Eq.~\eqref{eq:grad_mij}, we get the full Jacobian entry:
\begin{equation}
\frac{\partial M_i}{\partial s_j} = \sigma'_i \left( \delta_{ij} - \frac{\sigma'_j}{\sum_{l=1}^N \sigma'_l} \right).
\label{eq:jacobian}
\end{equation}
The derivative of the Sigmoid function, $\sigma'(z) = \sigma(z)(1-\sigma(z))$, has a maximum value of $1/4$ (at $z=0$). This means both $\sigma'_i$ and $\sigma'_j$ are bounded by $1/4$. The term $\sum_{l=1}^N \sigma'_l$ is a sum of $N$ non-negative, bounded values. As long as at least one score $s_l$ is near the threshold $-t$, this sum is non-zero, preventing the gradient from becoming undefined.

Crucially, the gradient magnitude does not depend on the scale of the input scores $\mathbf{s}$, but rather on how close the scores are to the decision boundary $-t$. This structure ensures that gradients are well-behaved and do not explode, which is a key advantage over sorting-based relaxation methods that can suffer from vanishing or exploding gradients\cite{zhu2025dftopk}. This bounded-gradient property allows for a stable soft-to-hard transition during training, enabling robust optimization.

\subsection{Proof of Gradient Maximization at the Decision Boundary}
\label{sec:decision_boundary}
We analyze the magnitude of the Jacobian entries and prove that they are maximized when the selection probabilities are closest to the decision boundary (i.e., $M_i \approx 0.5$).

Recall the Jacobian derived in Eq.~\eqref{eq:jacobian}:
\begin{equation}
J_{ij} = \frac{\partial M_i}{\partial s_j} = \sigma'_i \left( \delta_{ij} - \frac{\sigma'_j}{\sum_{l=1}^N \sigma'_l} \right),
\end{equation}
where $\sigma'_k = \sigma(s_k+t)(1-\sigma(s_k+t))$.
The function $h(z) = \sigma(z)(1-\sigma(z))$ achieves its global maximum of $0.25$ when $\sigma(z) = 0.5$ (i.e., $z=0$). Conversely, $h(z) \to 0$ as $\sigma(z) \to 0$ or $1$. Thus, the term $\sigma'_k$ can be viewed as an "uncertainty indicator," peaking when the token is at the decision boundary.

We analyze the gradient magnitude in two cases:

\textbf{Case 1: Self-Influence (Diagonal terms, $i=j$).}
This term represents how the score of token $i$ affects its own selection probability.
\begin{equation}
J_{ii} = \sigma'_i \left( 1 - \frac{\sigma'_i}{\sum_{l=1}^N \sigma'_l} \right).
\end{equation}
Let $R_i = \sum_{l \neq i} \sigma'_l$ be the sum of the derivative terms of all other tokens. Since $\sigma'_l \ge 0$, we have $R_i \ge 0$. The expression becomes:
\begin{equation}
J_{ii} = \sigma'_i \left( 1 - \frac{\sigma'_i}{\sigma'_i + R_i} \right) = \sigma'_i \left( \frac{R_i}{\sigma'_i + R_i} \right) = \frac{R_i \cdot \sigma'_i}{R_i + \sigma'_i}.
\end{equation}
Consider $J_{ii}$ as a function of the variable $u = \sigma'_i$, denoted as $g(u) = \frac{R_i u}{R_i + u}$, where $u \in (0, 0.25]$.
Taking the derivative with respect to $u$:
\begin{equation}
g'(u) = \frac{R_i(R_i + u) - R_i u}{(R_i + u)^2} = \frac{R_i^2}{(R_i + u)^2}.
\end{equation}
Assuming $R_i > 0$ (which holds if there is at least one other uncertain token), then $g'(u) > 0$. This implies that $J_{ii}$ is a strictly monotonically increasing function of $\sigma'_i$.
Since $\sigma'_i$ is maximized when $\sigma(s_i+t) = 0.5$, the gradient $J_{ii}$ is also maximized at this point.

\textbf{Case 2: Cross-Influence (Off-diagonal terms, $i \neq j$).}
This term represents how the score of token $j$ affects the probability of token $i$.
\begin{equation}
|J_{ij}| = \left| - \frac{\sigma'_i \sigma'_j}{\sum_{l=1}^N \sigma'_l} \right| = \frac{\sigma'_i \sigma'_j}{\sigma'_i + \sigma'_j + R_{ij}},
\end{equation}
where $R_{ij} = \sum_{l \neq i,j} \sigma'_l \ge 0$.
Let $g(u, v) = \frac{uv}{u+v+R_{ij}}$ where $u=\sigma'_i, v=\sigma'_j$. Taking partial derivatives:
\begin{equation}
\frac{\partial g}{\partial u} = \frac{v(u+v+R_{ij}) - uv}{(u+v+R_{ij})^2} = \frac{v^2 + vR_{ij}}{(u+v+R_{ij})^2} > 0.
\end{equation}
By symmetry, $\frac{\partial g}{\partial v} > 0$. Thus, the magnitude of the cross-gradient is maximized when both $\sigma'_i$ and $\sigma'_j$ are maximized, i.e., when both tokens are near the decision boundary ($0.5$).

The gradients are theoretically proven to be strongest for tokens with probabilities near $0.5$. This confirms that the differentiable operator naturally focuses the optimization effort on refining the "hard-to-classify" tokens at the decision boundary, rather than wasting updates on already saturated (confident) tokens.

\subsection{Backpropagation and Constraint Satisfaction}
To understand how the operator updates the input scores $\mathbf{s}$ during training, we analyze the relationship between the upstream gradient $\frac{\partial L}{\partial \mathbf{M}}$ (where $L$ is the loss function) and the computed gradient $\frac{\partial L}{\partial \mathbf{s}}$. By the chain rule, the gradient with respect to the $j$-th input score $s_j$ is computed as the vector-Jacobian product:
\begin{equation}
\frac{\partial L}{\partial s_j} = \sum_{i=1}^N \frac{\partial L}{\partial M_i} \frac{\partial M_i}{\partial s_j}.
\end{equation}
Substituting the Jacobian form derived in Eq.~\eqref{eq:jacobian}, we have:
\begin{equation}
\frac{\partial L}{\partial s_j} = \sum_{i=1}^N \frac{\partial L}{\partial M_i} \left[ \sigma'_i \left( \delta_{ij} - \frac{\sigma'_j}{\sum_{l=1}^N \sigma'_l} \right) \right].
\end{equation}
We can separate this summation into two distinct terms:
\begin{equation}
\frac{\partial L}{\partial s_j} = \underbrace{\frac{\partial L}{\partial M_j} \sigma'_j}_{\text{Local Gradient}} - \underbrace{\left( \sum_{i=1}^N \frac{\partial L}{\partial M_i} \sigma'_i \right) \frac{\sigma'_j}{\sum_{l=1}^N \sigma'_l}}_{\text{Global Correction}}.
\end{equation}
This formulation reveals that the gradient update consists of two mechanisms:
\begin{itemize}
    \item \textbf{Local Gating:} The first term, $\frac{\partial L}{\partial M_j} \sigma'_j$, indicates that the upstream gradient is gated by the derivative $\sigma'_j$. As shown in the previous section, this term is maximized when $s_j$ is near the threshold (where $\sigma'_j$ peaks), effectively focusing updates on uncertain tokens.
    \item \textbf{Global Constraint Correction:} The second term subtracts a weighted average of the gradients from all tokens. We can rewrite the final gradient as:
\end{itemize}
\begin{equation}
\frac{\partial L}{\partial s_j} = \sigma'_j \left( \frac{\partial L}{\partial M_j} - \frac{\sum_{i=1}^N \frac{\partial L}{\partial M_i} \sigma'_i}{\sum_{l=1}^N \sigma'_l} \right).
\end{equation}
The term inside the parentheses represents a "centered" gradient. By subtracting the weighted mean of the upstream gradients, the operator ensures that the total change in the output probabilities respects the cardinality constraint. This mechanism introduces a competitive dynamic: for a token to increase its selection probability ($s_j \uparrow$), it must overcome not just its own threshold, but the collective tendency of all other tokens, ensuring the fixed number $k$ is maintained.

\subsection{Optimization Dynamics for Token Retention}

We further analyze the gradient dynamics to understand how the operator promotes the retention of task-relevant tokens. Our derivation indicates that the competitive nature of the gradient update creates a corrective pressure that favors tokens with significant contributions to loss reduction.

\textbf{Gradient Competition Analysis.}
Based on the Jacobian derivation in Eq.~\eqref{eq:jacobian}, the gradient with respect to the input score $s_k$ incorporates a centering term:
\begin{equation}
\frac{\partial L}{\partial s_k} = \sigma'_k \left( \frac{\partial L}{\partial M_k} - \bar{g}_{\text{avg}} \right), \quad \text{where } \bar{g}_{\text{avg}} = \frac{\sum_{l} \frac{\partial L}{\partial M_l} \sigma'_l}{\sum_{l} \sigma'_l}.
\end{equation}
Here, $\bar{g}_{\text{avg}}$ represents the weighted average gradient of the current token set. This formulation implies that the update direction for a specific token $k$ depends on the relative magnitude of its individual gradient $\frac{\partial L}{\partial M_k}$ compared to the group average $\bar{g}_{\text{avg}}$.

\textbf{Corrective Update Trajectory.}
We consider a scenario where the model excludes a potentially informative token $i$ while retaining a less relevant token $j$. An informative token typically exhibits a large negative gradient component $\frac{\partial L}{\partial M_i}$ (indicating that its inclusion effectively reduces the loss), whereas a less relevant token $j$ has a gradient $\frac{\partial L}{\partial M_j}$ close to zero. Assuming the signal from the informative token is sufficiently strong such that $|\frac{\partial L}{\partial M_i}| > |\bar{g}_{\text{avg}}|$, we observe the following dynamics:
\begin{enumerate}
    \item \textbf{Upward Pressure on Informative Token $i$:}
    Since $\frac{\partial L}{\partial M_i}$ is negative and dominates the average term, the difference $\left( \frac{\partial L}{\partial M_i} - \bar{g}_{\text{avg}} \right)$ remains negative. In gradient descent optimization ($s \leftarrow s - \eta \nabla s$), this negative gradient results in a positive update to the score $s_i$. Consequently, the optimization applies upward pressure on the informative token, promoting its entry into the selected set.
    \item \textbf{Downward Pressure on Less Relevant Token $j$:}
    For the less relevant token where $\frac{\partial L}{\partial M_j} \approx 0$, the gradient is dominated by the term $-\bar{g}_{\text{avg}}$. If the collective gradient average $\bar{g}_{\text{avg}}$ is negative (driven by other informative tokens in the batch), the resulting term $-\bar{g}_{\text{avg}}$ becomes positive. This positive gradient leads to a negative update for score $s_j$, thereby suppressing the score of the less relevant token.
\end{enumerate}
This analysis demonstrates that the differentiable operator enforces a competitive update process. Rather than relying on heuristic selection, the operator utilizes the relative gradient magnitude to iteratively refine the subset, naturally aligning the selection with the objective of minimizing the task loss.

\subsection{Summary}
Collectively, these theoretical analyses establish a rigorous mathematical foundation for the VisionSelector framework. The properties of \textbf{monotonicity} and \textbf{shift invariance} ensure that the soft operator preserves the ranking order of input scores and remains robust to constant additive shifts. The investigation into \textbf{gradient stability} and \textbf{maximization at the decision boundary} confirms that the operator produces bounded gradients that focus optimization efforts on refining uncertain tokens (where selection probability is near 0.5), rather than saturated ones. Furthermore, the decomposition of \textbf{backpropagation} reveals a mechanism of global constraint correction, which enforces the cardinality constraint by subtracting a weighted mean from the gradients. Finally, the analysis of \textbf{token retention} illustrates that this competitive gradient dynamic applies upward pressure to informative tokens and downward pressure to less relevant ones, thereby promoting the selection of task-essential features based on their contribution to loss reduction.

\section{Algorithmic Procedure of $\mathrm {DiffTopK}$}

Following Equation \ref{eq:dts_forward}, Differentiable Top‑K solves the sigmoid threshold $t$ during the forward pass via the method of undetermined coefficients and obtains $t$ by binary search, yielding a smooth approximation to Top‑K so that each probability $M_{i}$ lies in $[0,1]$. Owing to the strict monotonicity of $\sigma(\cdot)$, its compensability under global shifts (via adjusting $t$), and its saturation behavior in the low-temperature or extreme-threshold limit, the mapping $\mathbf{s}\to \mathbf{M}_{\text{soft}}=\sigma (\mathbf{s}+t)$ satisfies monotonicity, invariance to global shifts, and convergence toward standard Top‑K.

During backpropagation, the threshold $t$ varies with $\mathbf{s}$ but they satisfy an implicit constraint. Differentiating this implicit equation yields a closed‑form gradient, and the resulting backward computation appears in Algorithm \ref{alg:diff_topk_en}.

\begin{minipage}{0.55\columnwidth}
\begin{algorithm}[H]
    \small 
    \caption{Differentiable Top-$K$}
    \label{alg:diff_topk_en}
    \begin{algorithmic}
        \STATE {\bfseries Input:} $\mathbf{s} \in \mathbb{R}^{B \times N}$ (scores), $k \in \mathbb{Z}^+$ (select count)
        \STATE {\bfseries Output:} $\mathbf{M} \in [0,1]^{B \times N}$ (soft mask)
        
        \STATE {\bfseries Procedure} \textsc{FindThresh}($\mathbf{s}, k$):
        \STATE \quad $\mathit{lower} \gets -\max(\mathbf{s}, \mathrm{dim}=1) - 10$
        \STATE \quad $\mathit{upper} \gets -\min(\mathbf{s}, \mathrm{dim}=1) + 10$
        \FOR{$i = 1$ {\bfseries to} $64$}
            \STATE $\mathit{mid} \gets (\mathit{lower} + \mathit{upper})/2$
            \STATE $\mathit{m\_sum} \gets \sum \sigma(\mathbf{s} + \mathit{mid})$
            \STATE $\mathit{mask} \gets \mathit{m\_sum} < k$
            \STATE $\mathit{lower}[\mathit{mask}] \gets \mathit{mid}[\mathit{mask}]$
            \STATE $\mathit{upper}[\neg \mathit{mask}] \gets \mathit{mid}[\neg \mathit{mask}]$
        \ENDFOR
        \STATE \quad \textbf{return} $(\mathit{lower} + \mathit{upper})/2$
        
        \STATE {\bfseries Procedure} \textsc{Forward}($\mathbf{s}, k$):
        \STATE \quad $\mathbf{t} \gets \mathrm{FindThresh}(\mathbf{s}, k)$
        \STATE \quad \textbf{return} $\sigma(\mathbf{s} + \mathbf{t})$
        
        \STATE {\bfseries Procedure} \textsc{Backward}($\mathit{grad}$):
        \STATE \quad $\mathbf{v} \gets \sigma'(\mathbf{s} + \mathbf{t})$
        \STATE \quad $s_{\mathrm{sum}} \gets \sum \mathbf{v}$
        \STATE \quad $uv \gets \mathit{grad} \odot \mathbf{v}$
        \STATE \quad $\mathit{uv}_{\mathrm{sum}} \gets \sum uv$
        \STATE \quad $\mathit{grad}_s \gets uv - (\mathit{uv}_{\mathrm{sum}}/s_{\mathrm{sum}}) \odot \mathbf{v}$
        \STATE \quad \textbf{return} $\mathit{grad}_s$
    \end{algorithmic}
\end{algorithm}
\end{minipage}

\section{Additional Experiments on Ablation Study}
\label{sec:app_ablation}
To investigate the impact of hyperparameters (batch size, learning rate) and the score projector dimension $d$, we provide a detailed ablation study in Table \ref{tab:ablation_appendix}.

\begin{table*}[!t]
\centering
\caption{Ablation study on the impact of training data composition, curriculum annealing strategy ($\lambda_{t}$), and hyperparameters on model performance.}
\setlength{\tabcolsep}{2pt}
\resizebox{0.9\textwidth}{!}{
\begin{tabular}{c|ccc|ccccc|cccccc|c}
\toprule
                                  & \multicolumn{3}{c|}{\textbf{Dataset}} &                                          &                                      &                                         &                                &                                 & \textbf{DocVQA} & \textbf{OCRBench} & \textbf{ScienceQA} & \textbf{MMMU} & \textbf{MME} & \textbf{POPE} &                                \\ \cmidrule{2-4}
\multirow{-2}{*}{\textbf{Config}} & OCRVQA      & ChartQA      & COCO     & \multirow{-2}{*}{\textbf{$\lambda_{t}$}} & \multirow{-2}{*}{\textbf{BatchSize}} & \multirow{-2}{*}{\textbf{LearningRate}} & \multirow{-2}{*}{\textbf{Dim}} & \multirow{-2}{*}{\textbf{Init}} & ANLS            & Acc               & EM                 & Acc           & Score        & F1            & \multirow{-2}{*}{\textbf{Avg}} \\ \midrule
1                                 & \checkmark           &              &          & 0.1$\sim$3                               & 8                                    & 1e-4                                    & 1024                           & Near Zero                       & 86.85           & 700               & 85.47              & 48.78         & 2238.43      & 81.58         & 93.31\%                        \\
2                                 & \checkmark           & \checkmark            &          & 0.1$\sim$3                               & 8                                    & 1e-4                                    & 1024                           & Near Zero                       & 87.59           & 701               & 84.78              & 48.78         & 2270.09      & 81.84         & 93.60\%                        \\
3                                 & \checkmark           & \checkmark            & \checkmark        & 0.1$\sim$3                               & 8                                    & 1e-4                                    & 1024                           & Near Zero                       & 89.34           & 763               & 85.52              & 48.89         & 2282.37      & 83.93         & 95.81\%                        \\
4                                 & \checkmark           & \checkmark            & \checkmark        & 3                                        & 8                                    & 1e-4                                    & 1024                           & Near Zero                       & 83.69           & 613               & 84.48              & 47.22         & 2172.38      & 83.70         & 90.26\%                        \\
5                                 & \checkmark           & \checkmark            & \checkmark        & 0.1$\sim$2                               & 8                                    & 1e-4                                    & 1024                           & Near Zero                       & 89.54           & 779               & 84.68              & 49.00         & 2272.71      & 83.90         & 95.97\%                        \\
6                                 & \checkmark           & \checkmark            & \checkmark        & 0.1$\sim$2                               & 16                                   & 1e-4                                    & 1024                           & Near Zero                       & 89.98           & 772               & 84.88              & 49.22         & 2279.07      & 83.94         & 96.07\%                        \\
7                                 & \checkmark           & \checkmark            & \checkmark        & 0.1$\sim$2                               & 32                                   & 1e-4                                    & 1024                           & Near Zero                       & 88.51           & 730               & 85.13              & 48.56         & 2247.40      & 82.89         & 94.38\%                        \\
8                                 & \checkmark           & \checkmark            & \checkmark        & 0.1$\sim$2                               & 16                                   & 5e-5                                    & 1024                           & Near Zero                       & 89.95           & 756               & 85.57              & 49.67         & 2303.66      & 83.56         & 96.13\%                        \\
9                                 & \checkmark           & \checkmark            & \checkmark        & 0.1$\sim$2                               & 16                                   & 5e-5                                    & 1792                           & Random                          & 89.36           & 777               & 85.42              & 48.56         & 2311.13      & 84.92         & 96.36\%                        \\
10                                & \checkmark           & \checkmark            & \checkmark        & 0.1$\sim$2                               & 16                                   & 5e-5                                    & 1792                           & Kaiming                         & 89.61           & 761               & 85.52              & 49.00         & 2311.41      & 84.83         & 96.23\%                        \\
\rowcolor[HTML]{ECF4FF} 
11(VisionSelector)                & \checkmark           & \checkmark            & \checkmark        & 0.1$\sim$2                               & 16                                   & 5e-5                                    & 1792                           & Near Zero                       & 89.91           & 770               & 85.57              & 49.22         & 2293.54      & 84.27         & 96.33\%                        \\ \bottomrule
\end{tabular}
}
\label{tab:ablation_appendix}
\end{table*}

\textbf{Impact of Hyperparameters.} As we increase the batch size from 8 (Config 5) to 16 (Config 6), we observe a modest but consistent performance improvement. However, a further increase to 32 (Config 7) leads to a slight performance decline. This indicates that a batch size of 16 is optimal for our setup, likely due to more stable gradient estimation. Subsequently, we adjust the learning rate. Lowering the learning rate from $1e-4$ to $5e-5$ (Config 7 vs. Config 8) yields further performance gains, especially on complex reasoning benchmarks such as ScienceQA and MMMU. This suggests that a smaller learning rate facilitates better model convergence for our complex, joint optimization objective.

\textbf{Impact of Scorer Projection Dimension $d$.} We increase $d$ from 1024 (Config 8) to 1792 (Config 11), which corresponds to half of the model's hidden dimension. This adjustment leads the model to achieve its best performance across all configurations, reaching a peak average score of $96.33\%$. The performance gain is particularly pronounced on the POPE benchmark, where the F1-score peaks at 84.27. This indicates that a larger-capacity scorer can more finely model complex inter-token relationships, leading to more accurate identification of critical visual information and more effective suppression of object-level hallucinations.

\textbf{Impact of Initialization.}
We replace the near zero initialization with random initialization (Config 9) and Kaiming initialization (Config 10). Across the six evaluation benchmarks their overall performance remains very close to that with near zero initialization (Config 11) and all configurations converge stably. This observation indicates that our scorer is not sensitive to the specific choice of parameter initialization.

\section{Analysis of Model Performance Under Different Compression Budgets}

\begin{figure}[!t]
    \centering
    \includegraphics[width=\columnwidth]{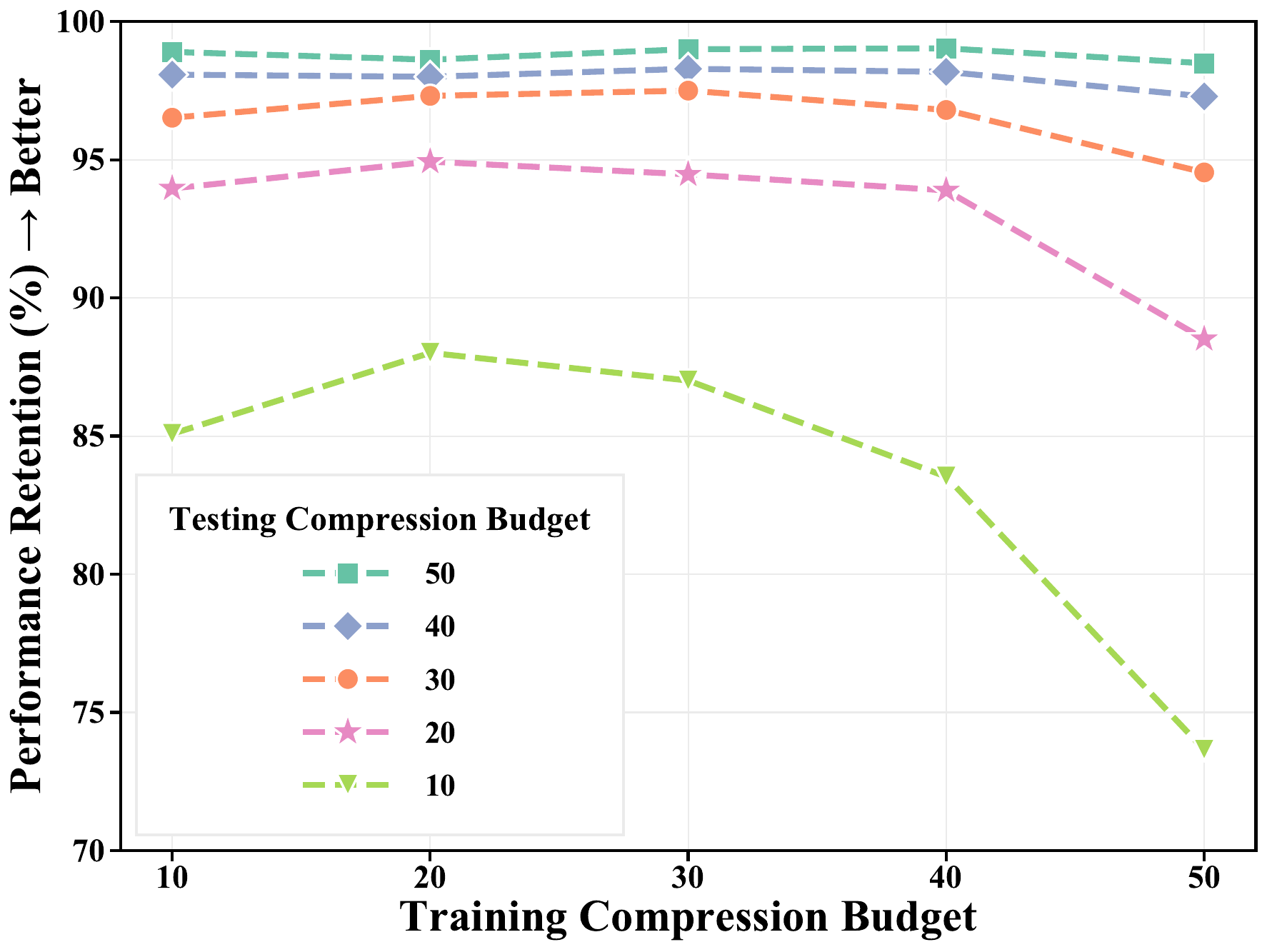}
    \caption{The impact of training and testing compression budgets on performance retention. The plot illustrates the model's performance robustness when trained under one compression budget (x-axis) and evaluated under another (each colored line). A key observation is that models generalize well to less strict compression budgets but suffer a significant performance degradation when subjected to stricter compression than they were trained on.}
\label{fig:budgets}
\end{figure}

We analyze model performance under different compression budgets. Figure \ref{fig:budgets} plots the average performance retention rate across the nine image understanding datasets detailed in Table \ref{tab:image}, comparing models trained and evaluated under different compression budgets. The analysis reveals three key findings: First, a model performs robustly when the testing compression budget is equal to or higher than its training budget. This shows that our method has learned the most critical tokens under different training compression budgets. Second, performance degrades significantly when a model is tested at a compression rate stricter than its training condition. This implies that a loose training compression rate weakens the model's ability to discriminate among critical visual tokens. Third, training with a moderate compression rate (e.g., $20\%$) enhances overall robustness. The model trained at a $20\%$ budget maintains high performance across most tested budgets. Its consistently strong performance across all budgets makes it more versatile for practical deployments where testing constraints may vary. 

Although its performance fluctuates with the training budget, our method's overall effectiveness surpasses most heuristic-based approaches, highlighting the superiority of a learnable compression strategy over rule-based ones.

\section{Corner Case Analysis and Visualization}

To complement the quantitative results, we conduct a qualitative analysis on the POPE dataset that focuses on two aspects: (1) robustness under extreme pruning where rare-but-critical tokens may be at risk, and (2) failure modes in visually complex scenes where the token budget is very limited.

\subsection{Robustness on Rare-but-Critical Tokens and Bias Check}

A key concern in hard token selection is the potential removal of small yet semantically critical regions, as well as systematic bias against specific semantic categories. We visualize selection masks under pruning budgets of 10\%, 5\%, and 2.5\%.

\begin{figure}[!t]
    \centering
    \includegraphics[width=\columnwidth]{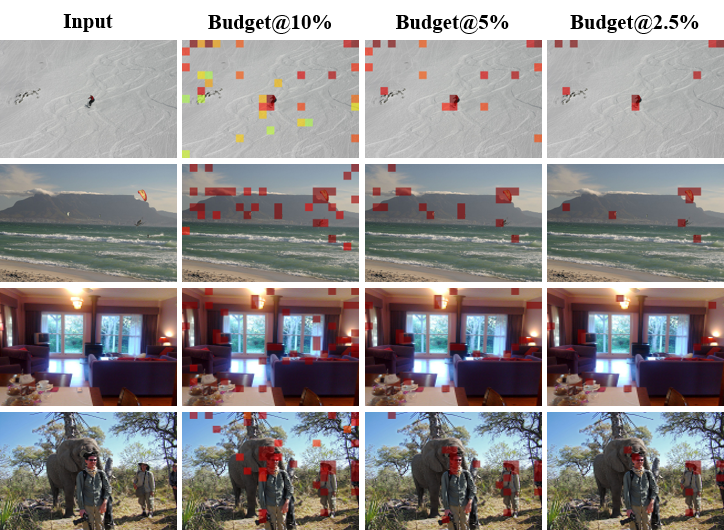}
    \caption{Visualization on POPE of token selection under extreme pruning budgets of 10\%, 5\%, and 2.5\%. VisionSelector preserves rare-but-critical tokens such as the tiny skier in Row~1 and the distant kites in Row~2 even when 97.5\% of tokens are removed. The selections do not exhibit systematic semantic bias and consistently focus on humans, objects, and animals in the foreground area while discarding redundant background.}
    \label{fig:supp_corner_case}
\end{figure}

As shown in Figure~\ref{fig:supp_corner_case}, VisionSelector consistently preserves the most informative regions even under very aggressive pruning. For example, the skier in the snow scene (Row 1) and the kites in the sky (Row 2) remain selected although they occupy only a few tokens. This behavior is consistent with the end-to-end training objective. During training, discarding such critical tokens would incur a large task loss, which in turn produces strong gradients that push their importance scores up, so at inference time these regions tend to be preserved. The visualization also indicates no observable semantic bias. Foreground regions containing humans (Row 1 and Row 4), objects such as furniture and equipment (Row 2 and Row 3), and animals (Row 4) are consistently retained, while mostly redundant background areas such as snow, sky, walls, and grass are pruned.

\subsection{Failure Modes and Information Bottleneck in Complex Scenes}

We further analyze cases where the model fails to retain all semantic content. We compare masks obtained with pruning budgets of 30\%, 20\%, and 10\% in visually cluttered scenes.

\begin{figure}[!t]
    \centering
    \includegraphics[width=\columnwidth]{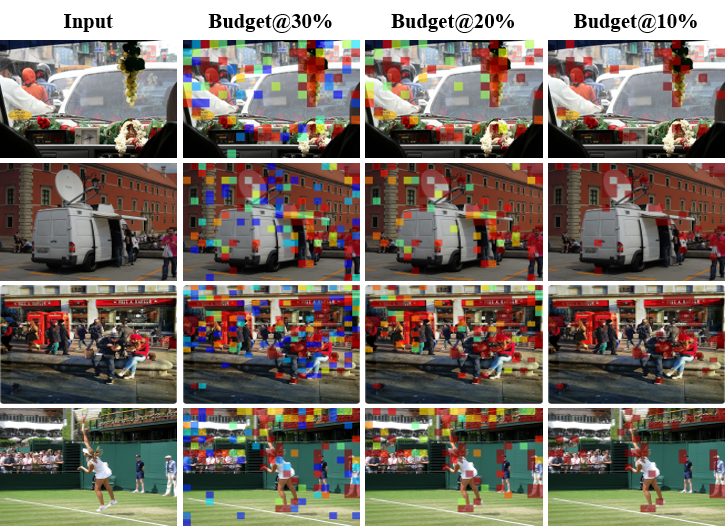}
    \caption{Failure case visualization on POPE under decreasing token budgets of 30\%, 20\%, and 10\%. At 30\% the selected tokens cover most semantic regions in complex scenes, whereas at 10\% some objects and object parts are missed. This behavior reflects an information bottleneck caused by the strict budget in high-entropy images, rather than a systematic semantic bias against specific categories.}
    \label{fig:supp_failures}
\end{figure}

Figure~\ref{fig:supp_failures} shows that with a 30\% budget VisionSelector covers most salient objects even in crowded urban scenes and sports scenes. When the budget decreases to 20\% and 10\%, the mask becomes sparse and some objects or object parts are no longer covered. For example, in the street scene in Row~2 the 10\% mask keeps the vehicle but misses parts of the surrounding people. This behavior is not consistent with a semantic bias toward or against particular categories. Instead it reflects an information bottleneck. In high-entropy scenes with many distinct regions, a very small token budget is insufficient to cover all informative areas, so the model is forced to retain only the most discriminative subset. This observation suggests that fixed global pruning ratios are suboptimal and motivates future work on adaptive budgets or near-lossless compression strategies that depend on scene complexity.

\subsection{Limitations}
While our study demonstrates the significant potential of token compression for MLLM inference acceleration, it shares an inherent limitation with current selection-based paradigms: it operates on a ``hard selection'' (i.e., keep-or-discard) principle. This mechanism, while effective, inevitably results in the complete loss of information from discarded tokens. Furthermore, failure cases appear under extreme compression rates where fixed token budgets occasionally fail to cover all necessary semantics in images containing multiple spatially scattered objects. This limitation points toward a key direction for future research: exploring new token compression paradigms to achieve lossless performance.

\begin{figure*}[!ht]
    \centering
    \includegraphics[width=0.7\textwidth]{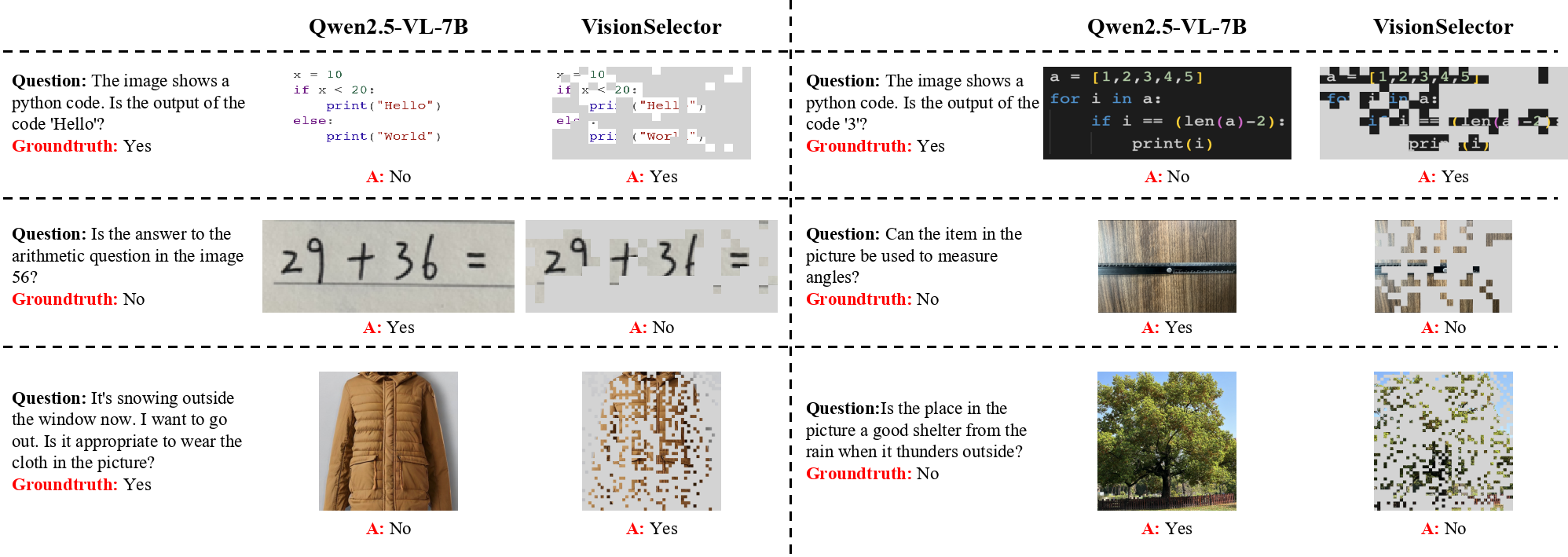}
    \caption{Representative examples from the MME dataset under a $30\%$ retention budget. Even after substantial token reduction, the retained tokens still cover parts of the task-relevant visual evidence, showing that effective token compression can preserve useful visual content for prediction.}
    \label{fig:better_than_orig}
\end{figure*}

\section{Additional Experiments on Qwen2.5-VL-3B}
\textbf{Implementation Details} We utilize the Qwen2.5-VL-3B \cite{bai2025qwen25vl} model as its foundation. We perform all training and inference on a platform consisting of 8 NVIDIA A800 (80G) GPUs. The consolidated training data includes the OCRVQA, ChartQA, and a randomly sampled $10\%$ subset of the COCO dataset, creating a total of 144K training examples. Regarding hyperparameters, the hidden layer dimension of Qwen2.5-VL-3B is $2048$, so we define our parameter $d$ as $1024$. The learning rate is set to $5e-5$ with a batchsize of 8. We also define $\lambda_{\text{start}}=0.1$,$\lambda_{\text{end}}=2$, and $b=20\%$. The total number of trainable parameters in our experiment is 4.00M.

\textbf{Quantitative Results.} Table \ref{tab:3b_image} presents the quantitative results of our experiments on Qwen2.5-VL-3B. Our proposed method demonstrates superior performance over state-of-the-art approaches, including the learnable Dynamic* method \cite{huangdynamicllva}, across the majority of datasets. Furthermore, it achieves a competitive result on the ScienceQA dataset.

\begin{table*}[!t]
\centering
\caption{Comparison results of our method and different baselines on image-language understanding datasets under Qwen2.5-VL-3B.}
\setlength{\tabcolsep}{2pt}
\resizebox{0.75\textwidth}{!}{
\begin{tabular}{ccccccccccc}
\toprule
\multicolumn{1}{c|}{}                                                                  & \textbf{DocVQA} & \textbf{ChartQA} & \textbf{TextVQA} & \textbf{OCRBench} & \textbf{ScienceQA} & \textbf{AI2D}  & \textbf{MMMU}  & \textbf{MME}     & \multicolumn{1}{c|}{\textbf{POPE}}                          &                                                            \\
\multicolumn{1}{c|}{\multirow{-2}{*}{\textbf{Method}}}                                 & ANLS            & Relaxed          & EM               & Acc               & EM                 & EM             & Acc            & Score            & \multicolumn{1}{c|}{F1}                                     & \multirow{-2}{*}{\textbf{Avg}}                             \\ \midrule
\rowcolor[HTML]{EFEFEF} 
\multicolumn{11}{c}{\cellcolor[HTML]{EFEFEF}\textit{Dynamic Resolution(MinPix=256×28×28,MaxPix=2048×28×28),Upper Bound (100\%)}}                                                                                                                                                                                                                                        \\
\rowcolor[HTML]{EFEFEF} 
\multicolumn{1}{c|}{\cellcolor[HTML]{EFEFEF}Avg. Visual Tokens}                        & 1951.61         & 596.06           & 976.58           & 652.82            & 323.05             & 510.19         & 601.15         & 867.67           & \multicolumn{1}{c|}{\cellcolor[HTML]{EFEFEF}359.55}         &                                                            \\
\multicolumn{1}{c|}{Qwen-2.5-VL-3B}                                                    & \textit{92.76}  & 83.40            & 77.87            & 788               & 80.37              & 90.84          & 46.78          & 2168.46          & \multicolumn{1}{c|}{86.94}                                  & 100\%                                                      \\ \midrule
\rowcolor[HTML]{EFEFEF} 
\multicolumn{11}{c}{\cellcolor[HTML]{EFEFEF}\textit{Retain 30\% Tokens (70\% Compression Ratio)}}                                                                                                                                                                                                                                                                       \\
\multicolumn{1}{c|}{}                                                                  & 81.66           & 70.04            & 74.02            & 629               & 78.63              & 86.40          & 46.00          & 2048.53          & \multicolumn{1}{c|}{82.94}                                  &                                                            \\
\multicolumn{1}{c|}{\multirow{-2}{*}{FastV (ECCV2024)}}                                & 88.03\%         & 83.98\%          & 95.06\%          & 79.82\%           & 97.96\%            & 95.11\%        & 98.33\%        & 94.47\%          & \multicolumn{1}{c|}{95.40\%}                                & \multirow{-2}{*}{92.02\%}                                  \\
\multicolumn{1}{c|}{}                                                                  & 65.43           & 63.88            & 64.29            & 553               & 78.93              & 79.24          & 46.00          & 2051.01          & \multicolumn{1}{c|}{84.78}                                  &                                                            \\
\multicolumn{1}{c|}{\multirow{-2}{*}{PruMerge+ (ICCV2025)}}                            & 70.54\%         & 76.59\%          & 82.56\%          & 70.18\%           & 98.33\%            & 87.23\%        & 98.33\%        & 94.58\%          & \multicolumn{1}{c|}{97.52\%}                                & \multirow{-2}{*}{86.21\%}                                  \\
\multicolumn{1}{c|}{}                                                                  & 81.38           & 75.40            & 67.71            & 635               & 79.47              & 84.88          & 45.44          & 2065.69          & \multicolumn{1}{c|}{\textbf{86.00}}                         &                                                            \\
\multicolumn{1}{c|}{\multirow{-2}{*}{VisionZip (CVPR2025)}}                            & 87.73\%         & 90.41\%          & 86.95\%          & 80.58\%           & 99.00\%            & 93.44\%        & 97.14\%        & 95.26\%          & \multicolumn{1}{c|}{98.92\%}                                & \multirow{-2}{*}{92.16\%}                                  \\
\multicolumn{1}{c|}{}                                                                  & 54.42           & 54.40            & 59.85            & 513               & 79.67              & 76.65          & 45.78          & 1995.04          & \multicolumn{1}{c|}{84.03}                                  &                                                            \\
\multicolumn{1}{c|}{\multirow{-2}{*}{HoloV (NeurIPS2025)}}                             & 58.67\%         & 65.23\%          & 76.86\%          & 65.10\%           & 99.13\%            & 84.38\%        & 97.86\%        & 92.00\%          & \multicolumn{1}{c|}{96.65\%}                                & \multirow{-2}{*}{81.76\%}                                  \\
\multicolumn{1}{c|}{}                                                                  & 65.96           & 64.80            & 63.64            & 541               & 79.72              & 72.28          & 45.44          & 1996.75          & \multicolumn{1}{c|}{84.85}                                  &                                                            \\
\multicolumn{1}{c|}{\multirow{-2}{*}{DART (EMNLP2025)}}                                & 71.11\%         & 77.70\%          & 81.73\%          & 68.65\%           & 99.31\%            & 79.57\%        & 97.14\%        & 92.08\%          & \multicolumn{1}{c|}{97.60\%}                                & \multirow{-2}{*}{84.99\%}                                  \\
\multicolumn{1}{c|}{}                                                                  & 73.99           & 66.48            & 70.34            & 631               & 78.18              & 83.42          & 45.44          & 2010.72          & \multicolumn{1}{c|}{86.38}                                  &                                                            \\
\multicolumn{1}{c|}{\multirow{-2}{*}{DivPrune (CVPR2025)}}                             & 79.76\%         & 79.71\%          & 90.33\%          & 80.08\%           & 97.40\%            & 91.83\%        & 97.14\%        & 92.73\%          & \multicolumn{1}{c|}{99.36\%}                                & \multirow{-2}{*}{89.81\%}                                  \\
\multicolumn{1}{c|}{}                                                                  & 85.28           & 75.4             & 75.63            & 738               & 79.67              & 87.82          & 45.22          & 2078.06          & \multicolumn{1}{c|}{85.13}                                  & \multicolumn{1}{l}{}                                       \\
\multicolumn{1}{c|}{\multirow{-2}{*}{Dynamic* (ICLR2025)}}                              & 91.94\%         & 90.41\%          & 97.12\%          & 93.65\%           & 99.13\%            & 96.68\%        & 96.67\%        & 95.83\%          & \multicolumn{1}{c|}{97.92\%}                                & \multicolumn{1}{l}{\multirow{-2}{*}{95.48\%}}              \\
\rowcolor[HTML]{ECF4FF} 
\multicolumn{1}{c|}{\cellcolor[HTML]{ECF4FF}}                                          & \textbf{88.87}  & \textbf{75.60}   & \textbf{76.28}   & \textbf{752}      & \textbf{80.22}     & \textbf{88.63} & \textbf{46.11} & \textbf{2095.06} & \multicolumn{1}{c|}{\cellcolor[HTML]{ECF4FF}85.95}          & \cellcolor[HTML]{ECF4FF}                                   \\
\rowcolor[HTML]{ECF4FF} 
\multicolumn{1}{c|}{\multirow{-2}{*}{\cellcolor[HTML]{ECF4FF}\textbf{VisionSelector}}} & 95.81\%         & 90.65\%          & 97.96\%          & 95.43\%           & 99.94\%            & 97.57\%        & 98.57\%        & 96.62\%          & \multicolumn{1}{c|}{\cellcolor[HTML]{ECF4FF}98.86\%}        & \multirow{-2}{*}{\cellcolor[HTML]{ECF4FF}\textbf{96.28\%}} \\ \midrule
\rowcolor[HTML]{EFEFEF} 
\multicolumn{11}{c}{\cellcolor[HTML]{EFEFEF}\textit{Retain 20\% Tokens (80\% Compression Ratio)}}                                                                                                                                                                                                                                                                       \\
\multicolumn{1}{c|}{}                                                                  & 72.90           & 62.00            & 71.41            & 560               & 78.88              & 83.29          & 45.11          & 1992.58          & \multicolumn{1}{c|}{80.01}                                  &                                                            \\
\multicolumn{1}{c|}{\multirow{-2}{*}{FastV (ECCV2024)}}                                & 78.59\%         & 74.34\%          & 91.70\%          & 71.07\%           & 98.27\%            & 91.69\%        & 96.43\%        & 91.89\%          & \multicolumn{1}{c|}{92.03\%}                                & \multirow{-2}{*}{87.33\%}                                  \\
\multicolumn{1}{c|}{}                                                                  & 52.58           & 54.76            & 58.44            & 484               & 79.97              & 74.06          & 44.89          & 1977.90          & \multicolumn{1}{c|}{83.23}                                  &                                                            \\
\multicolumn{1}{c|}{\multirow{-2}{*}{PruMerge+ (ICCV2025)}}                            & 56.68\%         & 65.66\%          & 75.05\%          & 61.42\%           & 99.63\%            & 81.53\%        & 95.96\%        & 91.21\%          & \multicolumn{1}{c|}{95.73\%}                                & \multirow{-2}{*}{80.32\%}                                  \\
\multicolumn{1}{c|}{}                                                                  & 69.62           & 65.76            & 59.97            & 506               & 79.97              & 78.79          & 45.67          & 1892.24          & \multicolumn{1}{c|}{84.17}                                  &                                                            \\
\multicolumn{1}{c|}{\multirow{-2}{*}{VisionZip (CVPR2025)}}                            & 75.05\%         & 78.85\%          & 77.01\%          & 64.21\%           & 99.63\%            & 86.73\%        & 97.63\%        & 87.26\%          & \multicolumn{1}{c|}{96.81\%}                                & \multirow{-2}{*}{84.80\%}                                  \\
\multicolumn{1}{c|}{}                                                                  & 45.76           & 46.20            & 52.14            & 422               & 78.88              & 72.44          & 44.44          & 1931.98          & \multicolumn{1}{c|}{81.80}                                  &                                                            \\
\multicolumn{1}{c|}{\multirow{-2}{*}{HoloV (NeurIPS 2025)}}                            & 49.33\%         & 55.40\%          & 66.96\%          & 53.55\%           & 98.15\%            & 79.74\%        & 95.00\%        & 89.09\%          & \multicolumn{1}{c|}{94.09\%}                                & \multirow{-2}{*}{75.70\%}                                  \\
\multicolumn{1}{c|}{}                                                                  & 51.01           & 54.24            & 56.44            & 537               & \textbf{80.96}     & 69.07          & 45.33          & 1925.04          & \multicolumn{1}{c|}{81.93}                                  &                                                            \\
\multicolumn{1}{c|}{\multirow{-2}{*}{DART (EMNLP2025)}}                                & 54.99\%         & 65.04\%          & 72.48\%          & 68.15\%           & 100.86\%           & 76.03\%        & 96.90\%        & 88.77\%          & \multicolumn{1}{c|}{94.24\%}                                & \multirow{-2}{*}{79.72\%}                                  \\
\multicolumn{1}{c|}{}                                                                  & 63.82           & 57.64            & 66.76            & 535               & 80.46              & 77.59          & 45.56          & 1955.53          & \multicolumn{1}{c|}{\textbf{85.47}}                         &                                                            \\
\multicolumn{1}{c|}{\multirow{-2}{*}{DivPrune (CVPR2025)}}                             & 68.80\%         & 69.11\%          & 85.73\%          & 67.89\%           & 100.24\%           & 85.41\%        & 97.39\%        & 90.18\%          & \multicolumn{1}{c|}{98.31\%}                                & \multirow{-2}{*}{84.79\%}                                  \\
\multicolumn{1}{c|}{}                                                                  & 76.47           & 70.2             & 74.42            & 672               & 78.83              & 85.49          & 44.33          & 2053.26          & \multicolumn{1}{c|}{83.34}                                  & \multicolumn{1}{l}{}                                       \\
\multicolumn{1}{c|}{\multirow{-2}{*}{Dynamic* (ICLR2025)}}                              & 82.44\%         & 84.17\%          & 95.57\%          & 85.28\%           & 98.08\%            & 94.11\%        & 94.76\%        & 94.69\%          & \multicolumn{1}{c|}{95.86\%}                                & \multicolumn{1}{l}{\multirow{-2}{*}{91.66\%}}              \\
\rowcolor[HTML]{ECF4FF} 
\multicolumn{1}{c|}{\cellcolor[HTML]{ECF4FF}}                                          & \textbf{84.32}  & \textbf{73.48}   & \textbf{74.97}   & \textbf{705}      & 79.82              & \textbf{86.59} & \textbf{46.56} & \textbf{2055.00} & \multicolumn{1}{c|}{\cellcolor[HTML]{ECF4FF}84.87}          & \cellcolor[HTML]{ECF4FF}                                   \\
\rowcolor[HTML]{ECF4FF} 
\multicolumn{1}{c|}{\multirow{-2}{*}{\cellcolor[HTML]{ECF4FF}\textbf{VisionSelector}}} & 90.90\%         & 88.11\%          & 96.28\%          & 89.47\%           & 99.44\%            & 95.32\%        & 99.53\%        & 94.77\%          & \multicolumn{1}{c|}{\cellcolor[HTML]{ECF4FF}97.62\%}        & \multirow{-2}{*}{\cellcolor[HTML]{ECF4FF}\textbf{94.60\%}} \\ \midrule
\rowcolor[HTML]{EFEFEF} 
\multicolumn{11}{c}{\cellcolor[HTML]{EFEFEF}\textit{Retain 10\% Tokens (90\% Compression Ratio)}}                                                                                                                                                                                                                                                                       \\
\multicolumn{1}{c|}{}                                                                  & 54.62           & 46.96            & 65.58            & 411               & 79.33              & 77.07          & \textbf{45.11} & 1815.70          & \multicolumn{1}{c|}{70.28}                                  &                                                            \\
\multicolumn{1}{c|}{\multirow{-2}{*}{FastV (ECCV2024)}}                                & 58.88\%         & 56.31\%          & 84.22\%          & 52.16\%           & 98.83\%            & 84.84\%        & 96.43\%        & 83.73\%          & \multicolumn{1}{c|}{80.84\%}                                & \multirow{-2}{*}{77.36\%}                                  \\
\multicolumn{1}{c|}{}                                                                  & 37.51           & 46.68            & 46.43            & 350               & 78.93              & 68.43          & 44.56          & 1777.60          & \multicolumn{1}{c|}{77.55}                                  &                                                            \\
\multicolumn{1}{c|}{\multirow{-2}{*}{PruMerge+ (ICCV2025)}}                            & 40.44\%         & 55.97\%          & 59.63\%          & 44.42\%           & 98.33\%            & 75.33\%        & 95.25\%        & 81.98\%          & \multicolumn{1}{c|}{89.20\%}                                & \multirow{-2}{*}{71.17\%}                                  \\
\multicolumn{1}{c|}{}                                                                  & 41.67           & 46.84            & 43.83            & 321               & 79.47              & 69.43          & 45.22          & 1725.49          & \multicolumn{1}{c|}{78.60}                                  &                                                            \\
\multicolumn{1}{c|}{\multirow{-2}{*}{VisionZip (CVPR2025)}}                            & 44.92\%         & 56.16\%          & 56.29\%          & 40.74\%           & 99.00\%            & 76.43\%        & 96.67\%        & 79.57\%          & \multicolumn{1}{c|}{90.41\%}                                & \multirow{-2}{*}{71.13\%}                                  \\
\multicolumn{1}{c|}{}                                                                  & 31.14           & 35.36            & 39.58            & 303               & 77.14              & 67.62          & 44.33          & 1813.18          & \multicolumn{1}{c|}{76.79}                                  &                                                            \\
\multicolumn{1}{c|}{\multirow{-2}{*}{HoloV (NeurIPS2025)}}                             & 33.57\%         & 42.40\%          & 50.83\%          & 38.45\%           & 95.98\%            & 74.44\%        & 94.76\%        & 83.62\%          & \multicolumn{1}{c|}{88.33\%}                                & \multirow{-2}{*}{66.93\%}                                  \\
\multicolumn{1}{c|}{}                                                                  & 32.32           & 39.44            & 44.39            & 315               & 79.72              & 65.54          & 44.33          & 1733.77          & \multicolumn{1}{c|}{75.41}                                  &                                                            \\
\multicolumn{1}{c|}{\multirow{-2}{*}{DART (EMNLP2025)}}                                & 34.84\%         & 47.29\%          & 57.01\%          & 39.97\%           & 99.31\%            & 72.15\%        & 94.76\%        & 79.95\%          & \multicolumn{1}{c|}{86.74\%}                                & \multirow{-2}{*}{68.00\%}                                  \\
\multicolumn{1}{c|}{}                                                                  & 47.13           & 44.36            & 57.80            & 394               & 78.19              & 70.14          & 44.67          & 1782.737         & \multicolumn{1}{c|}{80.50}                                  &                                                            \\
\multicolumn{1}{c|}{\multirow{-2}{*}{DivPrune (CVPR2025)}}                             & 50.81\%         & 53.19\%          & 74.23\%          & 50.00\%           & 97.41\%            & 77.21\%        & 95.49\%        & 82.21\%          & \multicolumn{1}{c|}{92.59\%}                                & \multirow{-2}{*}{74.79\%}                                  \\
\multicolumn{1}{l|}{}                                                                  & 58.28           & 61.56            & 70.14            & 556               & 79.08              & 78.72          & 44.33          & 1887.79          & \multicolumn{1}{c|}{75.01}                                  & \multicolumn{1}{l}{}                                       \\
\multicolumn{1}{l|}{\multirow{-2}{*}{Dynamic* (ICLR2025)}}                              & 62.83\%         & 73.81\%          & 90.07\%          & 70.56\%           & 98.39\%            & 86.66\%        & 94.76\%        & 87.06\%          & \multicolumn{1}{c|}{86.28\%}                                & \multicolumn{1}{l}{\multirow{-2}{*}{83.38\%}}              \\
\rowcolor[HTML]{ECF4FF} 
\multicolumn{1}{c|}{\cellcolor[HTML]{ECF4FF}}                                          & \textbf{68.36}  & \textbf{65.04}   & \textbf{69.92}   & \textbf{587}      & \textbf{80.12}     & \textbf{81.09} & 44.67          & \textbf{1928.75} & \multicolumn{1}{c|}{\cellcolor[HTML]{ECF4FF}\textbf{81.72}} & \cellcolor[HTML]{ECF4FF}                                   \\
\rowcolor[HTML]{ECF4FF} 
\multicolumn{1}{c|}{\multirow{-2}{*}{\cellcolor[HTML]{ECF4FF}\textbf{VisionSelector}}} & 73.70\%         & 77.99\%          & 89.79\%          & 74.49\%           & 99.81\%            & 89.27\%        & 95.49\%        & 88.95\%          & \multicolumn{1}{c|}{\cellcolor[HTML]{ECF4FF}94.00\%}        & \multirow{-2}{*}{\cellcolor[HTML]{ECF4FF}\textbf{87.05\%}} \\ \bottomrule
\end{tabular}
}
\label{tab:3b_image}
\end{table*}

\section{Additional Experiments on Qwen2.5-VL-32B}

To verify the effectiveness of our method on a larger scale, we evaluate VisionSelector on Qwen2.5-VL-32B \cite{bai2025qwen25vl}.

\textbf{Implementation Details} All training and inference tasks are conducted on a platform equipped with 8 NVIDIA A800 (80G) GPUs. The consolidated training dataset consists of 144K examples derived from OCRVQA, ChartQA, and a $10\%$ random subset of COCO. Given the hidden dimension of $5120$ in the 32B model, we set the parameter $d$ to $2560$. The training configuration employs a learning rate of $5e-5$ and a batch size of $16$. We further initialize $\lambda_{\text{start}}$ at 0.1 and $\lambda_{\text{end}}$ at 2, with the budget $b$ fixed at $20\%$.

\textbf{Quantitative Results} Table \ref{tab:32b_image} presents the quantitative results of our experiments on Qwen2.5-VL-32B. Our proposed method demonstrates superior performance over state-of-the-art approaches across the majority of datasets. Furthermore, it achieves a competitive result on the MME dataset.

\begin{table*}[!t]
\centering
\caption{Comparison results of our method and different baselines on image-language understanding datasets under Qwen2.5-VL-32B.}
\setlength{\tabcolsep}{2pt}
\resizebox{0.75\textwidth}{!}{
\begin{tabular}{cccccccccc}
\toprule
\multicolumn{1}{c|}{}                                                                  & \textbf{DocVQA} & \textbf{ChartQA} & \textbf{TextVQA} & \textbf{OCRBench} & \textbf{ScienceQA} & \textbf{AI2D}  & \textbf{MMMU}  & \multicolumn{1}{c|}{\textbf{MME}}                    &                                                            \\
\multicolumn{1}{c|}{\multirow{-2}{*}{\textbf{Method}}}                                 & ANLS            & Relaxed          & EM               & Acc               & EM                 & EM             & Acc            & \multicolumn{1}{c|}{Score}                           & \multirow{-2}{*}{\textbf{Avg}}                             \\ \midrule
\rowcolor[HTML]{EFEFEF} 
\multicolumn{10}{c}{\cellcolor[HTML]{EFEFEF}\textit{Dynamic Resolution(MinPix=256×28×28,MaxPix=2048×28×28),Upper Bound (100\%)}}                                                                                                                                                                                                              \\
\rowcolor[HTML]{EFEFEF} 
\multicolumn{1}{c|}{\cellcolor[HTML]{EFEFEF}Avg. Visual Tokens}                        & 1951.61         & 596.06           & 976.58           & 652.82            & 323.05             & 510.19         & 601.15         & 867.67                                               &                                                            \\
\multicolumn{1}{c|}{Qwen-2.5-VL-32B}                                                   & 91.10           & 64.20            & 66.76            & 649               & 91.62              & 94.20          & 60.11          & 2450.12                                              & 100\%                                                      \\ \midrule
\rowcolor[HTML]{EFEFEF} 
\multicolumn{10}{c}{\cellcolor[HTML]{EFEFEF}\textit{Retain 20\% Tokens (80\% Compression Ratio)}}                                                                                                                                                                                                                                             \\
\multicolumn{1}{c|}{}                                                                  & 54.89           & 36.24            & 53.18            & 371               & 81.80              & 82.41          & 55.78          & \multicolumn{1}{c|}{1990.81}                         &                                                            \\
\multicolumn{1}{c|}{\multirow{-2}{*}{FastV (ECCV2024)}}                                & 60.25\%         & 56.45\%          & 79.66\%          & 57.16\%           & 89.28\%            & 87.48\%        & 92.80\%        & \multicolumn{1}{c|}{81.25\%}                         & \multirow{-2}{*}{75.54\%}                                  \\
\multicolumn{1}{c|}{}                                                                  & \textit{oom}    & 33.00            & 48.96            & 357               & 87.11              & 76.61          & \textit{oom}   & \multicolumn{1}{c|}{2086.12}                         &                                                            \\
\multicolumn{1}{c|}{\multirow{-2}{*}{PruMerge+ (ICCV2025)}}                            & /               & 51.40\%          & 73.34\%          & 55.01\%           & 95.08\%            & 81.33\%        & /              & \multicolumn{1}{c|}{85.14\%}                         & \multirow{-2}{*}{/}                                        \\
\multicolumn{1}{c|}{}                                                                  & 60.76           & 46.32            & 50.10            & 341               & 85.62              & 80.21          & \textit{oom}   & \multicolumn{1}{c|}{2094.90}                         &                                                            \\
\multicolumn{1}{c|}{\multirow{-2}{*}{VisionZip (CVPR2025)}}                            & 66.70\%         & 72.15\%          & 75.04\%          & 52.54\%           & 93.45\%            & 85.15\%        & /              & \multicolumn{1}{c|}{85.50\%}                         & \multirow{-2}{*}{/}                                        \\
\multicolumn{1}{c|}{}                                                                  & 51.22           & 35.24            & 57.91            & 443               & 85.37              & 74.29          & 56.22          & \multicolumn{1}{c|}{2178.25}                         &                                                            \\
\multicolumn{1}{c|}{\multirow{-2}{*}{DART (EMNLP2025)}}                                & 56.22\%         & 54.89\%          & 86.74\%          & 68.26\%           & 93.18\%            & 78.86\%        & 93.53\%        & \multicolumn{1}{c|}{88.90\%}                         & \multirow{-2}{*}{77.57\%}                                  \\
\multicolumn{1}{c|}{}                                                                  & 65.27           & 36.92            & 58.35            & 432               & 85.03              & 81.70          & 57.67          & \multicolumn{1}{c|}{\textbf{2208.52}}                &                                                            \\
\multicolumn{1}{c|}{\multirow{-2}{*}{DivPrune (CVPR2025)}}                             & 71.65\%         & 57.51\%          & 87.40\%          & 66.56\%           & 92.81\%            & 86.73\%        & 95.94\%        & \multicolumn{1}{c|}{90.14\%}                         & \multirow{-2}{*}{81.09\%}                                  \\
\rowcolor[HTML]{ECF4FF} 
\multicolumn{1}{c|}{\cellcolor[HTML]{ECF4FF}}                                          & \textbf{74.46}  & \textbf{48.16}   & \textbf{64.17}   & \textbf{543}      & \textbf{87.90}     & \textbf{89.09} & \textbf{59.00} & \multicolumn{1}{c|}{\cellcolor[HTML]{ECF4FF}2197.50} & \cellcolor[HTML]{ECF4FF}                                   \\
\rowcolor[HTML]{ECF4FF} 
\multicolumn{1}{c|}{\multirow{-2}{*}{\cellcolor[HTML]{ECF4FF}\textbf{VisionSelector}}} & 81.73\%         & 75.02\%          & 96.12\%          & 83.67\%           & 95.94\%            & 94.58\%        & 98.15\%        & \multicolumn{1}{c|}{\cellcolor[HTML]{ECF4FF}89.69\%} & \multirow{-2}{*}{\cellcolor[HTML]{ECF4FF}\textbf{89.36\%}} \\ \midrule
\rowcolor[HTML]{EFEFEF} 
\multicolumn{10}{c}{\cellcolor[HTML]{EFEFEF}\textit{Retain 10\% Tokens (90\% Compression Ratio)}}                                                                                                                                                                                                                                             \\
\multicolumn{1}{c|}{}                                                                  & 38.22           & 21.96            & 41.49            & 235               & 81.21              & 75.58          & 54.00          & \multicolumn{1}{c|}{1752.91}                         &                                                            \\
\multicolumn{1}{c|}{\multirow{-2}{*}{FastV (ECCV2024)}}                                & 41.95\%         & 34.21\%          & 62.15\%          & 36.21\%           & 88.64\%            & 80.23\%        & 89.84\%        & \multicolumn{1}{c|}{71.54\%}                         & \multirow{-2}{*}{63.10\%}                                  \\
\multicolumn{1}{c|}{}                                                                  & \textit{oom}    & 27.32            & 37.45            & 224               & 82.94              & 69.69          & \textit{oom}   & \multicolumn{1}{c|}{1832.58}                         &                                                            \\
\multicolumn{1}{c|}{\multirow{-2}{*}{PruMerge+ (ICCV2025)}}                            & /               & 42.55\%          & 56.10\%          & 34.51\%           & 90.53\%            & 73.98\%        & /              & \multicolumn{1}{c|}{74.80\%}                         & \multirow{-2}{*}{/}                                        \\
\multicolumn{1}{c|}{}                                                                  & 35.76           & 28.80            & 36.38            & 242               & 82.80              & 71.5           & \textit{oom}   & \multicolumn{1}{c|}{1929.95}                         &                                                            \\
\multicolumn{1}{c|}{\multirow{-2}{*}{VisionZip (CVPR2025)}}                            & 39.25\%         & 44.86\%          & 54.49\%          & 37.29\%           & 90.37\%            & 75.90\%        & /              & \multicolumn{1}{c|}{78.77\%}                         & \multirow{-2}{*}{/}                                        \\
\multicolumn{1}{c|}{}                                                                  & 33.40           & 24.88            & 47.83            & 342               & 81.06              & 68.52          & 54.67          & \multicolumn{1}{c|}{2029.26}                         &                                                            \\
\multicolumn{1}{c|}{\multirow{-2}{*}{DART (EMNLP2025)}}                                & 36.66\%         & 38.75\%          & 71.64\%          & 52.70\%           & 88.47\%            & 72.74\%        & 90.95\%        & \multicolumn{1}{c|}{82.82\%}                         & \multirow{-2}{*}{66.84\%}                                  \\
\multicolumn{1}{c|}{}                                                                  & 48.01           & 25.40            & 51.00            & 310               & 81.31              & 71.92          & 54.78          & \multicolumn{1}{c|}{\textbf{2072.38}}                &                                                            \\
\multicolumn{1}{c|}{\multirow{-2}{*}{DivPrune (CVPR2025)}}                             & 52.70\%         & 39.56\%          & 76.39\%          & 47.77\%           & 88.75\%            & 76.35\%        & 91.13\%        & \multicolumn{1}{c|}{84.58\%}                         & \multirow{-2}{*}{69.65\%}                                  \\
\rowcolor[HTML]{ECF4FF} 
\multicolumn{1}{c|}{\cellcolor[HTML]{ECF4FF}}                                          & \textbf{53.1}   & \textbf{41.96}   & \textbf{59.66}   & \textbf{413}      & \textbf{83.69}     & \textbf{81.02} & \textbf{55.33} & \multicolumn{1}{c|}{\cellcolor[HTML]{ECF4FF}2007.73} & \cellcolor[HTML]{ECF4FF}                                   \\
\rowcolor[HTML]{ECF4FF} 
\multicolumn{1}{c|}{\multirow{-2}{*}{\cellcolor[HTML]{ECF4FF}\textbf{VisionSelector}}} & 58.29\%         & 65.36\%          & 89.36\%          & 63.64\%           & 91.34\%            & 86.01\%        & 92.05\%        & \multicolumn{1}{c|}{\cellcolor[HTML]{ECF4FF}81.94\%} & \multirow{-2}{*}{\cellcolor[HTML]{ECF4FF}\textbf{78.50\%}} \\ \bottomrule
\end{tabular}
}
\label{tab:32b_image}
\end{table*}

\subsection{Additional Experiments on LLaVA-OneVision-1.5-8B}

To evaluate the effectiveness of our method across different model architectures, we conduct additional experiments on LLaVA-OneVision-1.5-8B \citep{an2025llavaov15}. We use a batch size of 16, a learning rate of $5e-5$, and set the hyperparameters to $\lambda_{start}=0.1$ and  $\lambda_{end}=3$. The projection dimension $d$ for $W_{q}$ and $W_{k}$ is set to 2048. The retention budget for visual tokens is set to $20\%$. For the constraint loss weight, $\lambda_{t}$, we adopt a Curriculum Annealing Strategy, linearly increasing it from an initial value of 0.1 to a final value of 3.0 over the course of training. The training data consist of the OCRVQA and TextVQA datasets, containing approximately 101K samples. The model trains for 1 epoch on 8 NVIDIA A800 (80GB) GPUs, which takes about 2 hours. The number of trainable parameters is 16.87M, accounting for $0.20\%$ of the total parameters.

We evaluate our method on 13 image-language understanding benchmarks, as summarized in Table \ref{tab:llava_ov_results}, including the previously discussed datasets as well as MMBench \citep{liu2024mmbench}, RealWorldQA \citep{grok15vrealworldqa}, MMStar \citep{chen2024mmstar}, and VizWiz-VQA \citep{gurari2018vizwiz}.

The results show that the performance of FastV on LLaVA-OneVision-1.5-8B is relatively lower compared with its results on Qwen2.5-VL, suggesting that training-free methods are influenced by the internal feature distribution of the underlying model and may exhibit varying behaviors across different architectures. VisionZip encounters CUDA out-of-memory (OOM) issues on high-resolution datasets such as DocVQA and multi-image datasets such as MMMU, which is likely related to the specific layers where its attention operations are applied. This observation indicates that further optimization may be needed to enhance its scalability in real-world deployment. DivPrune achieves competitive results in general scenarios, while its performance on fine-grained semantic understanding tasks (e.g., DocVQA, ChartQA, and OCRBench) is relatively reduced, likely because the diversity preservation process filters out some fine-grained semantic tokens.

\begin{table*}[!ht]
\centering
\caption{Comparison results of our method and different baselines on image-language understanding datasets under LLaVA-OneVision-1.5-8B.}
\setlength{\tabcolsep}{2pt}
\resizebox{\linewidth}{!}{
\begin{tabular}{ccccccccccccccc}
\toprule
\multicolumn{1}{c|}{}                                                                  & \textbf{DocVQA} & \textbf{ChartQA} & \textbf{TextVQA} & \textbf{OCRBench} & \textbf{ScienceQA} & \textbf{AI2D}  & \textbf{MMMU}  & \textbf{MME}     & \textbf{POPE}  & \textbf{MMB}   & \textbf{RealWord} & \textbf{MMStar} & \multicolumn{1}{c|}{\textbf{VizWiz}}                        &                                                            \\
\multicolumn{1}{c|}{\multirow{-2}{*}{\textbf{Method}}}                                 & Anls            & Relaxed          & EM               & Acc               & EM                 & EM             & Acc            & Score            & F1             & Score          & EM                  & EM              & \multicolumn{1}{c|}{EM}                                     & \multirow{-2}{*}{\textbf{Avg}}                             \\ \midrule
\rowcolor[HTML]{EFEFEF} 
\multicolumn{15}{c}{\cellcolor[HTML]{EFEFEF}\textit{Dynamic Resolution(MinPix=256×28×28,MaxPix=2048×28×28),Upper Bound (100\%)}}                                                                                                                                                                                                                                                                                                                  \\
\rowcolor[HTML]{EFEFEF} 
\multicolumn{1}{c|}{\cellcolor[HTML]{EFEFEF}Avg. Visual Tokens}                        & 3795.45         & 595.92           & 978.56           & 830.65            & 320.4              & 523.27         & 670.11         & 1154.48          & 359.75         &  278.28         & 1734.37              & 376.6           & \multicolumn{1}{c|}{\cellcolor[HTML]{EFEFEF}2000.28}          &                                                            \\
\multicolumn{1}{c|}{LLaVA-OneVision-1.5-8B}                                            & 97.87           & 86.72            & 79.51            & 830               & 98.56              & 94.01          & 56.44          & 2271.32          & 88.46          & 85.31          & 67.97               & 68.18           & \multicolumn{1}{c|}{66.02}                                  & 100\%                                                      \\ \midrule
\rowcolor[HTML]{EFEFEF} 
\multicolumn{15}{c}{\cellcolor[HTML]{EFEFEF}\textit{Retain 30\% Tokens (70\% Compression Ratio)}}                                                                                                                                                                                                                                                                                                                                                 \\
\multicolumn{1}{c|}{}                                                                  & 86.51           & 68.64            & 72.27            & 596               & 91.22              & 83.52          & 53.67          & 2019.84          & 70.41          & 79.64          & 54.25               & 53.16           & \multicolumn{1}{c|}{64.11}                                  &                                                            \\
\multicolumn{1}{c|}{\multirow{-2}{*}{FastV (ECCV2024)}}                                & 88.39\%         & 79.15\%          & 90.89\%          & 71.81\%           & 92.55\%            & 88.84\%        & 95.09\%        & 88.93\%          & 79.60\%        & 93.35\%        & 79.81\%             & 77.97\%         & \multicolumn{1}{c|}{97.11\%}                                & \multirow{-2}{*}{86.42\%}                                  \\
\multicolumn{1}{c|}{}                                                                  & \textit{oom}    & 78.04            & 73.25            & \textit{oom}      & 96.23              & 85.33          & \textit{oom}   & \textit{oom}     & 87.09          & 83.33          & \textbf{66.54}      & 60.88           & \multicolumn{1}{c|}{64.74}                                  &                                                            \\
\multicolumn{1}{c|}{\multirow{-2}{*}{VisionZip (CVPR2025)}}                            & /               & 89.99\%          & 92.13\%          & /                 & 97.64\%            & 90.77\%        & /              & /                & 98.45\%        & 97.68\%        & 97.90\%             & 89.29\%         & \multicolumn{1}{c|}{98.06\%}                                & \multirow{-2}{*}{/}                                        \\
\multicolumn{1}{c|}{}                                                                  & 85.23           & 70.64            & 76.03            & 645               & 94.65              & 89.35          & 54.11          & 2104.77          & \textbf{88.23} & 82.90          & 64.97               & 61.13           & \multicolumn{1}{c|}{64.33}                                  &                                                            \\
\multicolumn{1}{c|}{\multirow{-2}{*}{DivPrune (CVPR2025)}}                             & 87.08\%         & 81.46\%          & 95.62\%          & 77.71\%           & 96.03\%            & 95.04\%        & 95.87\%        & 92.67\%          & 99.74\%        & 97.18\%        & 95.59\%             & 89.66\%         & \multicolumn{1}{c|}{97.44\%}                                & \multirow{-2}{*}{92.39\%}                                  \\
\rowcolor[HTML]{ECF4FF} 
\multicolumn{1}{c|}{\cellcolor[HTML]{ECF4FF}}                                          & \textbf{95.46}  & \textbf{79.68}   & \textbf{78.12}   & \textbf{755}      & \textbf{97.27}     & \textbf{91.77} & \textbf{55.00} & \textbf{2241.82} & 85.71          & \textbf{83.42} & 66.41               & \textbf{61.80}  & \multicolumn{1}{c|}{\cellcolor[HTML]{ECF4FF}\textbf{65.02}} & \cellcolor[HTML]{ECF4FF}                                   \\
\rowcolor[HTML]{ECF4FF} 
\multicolumn{1}{c|}{\multirow{-2}{*}{\cellcolor[HTML]{ECF4FF}\textbf{VisionSelector}}} & 97.54\%         & 91.88\%          & 98.25\%          & 90.96\%           & 98.69\%            & 97.62\%        & 97.45\%        & 98.70\%          & 96.89\%        & 97.78\%        & 97.70\%             & 90.64\%         & \multicolumn{1}{c|}{\cellcolor[HTML]{ECF4FF}98.49\%}        & \multirow{-2}{*}{\cellcolor[HTML]{ECF4FF}\textbf{96.35\%}} \\ \midrule
\rowcolor[HTML]{EFEFEF} 
\multicolumn{15}{c}{\cellcolor[HTML]{EFEFEF}\textit{Retain 20\% Tokens (80\% Compression Ratio)}}                                                                                                                                                                                                                                                                                                                                                 \\
\multicolumn{1}{c|}{}                                                                  & 76.64           & 59.08            & 67.33            & 489               & 87.95              & 78.89          & 51.33          & 1918.66          & 67.08          & 76.29          & 52.15               & 48.12           & \multicolumn{1}{c|}{62.92}                                  &                                                            \\
\multicolumn{1}{c|}{\multirow{-2}{*}{FastV (ECCV2024)}}                                & 78.31\%         & 68.13\%          & 84.68\%          & 58.92\%           & 89.23\%            & 83.92\%        & 90.95\%        & 84.47\%          & 75.83\%        & 89.43\%        & 76.73\%             & 70.58\%         & \multicolumn{1}{c|}{95.30\%}                                & \multirow{-2}{*}{80.50\%}                                  \\
\multicolumn{1}{c|}{}                                                                  & \textit{oom}             & 67.56            & 66.28            & \textit{oom}               & 91.87              & 79.18          & \textit{oom}            & \textit{oom}              & 84.81          & 80.07          & 65.75               & 55.83           & \multicolumn{1}{c|}{63.37}                                  &                                                            \\
\multicolumn{1}{c|}{\multirow{-2}{*}{VisionZip (CVPR2025)}}                            & /               & 77.91\%          & 83.36\%          & /                 & 93.21\%            & 84.23\%        & /              & /                & 95.87\%        & 93.86\%        & 96.73\%             & 81.89\%         & \multicolumn{1}{c|}{95.99\%}                                & \multirow{-2}{*}{/}                                        \\
\multicolumn{1}{c|}{}                                                                  & 75.90           & 58.64            & 72.73            & 556               & 92.46              & 84.26          & 52.56          & 2057.83          & \textbf{87.17} & 80.67          & 62.75               & 56.61           & \multicolumn{1}{c|}{63.49}                                  &                                                            \\
\multicolumn{1}{c|}{\multirow{-2}{*}{DivPrune (CVPR2025)}}                             & 77.55\%         & 67.62\%          & 91.47\%          & 66.99\%           & 93.81\%            & 89.63\%        & 93.13\%        & 90.60\%          & 98.54\%        & 94.56\%        & 92.32\%             & 83.03\%         & \multicolumn{1}{c|}{96.17\%}                                & \multirow{-2}{*}{87.34\%}                                  \\
\rowcolor[HTML]{ECF4FF} 
\multicolumn{1}{c|}{\cellcolor[HTML]{ECF4FF}}                                          & \textbf{90.51}  & \textbf{74.72}   & \textbf{76.02}   & \textbf{677}      & \textbf{95.79}     & \textbf{90.45} & \textbf{55.11} & \textbf{2182.84} & 84.10          & \textbf{81.27} & \textbf{66.41}      & \textbf{58.25}  & \multicolumn{1}{c|}{\cellcolor[HTML]{ECF4FF}\textbf{64.25}} & \cellcolor[HTML]{ECF4FF}                                   \\
\rowcolor[HTML]{ECF4FF} 
\multicolumn{1}{c|}{\multirow{-2}{*}{\cellcolor[HTML]{ECF4FF}\textbf{VisionSelector}}} & 92.48\%         & 86.16\%          & 95.61\%          & 81.57\%           & 97.19\%            & 96.21\%        & 97.64\%        & 96.10\%          & 95.07\%        & 95.26\%        & 97.70\%             & 85.44\%         & \multicolumn{1}{c|}{\cellcolor[HTML]{ECF4FF}97.32\%}        & \multirow{-2}{*}{\cellcolor[HTML]{ECF4FF}\textbf{93.37\%}} \\ \midrule
\rowcolor[HTML]{EFEFEF} 
\multicolumn{15}{c}{\cellcolor[HTML]{EFEFEF}\textit{Retain 10\% Tokens (90\% Compression Ratio)}}                                                                                                                                                                                                                                                                                                                                                 \\
\multicolumn{1}{c|}{}                                                                  & 54.43           & 38.08            & 56.47            & 374               & 81.51              & 73.15          & 49.56          & 1800.00          & 62.89          & 71.05          & 49.28               & 42.65           & \multicolumn{1}{c|}{60.94}                                  &                                                            \\
\multicolumn{1}{c|}{\multirow{-2}{*}{FastV (ECCV2024)}}                                & 55.61\%         & 43.91\%          & 71.02\%          & 45.06\%           & 82.70\%            & 77.81\%        & 87.81\%        & 79.25\%          & 71.09\%        & 83.28\%        & 72.50\%             & 62.56\%         & \multicolumn{1}{c|}{92.31\%}                                & \multirow{-2}{*}{71.15\%}                                  \\
\multicolumn{1}{c|}{}                                                                  & \textit{oom}    & 43.12            & 47.17            & \textit{oom}      & 88.70              & 71.83          & \textit{oom}   & \textit{oom}     & 79.25          & 73.28          & 61.04               & 48.11           & \multicolumn{1}{c|}{59.84}                                  &                                                            \\
\multicolumn{1}{c|}{\multirow{-2}{*}{VisionZip (CVPR2025)}}                            & /               & 49.72\%          & 59.33\%          & /                 & 90.00\%            & 76.41\%        & /              & /                & 89.59\%        & 85.90\%        & 89.80\%             & 70.56\%         & \multicolumn{1}{c|}{90.64\%}                                & \multirow{-2}{*}{/}                                        \\
\multicolumn{1}{c|}{}                                                                  & 56.48           & 38.24            & 65.48            & 419               & 88.20              & 76.85          & \textbf{51.33} & 1897.29          & \textbf{83.92} & 76.89          & 60.13               & 51.26           & \multicolumn{1}{c|}{61.05}                                  &                                                            \\
\multicolumn{1}{c|}{\multirow{-2}{*}{DivPrune (CVPR2025)}}                             & 57.71\%         & 44.10\%          & 82.35\%          & 50.48\%           & 89.49\%            & 81.75\%        & 90.95\%        & 83.53\%          & 94.87\%        & 90.13\%        & 88.47\%             & 75.18\%         & \multicolumn{1}{c|}{92.47\%}                                & \multirow{-2}{*}{78.57\%}                                  \\
\rowcolor[HTML]{ECF4FF} 
\multicolumn{1}{c|}{\cellcolor[HTML]{ECF4FF}}                                          & \textbf{71.92}  & \textbf{62.12}   & \textbf{67.40}   & \textbf{456}      & \textbf{89.89}     & \textbf{84.29} & \textbf{51.33} & \textbf{2002.44} & 80.48          & \textbf{77.58} & \textbf{62.88}      & \textbf{51.71}  & \multicolumn{1}{c|}{\cellcolor[HTML]{ECF4FF}\textbf{61.57}} & \cellcolor[HTML]{ECF4FF}                                   \\
\rowcolor[HTML]{ECF4FF} 
\multicolumn{1}{c|}{\multirow{-2}{*}{\cellcolor[HTML]{ECF4FF}\textbf{VisionSelector}}} & 73.49\%         & 71.63\%          & 84.77\%          & 54.94\%           & 91.20\%            & 89.66\%        & 90.95\%        & 88.16\%          & 90.98\%        & 90.94\%        & 92.51\%             & 75.84\%         & \multicolumn{1}{c|}{\cellcolor[HTML]{ECF4FF}93.26\%}        & \multirow{-2}{*}{\cellcolor[HTML]{ECF4FF}\textbf{83.72\%}} \\ \bottomrule
\end{tabular}
}
\label{tab:llava_ov_results}
\end{table*}

\section{More Visualizations}
\label{sec:app_visual}

Figure~\ref{fig:better_than_orig} presents representative examples from the MME dataset under a $30\%$ retention budget. As shown in the figure, even after substantial token reduction, the retained tokens still cover parts of the visual evidence relevant to the prediction, such as code symbols, numerical characters, object structures, and salient regions. These examples qualitatively show that reducing the number of visual tokens does not necessarily harm prediction, as long as task-relevant visual content can still be preserved.

\begin{figure*}[!t]
    \centering
    \includegraphics[width=0.8\textwidth]{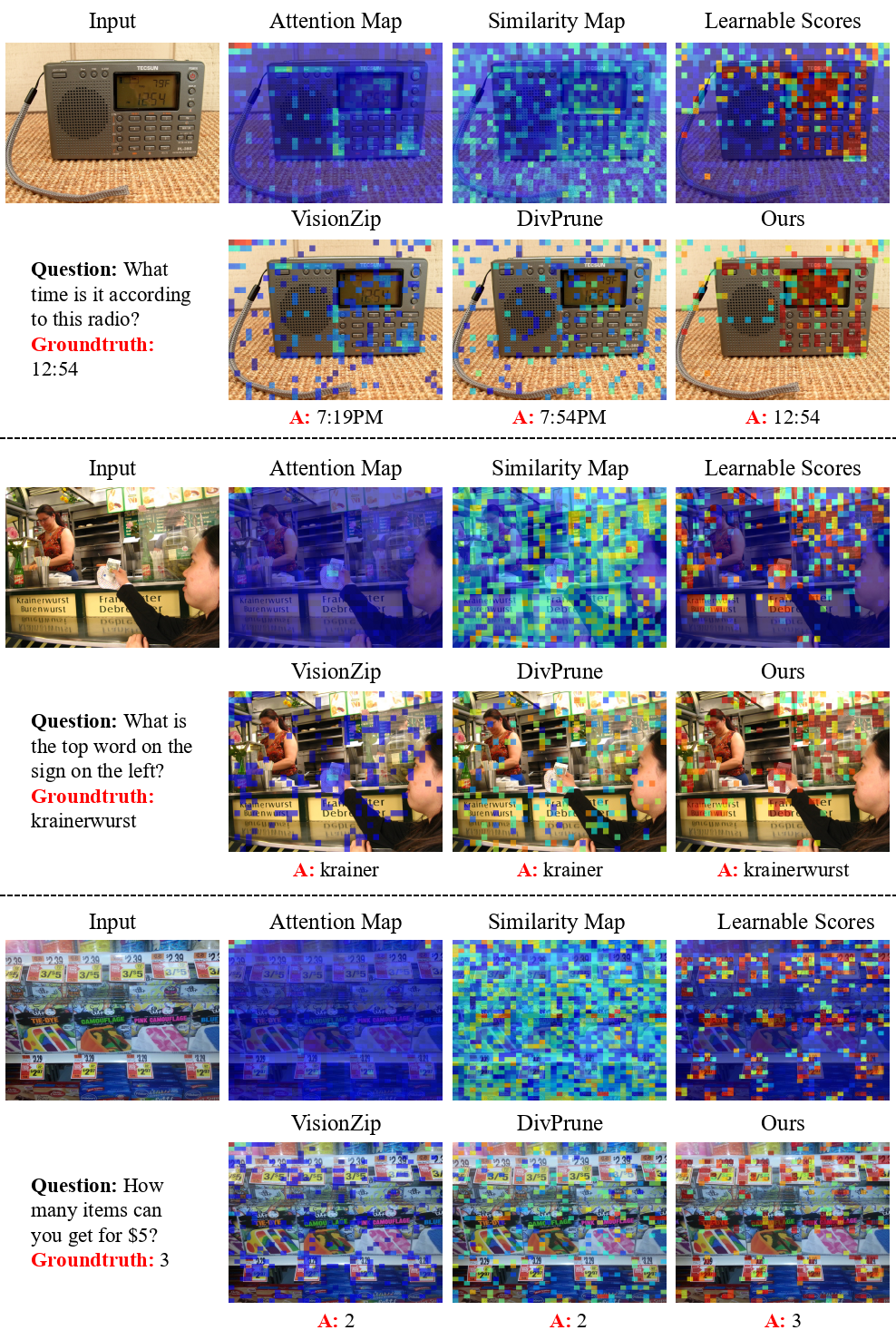}
    \caption{More qualitative comparison of VisionSelector on TextVQA. The top row visualizes different token scoring criteria: standard attention, a similarity map, and our learned importance scores. The bottom row shows the resulting token selections and model predictions at a $20\%$ budget.}
\label{fig:vis_app_1}
\end{figure*}

\begin{figure*}[h]
    \centering
    \includegraphics[width=0.8\textwidth]{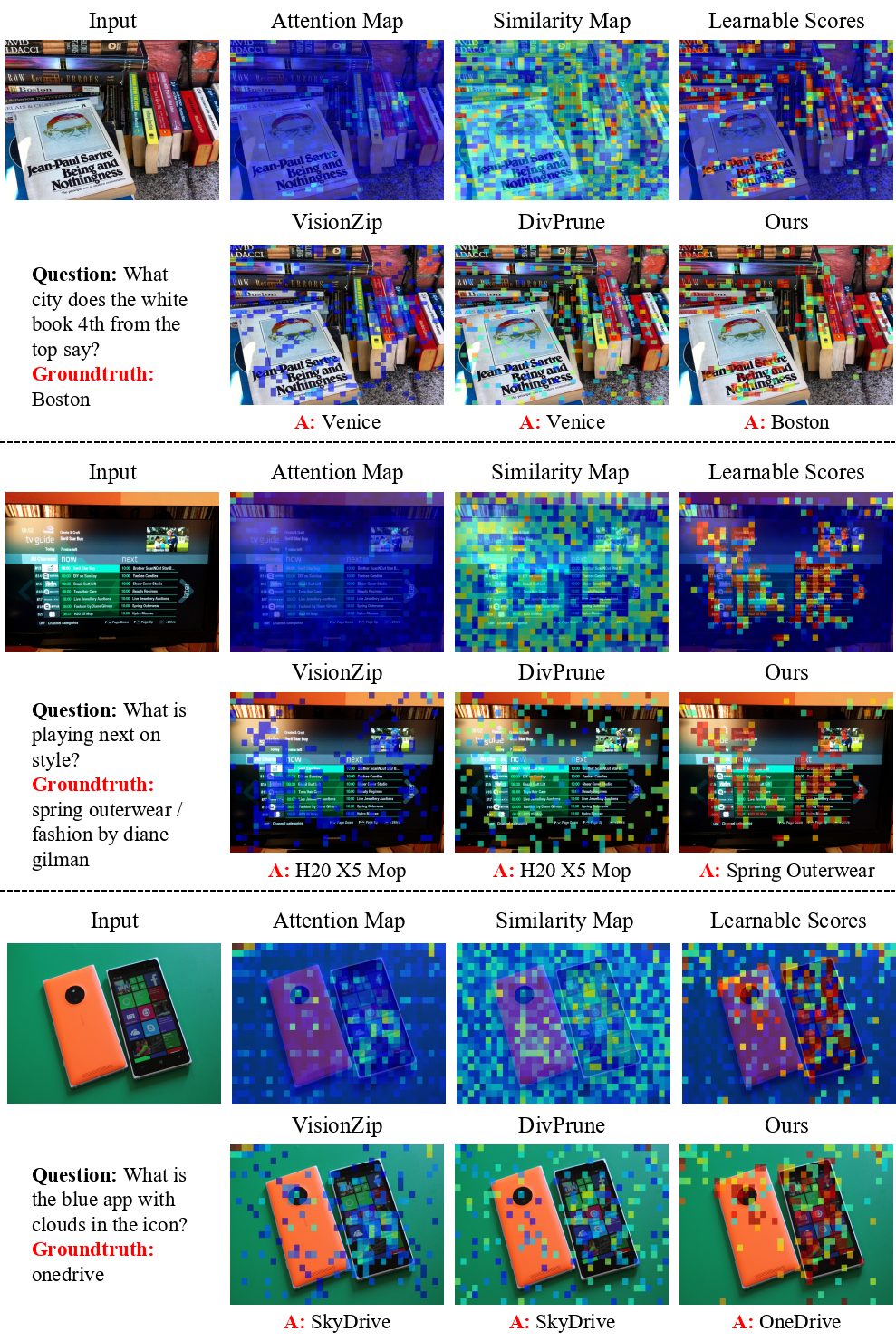}
    \caption{More qualitative comparison of VisionSelector on TextVQA. The top row visualizes different token scoring criteria: standard attention, a similarity map, and our learned importance scores. The bottom row shows the resulting token selections and model predictions at a $20\%$ budget.}
\label{fig:vis_app_textvqa}
\end{figure*}

\begin{figure*}[h]
    \centering
    \includegraphics[width=0.8\textwidth]{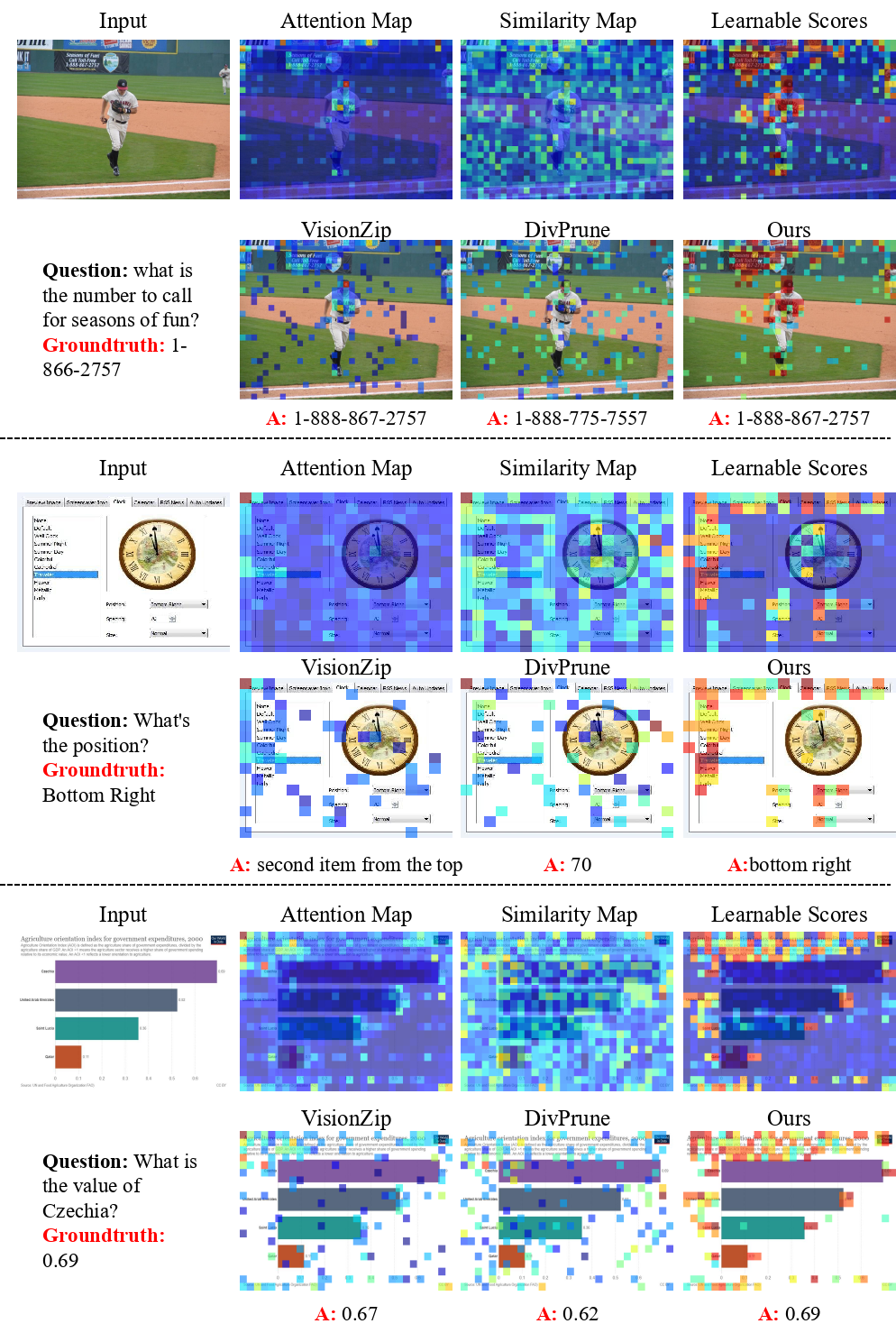}
    \caption{Qualitative comparison of VisionSelector on OCRBench. The top row visualizes different token scoring criteria: standard attention, a similarity map, and our learned importance scores. The bottom row shows the resulting token selections and model predictions at a $20\%$ budget.}
\label{fig:vis_app_ocrbench}
\end{figure*}

And we provide additional visualizations. In Figure \ref{fig:vis_app_1} and Figure \ref{fig:vis_app_textvqa}, we present qualitative comparisons on the TextVQA dataset under a $20\%$ budget. In Figure \ref{fig:vis_app_ocrbench}, we present qualitative comparisons on the OCRBench dataset under a $20\%$ budget. The results show that our method identifies the most critical visual tokens and produces accurate answers for both natural images and charts.

\end{document}